# From Sensors to Silicon: Causal Cross-Layer Self-Healing Edge AI for Secure and Energy-Adaptive IoT Systems

Mostafa Darvishi, *Senior Member, IEEE*

*Abstract*— **The migration of artificial intelligence from centralized cloud infrastructure to resource-constrained Internet-of-Things (IoT) devices reduces communication latency and improves data locality, but it also tightly couples sensing quality, inference behavior, microarchitectural activity, hardware reliability, security, and energy management within the same physical node. Existing edge-intelligence solutions predominantly address these dimensions in isolation, making it difficult to determine whether an observed degradation originates from sensor malfunction, data-distribution shift, malicious execution, hardware faults, or resource and thermal stress. This paper presents a causal cross-layer self-healing Edge-AI framework that continuously observes synchronized telemetry spanning sensor health, inference uncertainty, processor hardware-performance counters, accelerator and memory activity, power and thermal state, and communication conditions. A temporally informed causal reasoning engine estimates the most probable origin of abnormal behavior and quantifies diagnostic uncertainty, including under concurrent perturbations. The inferred cause and confidence are subsequently supplied to a risk-constrained recovery controller that selects among sensor failover, model and numerical-precision adaptation, dynamic voltage and frequency scaling, accelerator reset or reconfiguration, and selective edge–cloud migration while enforcing application-level latency, energy, inference-quality, and residual-risk constraints. The framework follows a local-first execution model in which time-critical diagnosis and recovery remain available without persistent cloud connectivity, while remote resources support slower model adaptation and knowledge aggregation. A heterogeneous FPGA-SoC prototype is used to evaluate the architecture across vision, audio, and time-series Edge-AI workloads under controlled sensor, cyber, hardware, resource, and compound perturbations. Evaluation considers root-cause identification, macro-F1 and AUPRC, detection and recovery latency, false-recovery rate, service availability, energy per correct inference, communication cost, and hardware overhead, together with cross-workload and unseen-condition generalization. By unifying observation, cause attribution, and constrained recovery across the sensor-to-silicon stack, the proposed framework establishes a systematic basis for developing Edge-AI nodes that are not only computationally efficient, but also adaptive to heterogeneous operational failures and threats.**

Mostafa Darvishi (S'06–M'18–SM'25) is with Electrical Engineering Department of École de technologie supérieure (ÉTS), Montreal, Canada. (e-mail: darvishi@ieee.org).

## I. INTRODUCTION

The continuing migration of artificial intelligence (AI) from centralized cloud infrastructure toward Internet-of-Things (IoT) endpoints is changing both the computational architecture and the dependability requirements of intelligent cyber-physical systems. Connected sensors increasingly generate high-rate, heterogeneous data streams whose continuous transmission and remote processing impose communication, latency, privacy, and energy costs. Recent studies therefore emphasize local or near-sensor processing as a means of reducing unnecessary data movement while enabling timely decision making at the point of data generation [1], [2]. This transition is reinforced by advances in specialized neural-processing units, heterogeneous system-on-chip (SoC) architectures, quantization, sparsity-aware execution, near-memory processing, and neuromorphic computing, which collectively make increasingly sophisticated inference feasible within resource-constrained platforms [3], [4]. At the same time, the corresponding design objective is shifting away from maximizing arithmetic throughput alone: memory hierarchy, interconnect activity, thermal behavior, communication cost, hardware reliability, and software–hardware interaction increasingly determine the practical efficiency of AI-enabled devices [3].

Moving intelligence to the edge, however, also transfers failure exposure and operational responsibility from comparatively centralized computing infrastructure to heterogeneous devices operating directly within uncertain physical environments. An intelligent IoT node is not merely an AI accelerator connected to a sensor. Its output emerges from a chain involving the physical sensing interface, signal conditioning and preprocessing, learned inference model, processor and accelerator microarchitecture, memory subsystem, power-management circuitry, thermal state, communication interface, operating software, and potentially remote edge or cloud services [5], [6]. Contemporary Edge-AI architectures already reflect this interdependence through application-specific compute fabrics, memory hierarchies, reconfigurable dataflow, dynamic voltage and frequency scaling (DVFS), heterogeneous processing, and local–cloud collaboration [3], [4], [7], [8]. Consequently, degradation originating in one layer can propagate through several others before becoming observable at the application output. This creates a dependability problem that is fundamentally different from optimizing inference accuracy, energy efficiency, hardware security, or fault tolerance independently.

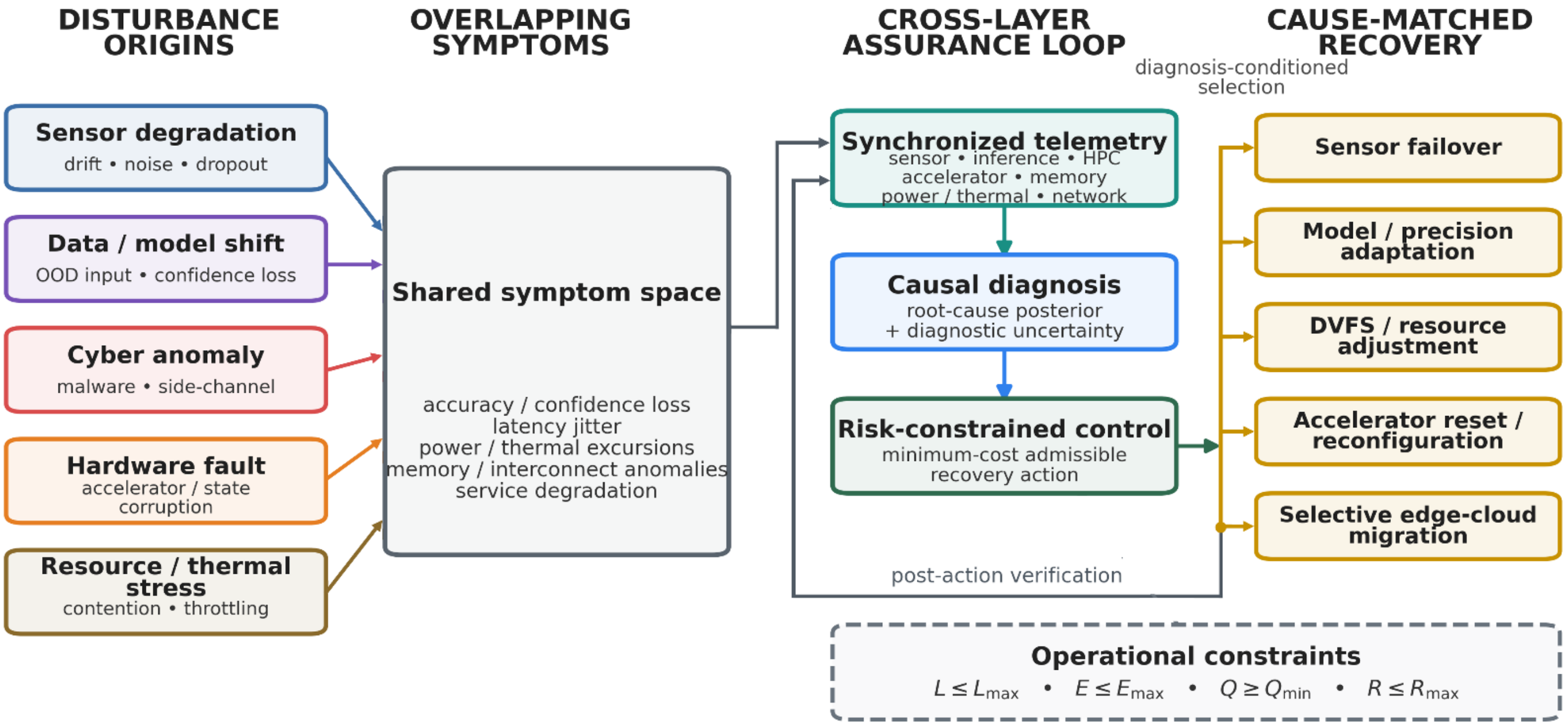

Fig. 1. Cross-layer ambiguity and cause-matched self-healing principle for an intelligent IoT node. Heterogeneous disturbances originating in sensing, data/model behavior, cyber activity, hardware state, and resource/thermal conditions can produce overlapping application-level symptoms. The proposed assurance loop combines synchronized telemetry for cause attribution and uncertainty estimation before selecting a recovery action subject to latency, energy, inference-quality, and residual-risk constraints. The feedback path represents post-action verification rather than an assumption that every corrective action succeeds.

A particularly important consequence is "*cross-layer ambiguity*". A decrease in classification confidence, for example, may be produced by sensor noise or drift, an out-of-distribution input, malicious software activity, corruption of accelerator state, memory interference, thermal throttling, or contention for processor resources. Likewise, increased execution latency or power consumption does not uniquely indicate hardware degradation: it may instead result from a legitimate change in workload, altered communication conditions, a malicious execution pattern, or an adaptive model configuration. These conditions can therefore generate partially overlapping observable signatures while requiring substantially different corrective actions. Sensor degradation may call for sensor failover or fusion; model mismatch may require precision or model adaptation; resource stress may favor DVFS or task migration; suspected malicious execution may require isolation; and accelerator corruption may require reset or reconfiguration. Applying the wrong response can consequently preserve, or even amplify, the original failure despite successful anomaly detection.

Fig. 1 captures this distinction. Five disturbance domains converge toward a partially shared symptom space, whereas recovery becomes meaningful only after evidence from multiple layers has been combined to infer a likely cause. The figure deliberately separates *detection* from *cause attribution* and, in turn, separates cause attribution from *action selection*. This distinction provides the conceptual basis for the proposed work: observing that system behavior is abnormal is not equivalent to determining why it is abnormal, and determining the probable cause is not equivalent to selecting a recovery action that remains admissible under application constraints.

Existing research provides important elements of such an assurance loop, but largely as separate mechanisms. At the sensing end of the system, machine-learning-enhanced nanosensors and emerging neuromorphic structures increasingly merge sensing with local computation, reducing dependence on continuous raw-data transfer and enabling low-latency pattern recognition and anomaly detection [2]. At the processor level, hardware performance counters (HPCs) provide low-level execution signatures that can support lightweight detection of malware and microarchitectural side-channel activity; Wang *et al.* demonstrated that feature-selected HPC measurements and relatively compact classifiers can achieve effective run-time detection while explicitly considering latency, power, and hardware overhead [5]. At the circuit level, AI-assisted calibration, fault localization, and self-healing concepts increasingly combine on-chip monitoring, diagnosis, decision logic, and reconfigurable resources [6]. Meanwhile, low-power Edge-AI research has developed DVFS, power and clock gating, heterogeneous compute resources, near-data processing, quantization, pruning, and specialized accelerators to adapt computation to restricted energy budgets [4], [7], [8].

These advances nevertheless leave a system-level gap. Sensor-oriented methods generally lack visibility into processor or accelerator health; microarchitectural security detectors cannot determine whether degraded application behavior originates upstream at the sensor or algorithmic level; conventional power-management mechanisms are designed primarily around workload and energy state rather than fault provenance; and fault-diagnosis mechanisms do not inherently distinguish malicious behavior from benign resource contention or distribution shift. Furthermore, most detection mechanisms terminate with a classification or alarm rather than determining whether a specific corrective action will improve system state without violating another operational objective. This limitation becomes increasingly consequential as AI hardware moves toward runtime adaptability and heterogeneous task partitioning, trends that require tighter coordination across models, runtime systems, memory organization, communication, power management, and hardware resources [3].

The problem addressed in this work is therefore not conventional anomaly detection. We consider a resource-constrained Edge-AI node in which a degradation may originate from multiple physical, algorithmic, cyber, microarchitectural, or resource-related causes and in which several causes may occur concurrently. Given synchronized but imperfect observations from these layers, the system must estimate the most plausible origin of abnormal behavior, quantify uncertainty in that estimate, and determine whether an available recovery action can restore acceptable operation subject to latency, energy, inference-quality, and security/reliability constraints. A successful solution must additionally avoid excessive monitoring and decision overhead, since the resource limitations that motivate Edge AI also constrain the assurance mechanism itself. This requirement is particularly important for HPC-based monitoring, for which both the number of simultaneously observable events and the computational cost of the classifier can be limited on embedded processors [5].

To address this problem, this paper proposes a *causal cross-layer self-healing Edge-AI framework* spanning the path from physical sensing to silicon execution. The framework constructs a synchronized telemetry state from sensor-health indicators, inference behavior and uncertainty, processor HPCs, accelerator and memory activity, power and thermal observations, and communication conditions. Rather than directly mapping this aggregate state to a generic anomaly label, a temporally informed cause-attribution mechanism evaluates the dependencies among observable variables and candidate disturbance sources. In this work, the term *causal* is used deliberately: cause attribution is anchored to an explicit dependency structure and to controlled interventions introduced during characterization and evaluation, rather than being inferred solely from correlation among naturally occurring telemetry. The resulting cause posterior and its uncertainty are subsequently supplied to a recovery controller that selects an admissible action from sensor, model, compute-resource, reconfiguration, and edge–cloud alternatives.

The recovery problem is treated as constrained decision making rather than as a fixed lookup between detected faults and predetermined responses. A configuration that minimizes instantaneous power, for example, is not useful if it violates application latency or substantially degrades inference quality; similarly, aggressive task migration may restore functionality but incur communication or privacy costs. The controller therefore evaluates candidate actions against explicit operating bounds on latency, energy, inference quality, and residual risk. Recovery is also followed by post-action observation so that ineffective or destabilizing actions can be identified rather than assumed successful. This closed-loop structure distinguishes *self-healing* in the present work from fault notification alone: detection initiates a decision process whose outcome is measured against the restored operating state.

The proposed framework makes five principal contributions. **First**, it introduces a synchronized cross-layer telemetry representation that combines sensing, inference, microarchitectural, accelerator/memory, power/thermal, and communication observations rather than treating these domains as independent monitors. **Second**, it formulates abnormal Edge-AI operation as a cause-attribution problem and develops a temporally informed diagnosis mechanism that estimates both root-cause likelihood and diagnostic uncertainty, including cases involving overlapping or compound perturbations. **Third**, it couples diagnosis to a risk-constrained self-healing controller capable of selecting among sensor failover, model or numerical-precision adaptation, DVFS/resource adjustment, accelerator reset or reconfiguration, and selective edge–cloud migration while respecting application-level operating constraints. **Fourth**, it adopts a local-first execution model in which time-critical diagnosis and recovery do not depend on persistent cloud connectivity, while remote resources remain available for slower adaptation and aggregated learning. **Fifth**, it establishes an FPGA-SoC experimental methodology that synchronizes controlled sensor, cyber, hardware, workload, and resource perturbations with multi-layer telemetry and evaluates not only detection performance but also cause identification, recovery correctness, false recovery, recovery latency, service continuity, communication overhead, energy per correct inference, implementation cost, and generalization to unseen operating conditions.

The emphasis on generalization is intentional. AI-assisted circuit and system design literature continues to identify data efficiency and generalization as persistent challenges [6], while Edge-AI deployments are intrinsically heterogeneous in workload, environment, hardware utilization, and sensor behavior. Accordingly, the evaluation in this work is designed not only around randomized train/test partitions from the same execution traces, but also around cross-workload and unseen-perturbation experiments. Such evaluation is necessary to determine whether cross-layer diagnosis captures relationships that remain useful outside the conditions used to train the model rather than merely memorizing signatures associated with particular injected events.

The scope of the proposed framework is intentionally bounded. It does not assume that every physical or cyber failure can be uniquely diagnosed from finite telemetry, nor does it claim that autonomous recovery can always restore nominal performance. Ambiguous cases are retained through diagnostic uncertainty, and recovery is performed only when an action satisfies the admissibility conditions defined by the controller. Similarly, cloud infrastructure is treated as complementary rather than mandatory: local monitoring and essential recovery remain available during communication loss, whereas computationally expensive adaptation, fleet-level aggregation, or non-time-critical model updates may be executed remotely. This division is consistent with the emerging view of Edge AI as part of a heterogeneous edge–cloud continuum rather than a complete replacement for centralized computation [2], [4].

The remainder of the paper follows the problem decomposition established above. Section II reviews sensor-side intelligence, Edge-AI hardware, microarchitectural security, AI-assisted fault diagnosis, self-healing hardware, and runtime adaptation, and identifies the specific gap between isolated detection mechanisms and cross-layer cause-conditioned recovery. Section III formalizes the system model, telemetry state, disturbance taxonomy, causal dependencies, recovery-action space, and latency–energy–quality–risk constraints. Section IV presents the complete sensors-to-silicon architecture and implementation of the cross-layer telemetry and assurance plane. Section V develops the temporal causal-diagnosis method, including uncertainty and compound-event handling. Section VI formulates the risk-constrained self-healing policy and post-action verification mechanism. Section VII describes the FPGA-SoC implementation, Edge-AI workloads, telemetry instrumentation, perturbation-injection methodology, and measurement procedure. Section VIII evaluates diagnosis, recovery, implementation overhead, energy and latency behavior, cross-workload generalization, and robustness to unseen and concurrent perturbations. Section IX provides ablation and sensitivity studies, discusses failure cases and limitations, and examines the implications for broader Edge-AI deployments. Finally, Section X concludes the paper.

## II. RELATED WORK AND RESEARCH GAP

### *A. From Intelligent Sensing to Resource-Constrained Edge AI*

The progression from cloud-centric sensing toward localized intelligence has been driven largely by the cost of transferring, storing, and repeatedly processing sensor-generated data. Machine-learning-enhanced nanosensors increasingly incorporate feature extraction, adaptive sensing, and predictive processing close to the physical signal source, while emerging in-sensor and neuromorphic approaches seek to reduce the separation between sensing, memory, and computation. Zhang *et al.* [2] identify this migration from cloud-assisted nanosensing toward chip-level edge processing and emphasize that energy-efficient hardware, algorithm optimization, and scalable integration remain key challenges. Their discussion further highlights the potential of near-sensor and neuromorphic processing to support local pattern recognition and anomaly detection while reducing dependence on continuous raw-data transmission. More recent in-sensor-AI research similarly frames sensing and computation as increasingly co-located functions rather than independent stages of an IoT pipeline [9], [10].

At the computational layer, contemporary Edge-AI architectures increasingly combine specialized processing elements, hierarchical local memories, variable-precision arithmetic, event-driven execution, and heterogeneous sensor interfaces. Chung [11] describes this design space using four interacting layers, i.e., compute fabric, memory subsystem, interconnect, and system integration, and explicitly includes reconfigurable precision, local SRAM, sparsity/compression, sensor fusion, DVFS, power gating, and hardware security among the mechanisms needed for practical on-device intelligence. Karpagam *et al.* [7] similarly motivate heterogeneous integration of AI accelerators, reconfigurable logic, conventional processors, near-memory computation, quantization, and sparsity to reduce latency and data movement in smart devices. Low-power VLSI studies further identify clock/power gating, DVFS, heterogeneous cores, and near-data processing as relevant mechanisms for resource-constrained IoT inference [8].

These developments are important to the present work because they provide several of the *actuation mechanisms* required by a self-healing node: precision can be changed, resources can be scaled, components can be duty-cycled, computation can be migrated, and reconfigurable hardware can be modified at runtime. However, these mechanisms are primarily introduced to optimize efficiency or accommodate workload variation. They do not by themselves establish *why* the current operating state has degraded or *which* adaptive action is appropriate.

This distinction becomes increasingly important as AI hardware moves away from a compute-centric view. Recent cross-layer reviews emphasize that practical AI-system behavior depends jointly on memory traffic, interconnect activity, thermal constraints, packaging, software mapping, runtime scheduling, and hardware reliability, rather than solely on arithmetic throughput [3]. Therefore, the same variables that determine efficiency also become potential indicators, or confounders, when diagnosing abnormal behavior. For example, a rise in memory traffic can indicate a legitimate workload change, a poorly mapped model, resource contention, or anomalous execution. Consequently, optimization telemetry cannot automatically be interpreted as fault telemetry.

The first gap emerging from this body of work is therefore clear: *sensor intelligence and Edge-AI adaptation provide rich observability and control mechanisms, but they generally do not close the loop between multi-layer observation, origin-aware diagnosis, and cause-conditioned recovery*.

### *B. Microarchitectural Telemetry for Runtime Security*

At the processor level, HPCs provide a particularly attractive observation channel because they expose low-level processor behavior without requiring intrusive instrumentation of the running application. HPCs record events such as executed instructions, cache misses, and branch activity and have historically been used for performance analysis, energy estimation, debugging, and increasingly security monitoring [5].

Wang *et al.* [5] demonstrated that selected HPC features can support lightweight run-time malware and side-channel-attack detection in resource-constrained edge devices. Their study is significant not only because of its detection performance, but also because it evaluates classifiers in terms of accuracy, F-measure, latency, power, and hardware overhead, thereby demonstrating that a more complex learning algorithm is not automatically preferable for an embedded assurance task. The study also exposes a practical limitation that is directly relevant to the present work: *embedded processors may permit only a small number of HPC registers to be monitored concurrently, while complex classifiers can compete with the protected application for already limited computational resources*.

More recently, APPARENT [12] extends hardware-event monitoring toward unknown-program anomaly detection. Rather than relying only on known malicious signatures, it combines low-level hardware-event measurements with self-supervised representation learning to recognize program behavior that was not present during training. The reported evaluation demonstrates that HPC-based observation remains relevant for contemporary edge processors and can generalize beyond a fixed catalog of attacks.

Nevertheless, both classes of methods answer primarily a security/anomaly question: whether execution appears benign or abnormal. They do not observe the complete sensor-to-application path. In particular, an HPC anomaly detector has no direct basis for deciding whether a simultaneous loss of inference confidence originates from malicious execution, sensor corruption, natural distribution shift, accelerator malfunction, or thermal/resource contention. This is not merely a missing feature channel. It changes the semantics of the inference task: *anomaly detection* asks whether current behavior differs from expected behavior, whereas *cause attribution* asks which mechanism most plausibly generated the observed joint state.

This paper therefore uses microarchitectural telemetry as one evidence domain rather than treating it as an independently sufficient security oracle. That decision also constrains the design developed later in Section III: HPC evidence must be temporally aligned with sensor, inference, accelerator, power, thermal, and communication observations before causal interpretation is attempted.

### C. AI-Assisted Fault Diagnosis and Self-Healing Hardware

A second relevant research direction applies AI directly to hardware diagnosis and post-silicon adaptation. Machine-learning approaches have been explored for parametric fault identification, circuit calibration, defect localization, and reliability analysis. The survey by Guven *et al.* [6] explicitly identifies autonomous self-healing as an emerging extension of these methods. In their formulation, a self-healing architecture requires on-chip health sensing, AI-based diagnosis, a decision mechanism, and reconfigurable elements capable of compensating for or bypassing the faulty region. Importantly, the same study notes that complete autonomous self-healing remains immature and calls for system-level pipelines that are data-efficient, localized, and applicable across increasingly heterogeneous hardware.

Recent work demonstrates substantial progress toward this goal. The Emulated Integrity Replica (EIR) framework [13] targets FPGA-SoC Edge-AI accelerators using two hierarchical digital twins: a coarse behavioral replica for rapid divergence detection and a fine-grained replica for reconstructing a fault-free state. EIR thereby performs actual in-place recovery rather than merely reporting an accelerator error, and it avoids the continuous fabric duplication of conventional DMR/TMR protection. This is a strong and highly relevant development because it demonstrates that autonomous hardware recovery can be practical on the same FPGA-SoC class considered in the present work. However, EIR intentionally addresses a bounded fault model centered on transient accelerator faults and checkpoint-observable execution. Its own stated scope assumes trusted supporting resources and focuses on recoverable PL state rather than simultaneous sensor degradation, cyber anomalies, natural model shift, or system-level thermal/resource stress. The distinction is central: EIR answers the question *“Has this accelerator diverged from its expected computation, and how can the correct state be reconstructed?”* The present paper instead addresses the preceding system-level question *“Which layer is responsible for the observed degradation, and therefore which recovery mechanism should be invoked?”*

A complementary recent study by Udkar *et al.* [14] integrates embedded telemetry, lightweight AI diagnosis, and dynamic FPGA-SoC reconfiguration for a high-power electrical platform. It demonstrates root-cause-oriented diagnosis and automated mitigation on a 50-kW motor-drive testbed, providing important evidence that edge-native diagnosis and hardware/firmware reconfiguration can form a practical closed loop. Its contribution, however, is application-specific: the telemetry and fault semantics arise from the electrical platform. It does not attempt to jointly distinguish the broader sensing, learned-model, cyber, microarchitectural, accelerator, and resource conditions encountered by a heterogeneous Edge-AI node.

Thus, the hardware self-healing literature establishes that autonomous recovery is feasible, but it also highlights why *diagnosis scope matters*. A sophisticated recovery mechanism can still select an inappropriate action if the disturbance has been attributed to the wrong layer.

### D. Cross-Domain and Cause-Oriented Anomaly Diagnosis

Recent work has begun to move beyond single-domain monitoring by combining heterogeneous telemetry. Wang and Chen's CPSNet [15] jointly models time-synchronized cyber and physical streams for Industrial-IoT anomaly diagnosis. Its dual-stream architecture explicitly couples network and physical-sensor information across multiple temporal scales and demonstrates that cross-modal interaction can improve fine-grained diagnosis relative to independently modeled modalities. Importantly, however, the authors explicitly describe the method as exploiting bidirectional contextual dependencies within a fixed observation window rather than claiming strict causal inference. CPSNet therefore provides strong evidence for the value of synchronized cross-domain telemetry, while leaving open the problem of intervention-grounded cause attribution and recovery.

Yu *et al.* [16] approach the problem from the physical-sensor side. Their decentralized edge framework goes beyond correcting anomalous measurements by identifying physical sources of sensor faults such as range, resolution, power-supply, and interference problems, and subsequently initiating sensor recovery. The work also considers combined sensor faults and demonstrates local execution on an IoT edge node. This is conceptually close to the distinction made in Fig. 1: *anomalous data are symptoms, whereas physical sensor failures are candidate causes*. Its diagnostic scope, however, remains intentionally sensor-centric.

A broader causal-inference literature independently reinforces the importance of heterogeneous evidence. Chen *et al.* [17] combine multiple operational data modalities for root-cause analysis in large complex systems and show that single-modal observations can omit information required to identify the initiating event. This supports the decision to formulate diagnosis over multiple telemetry domains. Yet such methods are not designed around the computational, hardware, and energy constraints of an embedded Edge-AI node, nor do they directly associate inferred causes with hardware/software recovery actions.

Security-oriented self-healing is also becoming more sophisticated. TrustEdge [18] integrates on-device TinyML intrusion detection, federated learning, cryptographic protection, and autonomous security responses such as node isolation or model rollback in an IoMT setting. This work is important because it demonstrates that detection and recovery can coexist at the resource-constrained edge. At the same time, its self-healing semantics are cyber-specific: the recovery space is constructed around sensor spoofing, model poisoning, secure learning, and node/model protection rather than hardware faults, thermal contention, benign distribution shift, or accelerator-level degradation.

Collectively, these studies show a clear evolution from isolated anomaly detection toward *domain-aware diagnosis and autonomous remediation*. The remaining difficulty is that each framework defines the system boundary differently. Physical-sensor recovery, cyber–physical anomaly diagnosis, hardware-state recovery, and security self-healing can each be effective within their assumed fault domain while providing insufficient evidence to distinguish disturbances originating outside that domain.

### *E. Runtime Adaptation and Edge–Cloud Coordination*

Runtime adaptation provides the final technical foundation required by the proposed framework. Edge-AI systems increasingly support mixed precision, reconfigurable dataflow, model substitution, heterogeneous scheduling, DVFS, task migration, and selective cloud offloading. These capabilities are usually introduced to maintain inference quality or latency as workload and resource demand change. For example, runtime model-reconfiguration research addresses mismatch between statically deployed models and time-varying edge workloads by switching configurations to preserve inference quality under latency constraints [19]. Reliability-aware edge scheduling similarly considers execution placement and resource control under device failures and heterogeneous computational tiers [20].

The broader AI-on-chip literature points in the same direction. Morsy [3] identifies runtime reconfigurability, mixed-precision execution, flexible scheduling, and heterogeneous task partitioning as increasingly important as workloads diversify. Chung's architecture [4] similarly treats DVFS, power gating, compression, sensor fusion, and cloud–edge cooperation as integral parts of a modern Edge-AI platform rather than external services.

However, *adaptation for efficiency and adaptation for recovery are not equivalent*. A scheduler that sees elevated latency may migrate a workload, but if that latency arose from malicious activity rather than resource pressure, migration alone may propagate the compromised state. Likewise, reducing frequency in response to elevated temperature may be appropriate under benign workload stress but inappropriate if redundant validation must temporarily be activated because of suspected hardware corruption. Recovery therefore requires that adaptation be conditioned on both *cause* and *risk*, rather than only on instantaneous performance metrics.

For this reason, the proposed framework does not introduce new actuation mechanisms merely for their own sake. Sensor failover, model/precision switching, DVFS, accelerator reset/reconfiguration, and edge–cloud migration are treated as an *action set* whose members already have technical precedent. The research contribution lies instead in determining *when* each action is justified, *whether* it remains admissible under the current application constraints, and *whether the post-action state actually improves*.

### *F. Research Gap and Design Requirements*

The comparison in Table I summarizes the distinction quantitatively at the capability level. The table is deliberately not a ranking. Each prior work solves a meaningful subproblem: Wang *et al.* [5] and APPARENT [12] expose microarchitectural security state; CPSNet [15] combines cyber and physical evidence; Yu *et al.* [16] diagnose and recover sensor faults; Udkar *et al.* [14] couple telemetry to hardware reconfiguration; EIR [13] provides fine-grained FPGA-SoC fault recovery; and TrustEdge [18] performs autonomous cyber-oriented remediation. What remains absent from the literature surveyed here is a single experimentally validated Edge-AI assurance loop that simultaneously treats *sensor degradation*, *natural model/data shift*, *cyber anomalies*, *hardware faults*, and *resource/thermal stress* as *competing* or *concurrent* causes, while combining physical, algorithmic, microarchitectural, hardware, power/thermal, and communication telemetry before selecting among heterogeneous recovery actions.

This synthesis leads to six requirements that determine the remainder of the paper. **R1 [Cross-layer observability]:** telemetry must span the physical sensor, inference model, processor microarchitecture, accelerator/memory subsystem, power/thermal state, and communication interface. **R2**

Table I. CAPABILITY COMPARISON OF PRIOR WORKS AND THE PROPOSED CROSS-LAYER SELF-HEALING FRAMEWORK

| Study | Sensor health | Model / data shift | Cyber state | HW fault | Resource / thermal | µArch. HPCs | Cross-layer fusion | Cause attribution | Uncertainty | Autonomous recovery | HW / edge validation |
|---|---|---|---|---|---|---|---|---|---|---|---|
| [5] | — | — | ✓ | — | — | ✓ | — | — | — | — | ✓ |
| [12] | — | — | ✓ | — | — | ✓ | — | — | — | — | ✓ |
| [15] | ✓ | — | ✓ | — | — | — | ✓ | ◐ | — | — | ✓ |
| [16] | ✓ | — | — | — | ◐ | — | ◐ | ✓ | — | ✓ | ✓ |
| [14] | ✓ | — | — | ✓ | ◐ | — | ◐ | ✓ | — | ✓ | ✓ |
| [13] | — | — | — | ✓ | — | — | ◐ | ✓ | — | ✓ | ✓ |
| [18] | ✓ | ◐ | ✓ | — | — | — | ◐ | ◐ | — | ✓ | ◐ |
| This Work | ✓ | ✓ | ✓ | ✓ | ✓ | ✓ | ✓ | ✓ | ✓ | ✓ | ✓ |

✓ = Explicitly addressed; ◐ = Partially addressed or restricted to a single domain/assumption; — = Not a stated capability.
Capability-based comparison only. “Cause attribution” denotes diagnosis beyond generic anomaly detection; the proposed row states design to be experimentally verified in Sections VII-IX.

**[Cause-oriented diagnosis]:** correlated abnormality alone is insufficient; diagnosis must explicitly represent dependencies among candidate causes and observed effects and must be evaluated using controlled interventions. **R3 [Uncertainty awareness]:** when available evidence does not uniquely support one cause, the system must preserve uncertainty rather than force a high-confidence label. **R4 [Cause-conditioned recovery]:** the diagnosed cause must be mapped to heterogeneous candidate actions and filtered by explicit latency, energy, inference-quality, and residual-risk constraints. **R5 [Local-first operation]:** time-critical diagnosis and recovery must remain functional without continuous cloud access, while slower learning and aggregation may exploit remote resources. **R6 [Experimental separability]:** the evaluation must contain controlled single and compound perturbations, unseen operating conditions, and multiple AI workloads so that improvements can be attributed to cross-layer reasoning rather than to memorization of one benchmark or disturbance signature.

These requirements also fix the technical flow of the remaining sections. Section III converts R1–R4 into a formal system model by defining the synchronized telemetry vector, disturbance variables, causal dependency structure, diagnostic uncertainty, recovery action space, and admissibility constraints. Section IV then translates that model into the sensors-to-silicon hardware/software architecture and specifies where each telemetry source and actuator resides. Section V develops the temporal causal-diagnosis engine and establishes how controlled interventions and uncertainty are incorporated so that the term *causal* is empirically defensible. Section VI formulates the risk-constrained recovery policy and post-action verification loop. Section VII implements the resulting architecture on the FPGA-SoC prototype and defines the workloads, instrumentation, injection procedures, and synchronization methodology required to satisfy R6. Section VIII evaluates diagnosis and recovery under single, compound, and unseen conditions, while Section IX isolates the contribution of individual telemetry domains, causal structure, uncertainty handling, and recovery constraints through ablation and sensitivity analysis.

## III. SYSTEM MODEL AND PROBLEM FORMULATION

Section II showed that existing Edge-AI systems expose increasingly rich sensing, microarchitectural, accelerator, power, and communication observability, while recent work on root-cause diagnosis demonstrates the value of combining heterogeneous information rather than relying on a single modality [3]. Recent causal-diagnosis frameworks likewise distinguish causal-structure degradation or multimodal root-cause reasoning from purely correlation-driven anomaly detection [17], [21]. Building on these observations, this section formulates the proposed sensors-to-silicon assurance problem as a partially observed, temporally evolving causal system in which heterogeneous disturbances affect multiple telemetry domains and a recovery controller must act under explicit operational constraints.

### A. System Boundary and Temporal Decision Cycle

Consider an intelligent IoT node $\mathcal{N}$ comprising a physical sensing interface, preprocessing pipeline, general-purpose processor, AI accelerator, local memory subsystem, power-management and thermal-monitoring resources, and a communication interface to an optional edge server or cloud. The remote resource is not assumed to be continuously available; time-critical diagnosis and recovery are therefore executed locally, consistent with the local-first requirement established in Section II.

The node operates over discrete decision intervals of duration $\Delta$. The index $t$ refers to the $t$-th synchronized decision interval rather than to an individual hardware clock cycle. This distinction is necessary because the relevant telemetry sources naturally operate at different rates: hardware counters may be sampled every few milliseconds, sensor-quality statistics may be associated with frames or acquisition windows, thermal variables evolve more slowly, and network metrics may be event driven.

Let $\mathbf{k}_t$ denote the active system configuration during interval $t$. It includes variables that can alter the behavior of the node, such as selected sensor path, inference model, numerical precision, processor/accelerator frequency, accelerator configuration, and current local/remote execution placement. Explicitly representing $\mathbf{k}_t$ is important because the same telemetry value may have different meaning under different configurations. For example, a cache-miss rate observed while executing an INT8 model cannot necessarily be interpreted using a baseline collected under an INT4 configuration. A complete operating cycle is therefore

$$\mathbf{k}_t \rightarrow \text{sense} \rightarrow \text{infer} \rightarrow \text{observe} \rightarrow \text{diagnose} \rightarrow \text{act} \rightarrow \mathbf{k}_{t+1},$$

with post-action observations determining whether the selected intervention actually restored an admissible operating state.

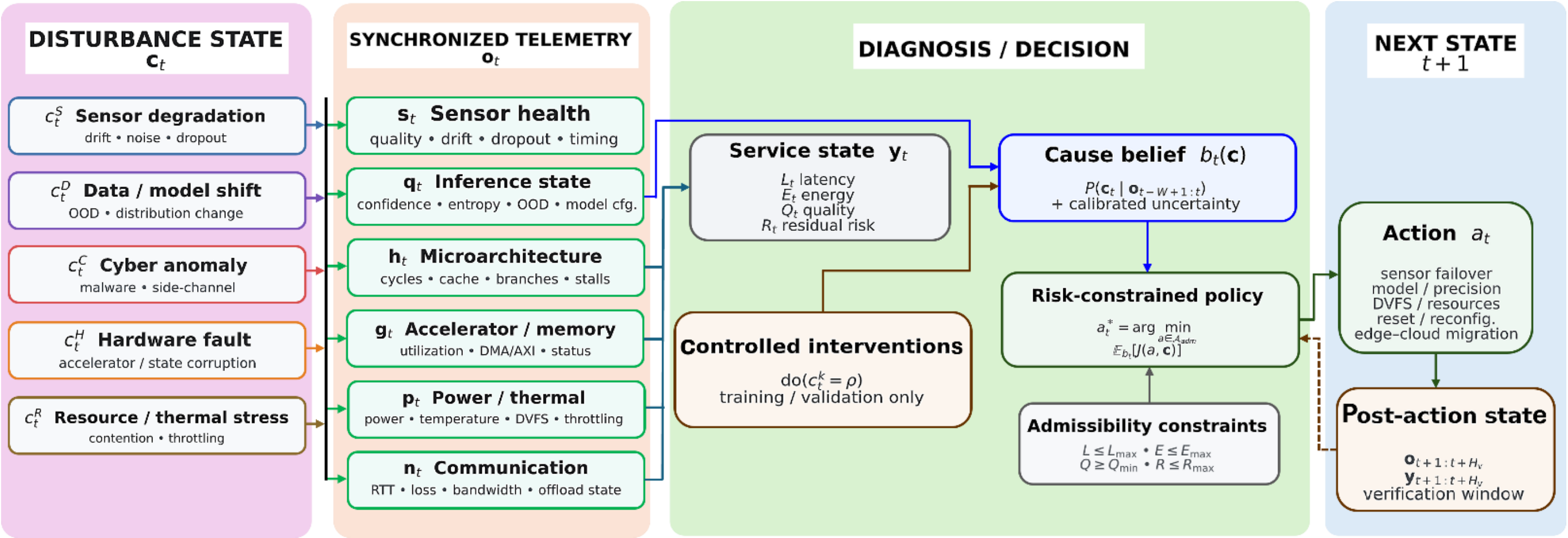

Fig. 2. Temporal causal system model and diagnosis–decision interface of the proposed framework. The multi-hot disturbance state $\boldsymbol{c}_t$ represents sensor degradation, data/model shift, cyber anomalies, hardware faults, and resource/thermal stress. Heterogeneous measurements are synchronized into the cross-layer observation vector $\boldsymbol{o}_t$. Controlled interventions are used during characterization and validation to establish and test cause–effect relationships; deployment-time diagnosis relies on passive telemetry. The inferred cause belief and associated uncertainty condition a risk-constrained recovery policy, whose action modifies the subsequent system state. Post-action observations close the verification loop. The disturbance-to-telemetry collection paths indicate candidate influence relationships; they do not imply that every disturbance directly affects every telemetry variable.

### *B. Synchronized Cross-Layer Observation State*

Following requirement R1 of Section II, the framework observes six telemetry domains. At each interval $t$, the synchronized system observation is defined as

$$\boldsymbol{o}_t = [\boldsymbol{s}_t^T, \boldsymbol{q}_t^T, \boldsymbol{h}_t^T, \boldsymbol{g}_t^T, \boldsymbol{p}_t^T, \boldsymbol{n}_t^T]^T \quad (1)$$

where $\mathbf{s}_t$ contains sensor-health information, $\mathbf{q}_t$ describes the AI-inference state, $\mathbf{h}_t$ contains processor microarchitectural observations, $\mathbf{g}_t$ represents accelerator and memory activity, $\mathbf{p}_t$ contains power/resource/thermal observations, and $\mathbf{n}_t$ represents communication state.

The individual domains are deliberately heterogeneous. Sensor observations can include signal-quality indicators, drift statistics, dropout rate, acquisition timing, or modality-specific health measures. Inference observations include confidence, predictive entropy, out-of-distribution scores, execution latency, active model identity, and numerical precision. Microarchitectural telemetry includes selected HPC events such as cycles, instructions, cache behavior, branches, and stalls. HPC availability is itself resource constrained, modern processors may expose only a limited subset simultaneously, which motivates feature selection rather than indiscriminate monitoring [5]. Accelerator/memory telemetry may include utilization, direct memory access (DMA) or Advanced eXtensible Interface (AXI) activity, buffer or memory traffic, completion/error status, and accelerator execution latency. Power/thermal telemetry includes voltage/current-derived power where available, temperature, operating frequency, DVFS state, and throttling indicators. Finally, communication telemetry may include round-trip delay, packet loss, effective bandwidth, and offloading state. Because these signals are asynchronous, let the raw measurements from telemetry domain $m$ be

$$X_t^{(m)} = \{\mathbf{x}^{(m)}(\tau_i) : \tau_i \in [t\Delta, (t+1)\Delta)\}.$$

A synchronization operator $\Phi_m(\cdot)$ maps all measurements occurring within the decision interval to a fixed-dimensional representation,

$$\tilde{\boldsymbol{x}}_t^{(m)} = \Phi_m\left(X_t^{(m)}\right) \quad (2)$$

where $\Phi_m$ may contain statistics appropriate to the modality, such as count, mean, maximum, variance, trend, or event rate. Section IV will specify the exact hardware/software collection procedure and Section VII will define the final sampling interval and aggregation functions. Missing measurements must not silently become ordinary numerical values. A telemetry-availability mask $\mathbf{m}_t \in \{0,1\}^{d_o}$ is therefore retained alongside $\mathbf{o}_t$. Consequently, the diagnostic input is the pair

$$\bar{\boldsymbol{o}}_t = (\boldsymbol{o}_t, \boldsymbol{m}_t) \quad (3)$$

so that unavailable telemetry is distinguishable from an actual zero-valued observation. This treatment will also allow Section VIII to evaluate a practical question that is often omitted in multimodal diagnosis: *how much diagnostic capability remains when one telemetry source disappears or* *becomes unreliable*?

### C. Disturbance State and Compound Events

The system is subject to the five disturbance domains introduced in Fig. 1. Rather than modeling them as mutually exclusive classes, we define the latent disturbance state as the multi-hot vector

$$\boldsymbol{c}_t = [c_t^S, c_t^D, c_t^C, c_t^H, c_t^R]^T \in \{0,1\}^5, \tag{4}$$

where $c_t^S$ denotes sensor degradation, $c_t^D$ data/model distribution shift, $c_t^C$ cyber or malicious-execution activity, $c_t^H$ hardware/accelerator faults, and $c_t^R$ resource or operating stress. Normal operation is represented by $\mathbf{c}_t = \mathbf{0}$. Crucially, $\|\mathbf{c}_t\|_0 > 1$ is permitted, meaning that the formulation inherently supports compound events rather than forcing every observation window into a single-cause category. A corresponding severity vector

$$\boldsymbol{\rho}_t = [\rho_t^S, \rho_t^D, \rho_t^C, \rho_t^H, \rho_t^R]^T, \qquad \rho_t^k \in [0,1], \tag{5}$$

is introduced for experimental characterization. The binary vector identifies *which* disturbance mechanisms are active, whereas $\boldsymbol{\rho}_t$ represents their controlled intensity. This distinction is important. A small sensor-offset perturbation and complete sensor dropout belong to the same causal family but may produce fundamentally different telemetry responses and recovery requirements.

Communication state is retained as a dedicated observation domain $\mathbf{n}_t$ rather than introducing a sixth primary cause in Fig. 1. When communication degradation itself is experimentally imposed, for example to alter the feasibility of edge–cloud migration, it is modeled as part of the broader operating/resource-stress condition $c_t^R$, while its physical manifestation remains explicitly observable through $\mathbf{n}_t$.

### D. Application-Level Outcome State

The controller does not attempt to maintain every telemetry variable at a fixed nominal value. The actual objective is to preserve application service. We therefore define

$$\boldsymbol{y}_t = [L_t \quad E_t \quad Q_t \quad R_t]^T, \tag{6}$$

where $L_t$ is end-to-end inference latency, $E_t$ is energy consumed over the decision interval, $Q_t$ represents application-level inference quality, and $R_t$ is residual operational risk. The interpretation of $Q_t$ requires care. Ground-truth accuracy is generally unavailable during normal deployment; consequently, runtime quality must be estimated using a calibrated surrogate based on quantities such as confidence, predictive entropy, agreement, or domain-specific consistency. During the labeled experimental campaign of Section VII, however, $Q_t$ can additionally be computed from the actual ground truth. The manuscript will therefore distinguish **online quality estimate** $\hat{Q}_t$ from **experimentally measured quality** $Q_t$ whenever necessary. Similarly, $R_t$ does not represent an abstract subjective safety score. It will be constructed in Section VI from the probability and severity of unresolved disturbances after considering the proposed action and diagnostic uncertainty.

### E. Temporal Structural Causal Model

The central modeling choice of this paper is to avoid treating the concatenated telemetry vector merely as input to an unconstrained classifier. Let $\mathcal{V}_t = \{\mathbf{c}_t, \mathbf{o}_t, \mathbf{k}_t, \mathbf{y}_t, \mathbf{a}_t\}$ denote the variables required to describe one decision interval. Their evolution is represented by a temporal structural causal model (SCM) as

$$v_{j,t} = f_j\big(Pa_G(v_{j,t}), \mathcal{V}_{t-1:t-p}, u_{j,t}\big), \tag{7}$$

where $v_{j,t} \in \mathcal{V}_t$, $\mathrm{Pa}_{\mathcal{G}}(v_{j,t})$ denotes the contemporaneous parents of variable $v_{j,t}$ in causal graph $G$, $p$ is the maximum temporal lag considered, and $u_{j,t}$ represents unmodeled exogenous influences.

The graph is **not learned blindly from observational correlations**. Instead, its allowable structure is constrained first by system architecture and temporal precedence. For example, a sensor-health change may influence inference confidence, whereas a change in inference confidence cannot retroactively cause the sensor drift that preceded it. Hardware-counter activity may influence power state and observed latency, while DVFS configuration can influence both counter rates and execution time. This architecture-informed prior is then refined and validated using the controlled interventions defined below.

This combination of structural prior and temporal evidence is consistent with recent causal-diagnosis research [21], which increasingly uses directed and lag-aware dependency structure to separate root-cause mechanisms from symptom correlations. Multimodal root-cause work likewise shows that combining heterogeneous system variables can reveal causal information unavailable to single-modal analysis [17]. The physical system evolution can be summarized more compactly as observation evolution (8), application/output evolution (9), and configuration evolution (10).

$$\boldsymbol{o}_t = F_o\big(\boldsymbol{c}_t, \boldsymbol{k}_t, \boldsymbol{o}_{t-1:t-p}, \boldsymbol{u}_t^o\big), \tag{8}$$

$$\boldsymbol{y}_t = F_y\big(\boldsymbol{o}_t, \boldsymbol{c}_t, \boldsymbol{k}_t, \boldsymbol{u}_t^y\big), \tag{9}$$

$$\boldsymbol{k}_{t+1} = F_k(\boldsymbol{k}_t, \boldsymbol{a}_t, \boldsymbol{u}_t^k). \tag{10}$$

Equations (8)–(10) make an important conceptual separation: *disturbances affect observations and application behavior, whereas the recovery action primarily changes the subsequent configuration through which the next system state evolves*.

### F. Controlled Interventions and Causal Validation

A causal interpretation requires stronger evidence than temporal correlation. For this reason, the experimental methodology will deliberately include controlled interventions during characterization and evaluation. For disturbance family $k$, a controlled intervention is denoted

$$do(c_t^k = \rho), \tag{11}$$

meaning that the corresponding mechanism is externally imposed at a controlled severity $\rho$ while the remaining experimental conditions are held or randomized according to the protocol which is explained later in Section VII. The average causal response of an observed or service variable $v$ to disturbance $k$ can then be evaluated as

$$\Delta_{k \to v}(\rho) = \mathbb{E}[v \mid do(c^k = \rho)] - \mathbb{E}[v \mid do(c^k = 0)]. \tag{12}$$

Equation (12) will not be used to claim that all disturbance mechanisms can be perfectly isolated in real deployments. Its purpose is to establish which telemetry variables react reproducibly when the experimenter deliberately perturbs a known mechanism. For compound disturbances $k$ and $\ell$, we additionally define the interaction response

$$\begin{aligned} \mathcal{I}_{k,\ell \to v} &= \mathbb{E}\left[v \mid do\left(c^k = \rho_k, c^\ell = \rho_\ell\right)\right] \\ &\quad - \mathbb{E}[v \mid do(c^k = \rho_k)] - \mathbb{E}\left[v \mid do\left(c^\ell = \rho_\ell\right)\right] \\ &\quad + \mathbb{E}[v \mid \boldsymbol{c} = \boldsymbol{0}]. \end{aligned} \tag{13}$$

A nonzero $\mathcal{I}_{k,\ell \to v}$ indicates that the joint response is not explained by a simple additive superposition of the two isolated disturbances. This quantity will be particularly useful when interpreting the compound-event experiments in Section VIII.

The causal claims of the paper will be restricted by explicit assumptions. Temporal precedence must be resolved at the selected synchronization interval; injected perturbations must target the stated mechanism with independently characterized side effects; configuration changes must be recorded rather than treated as latent variation; and every targeted disturbance must have at least one observable path to the monitored system boundary. When these conditions are not satisfied, the system will be allowed to return an ambiguous diagnosis rather than force a unique cause. This last condition is essential. *Causal identifiability is treated as an empirical property to be tested, not an assumption guaranteed by the model architecture*.

### *G. Cause Belief and Diagnostic Uncertainty*

During deployment, the controlled intervention label is unavailable. Diagnosis therefore operates on a temporal observation window

$$\boldsymbol{O}_t^{(W)} = \{\bar{\boldsymbol{o}}_{t-W+1}, \dots, \bar{\boldsymbol{o}}_t\}, \tag{14}$$

where $W$ is the diagnosis-window length. Let $\mathcal{C} \subseteq \{0,1\}^5$ denote the set of cause configurations considered during training, including normal operation and supported compound events. The diagnostic engine estimates the cause belief

$$b_t(\boldsymbol{c}) = P_\theta\left(\boldsymbol{c}_t = \boldsymbol{c} \mid \boldsymbol{O}_t^{(W)}, \boldsymbol{k}_t\right), \qquad \boldsymbol{c} \in \mathcal{C}. \tag{15}$$

The output of the diagnostic stage is therefore not only a hard label. A normalized belief entropy can be used as one complementary ambiguity measure,

$$U_t^H = -\frac{\sum_{c \in \mathcal{C}} b_t(\boldsymbol{c}) \log b_t(\boldsymbol{c})}{\log|\mathcal{C}|}, \tag{16}$$

where $U_t^H \to 0$ corresponds to a concentrated posterior and $U_t^H \to 1$ indicates substantial ambiguity. Entropy alone, however, does not guarantee calibration. Section V will therefore construct a calibrated prediction set

$$\Gamma_t(\alpha) \subseteq \mathcal{C}, \tag{17}$$

for target miscoverage level $\alpha$. The goal is that a confidently identifiable event produces a small set, ideally a singleton, whereas weak or conflicting evidence produces a larger set and can prevent an aggressive recovery decision. Conformal prediction has recently been explored for uncertainty-aware fault diagnosis precisely because conventional classifiers can remain overconfident under ambiguous operating conditions [22]. We will not claim distribution-free coverage under arbitrary nonstationary deployment conditions. Any conformal-style guarantee depends on its calibration assumptions; under distribution shift, coverage will therefore be measured experimentally rather than assumed.

### *H. Recovery Action Space*

The self-healing controller operates over the cause-matched action space introduced in Fig. 1,

$$\mathcal{A} = \{a^0, a^S, a^M, a^V, a^H, a^E\}, \tag{18}$$

where $a^0$ denotes hold/no intervention, $a^S$ sensor failover or sensor-selection adaptation, $a^M$ model or numerical-precision adaptation, $a^V$ DVFS/resource adjustment, $a^H$ accelerator reset or hardware reconfiguration, and $a^E$ selective edge–cloud migration. The no-intervention action $a^0$ is included mathematically even though it is omitted from the recovery column of Fig. 1 because it represents the controller's decision not to alter an otherwise admissible system.

The action set is deliberately heterogeneous because the central hypothesis of this paper is that *different causes require different recovery mechanisms*. A sensor-originated fault should not automatically trigger FPGA reconfiguration, and a hardware-state fault should not automatically trigger cloud migration.

### *I. Admissibility and Risk-Constrained Action Selection*

Not every technically available action is acceptable at every instant. For example, cloud migration may violate latency or communication requirements, higher precision may violate the available energy budget, and a reset may interrupt service beyond an application-specific limit. For each candidate action $a$, let the controller predict the next-window service state

$$\hat{\mathbf{y}}_{t+1}(a) = \left[\hat{L}_{t+1}(a), \hat{E}_{t+1}(a), \hat{Q}_{t+1}(a), \hat{R}_{t+1}(a)\right]^{\mathrm{T}}.$$

The admissible action set is then

$$\mathcal{A}_{adm}(t) = \left\{ a \in \mathcal{A}: \begin{array}{c} Pr[L_{t+1}(a) \le L_{max}] \ge 1 - \epsilon_L, \\ Pr[E_{t+1}(a) \le E_{max}] \ge 1 - \epsilon_E, \\ Pr[Q_{t+1}(a) \le Q_{max}] \ge 1 - \epsilon_Q, \\ Pr[R_{t+1}(a) \le R_{max}] \ge 1 - \epsilon_R \end{array} \right\}. \tag{19}$$

The $\epsilon$ parameters determine the tolerated probability of constraint violation. Among admissible actions, the controller minimizes the posterior-expected recovery cost

$$a_t^* = \underset{a \in \mathcal{A}_{adm}(t)}{arg\,min}\ \mathbb{E}_{c \sim b_t}[J(a, \boldsymbol{c})]. \tag{20}$$

A normalized cost function may be written as

$$J = w_L \tilde{L} + w_E \tilde{E} + w_B \tilde{B} + w_Q\left(1 - \tilde{Q}\right) + w_R \tilde{R} + w_S C_{sw}, \tag{21}$$

where $L$, $E$, $B$, $Q$, and $R$ denote normalized latency, energy, communication cost, inference quality, and residual risk, respectively, while $C_{\mathrm{sw}}$ represents switching or reconfiguration cost. The weights in (21) are not intended to conceal an arbitrary scalar ranking. Section VI will specify their normalization and sensitivity analysis, and Section IX will examine whether conclusions remain stable when the weights change.

Importantly, uncertainty affects the controller through the entire posterior $b_t$, rather than only through the maximum-probability diagnosis. If $\Gamma_t(\alpha)$ remains broad or no action satisfies (19), the controller must retain the current state, enter an application-defined safe configuration, or escalate rather than force a potentially harmful recovery.

### *J. Post-Action Verification*

Self-healing is not complete when an action has merely been issued. After executing $a_t^*$, the system observes a verification window of length $H_v$,

$$\mathcal{W}_t^v = \{\boldsymbol{o}_{t+1:t+H_v}, \boldsymbol{y}_{t+1:t+H_v}\}. \tag{22}$$

Recovery is accepted only if the resulting operating state satisfies the service constraints over the verification interval and the diagnosed disturbance evidence is sufficiently reduced. A compact verification indicator is

$$\chi_t = \mathbb{I}\left[\overline{L}_t^v \le L_{max}, \overline{E}_t^v \le E_{max}, \overline{Q}_t^v \ge Q_{min}, \overline{R}_t^v \le R_{max}\right], \tag{23}$$

where the barred quantities denote the predefined verification-window statistic rather than necessarily an arithmetic mean; for latency, for example, Section VI may use a tail statistic such as the 95th or 99th percentile. If $\chi_t = 0$, the action is considered unsuccessful or insufficient, and the recovery logic must either roll back, select the next admissible action, or escalate according to the policy developed in Section VI. This post-action validation prevents the circular assumption that a selected action must have corrected the diagnosed problem.

### *K. Formal Problem Statement*

The overall objective can now be stated precisely. Given a synchronized temporal observation window $\mathbf{O}_t^{(W)}$ from a resource-constrained Edge-AI node, determine a calibrated belief over the potentially compound disturbance state $\mathbf{c}_t$, and select an action $a_t$ that minimizes expected recovery cost while preserving application-level latency, energy, inference-quality, and residual-risk constraints. The learned causal structure must be supported by controlled perturbation experiments rather than observational association alone, and the selected recovery must be verified from the subsequent measured system state.

This formulation creates a direct dependency between the next sections. Section IV will realize every telemetry variable and actuator assumed in (1)–(3) and (18). Section V instantiates the causal model in (7)–(17) and demonstrates that cause attribution survives competing and compound symptoms. Section VI will implement (19)–(23) without introducing an optimization objective inconsistent with the system model. Section VII creates controlled interventions sufficiently clean to support (11)–(13), while Sections VIII–IX will test not only diagnosis accuracy but also uncertainty calibration, intervention consistency, recovery success, and sensitivity to missing telemetry and unseen conditions.

## IV. SENSORS-TO-SILICON ARCHITECTURE AND CROSS-LAYER TELEMETRY PLANE

The system model of Section III requires more than the simultaneous availability of heterogeneous measurements. The observations entering $\mathbf{o}_t$ must originate from identifiable physical or computational subsystems, share a consistent time reference, retain their validity and configuration context, and remain separable until their relationships are evaluated by the causal model. Likewise, each recovery action in $\mathcal{A}$ must correspond to a realizable control mechanism rather than an abstract decision label. This section maps those requirements onto the heterogeneous FPGA-SoC architecture shown in Fig. 3.

The reference prototype is organized around a ZCU104-class Zynq UltraScale+ MPSoC platform containing application processors, real-time processors, programmable logic, shared memory, and PS–PL interconnect resources within the same device [23], [24]. This heterogeneous organization is particularly useful for the proposed framework because sensing, software execution, AI acceleration, memory traffic, low-level hardware instrumentation, and recovery control can be observed within a common system boundary while retaining physically distinct acquisition paths.

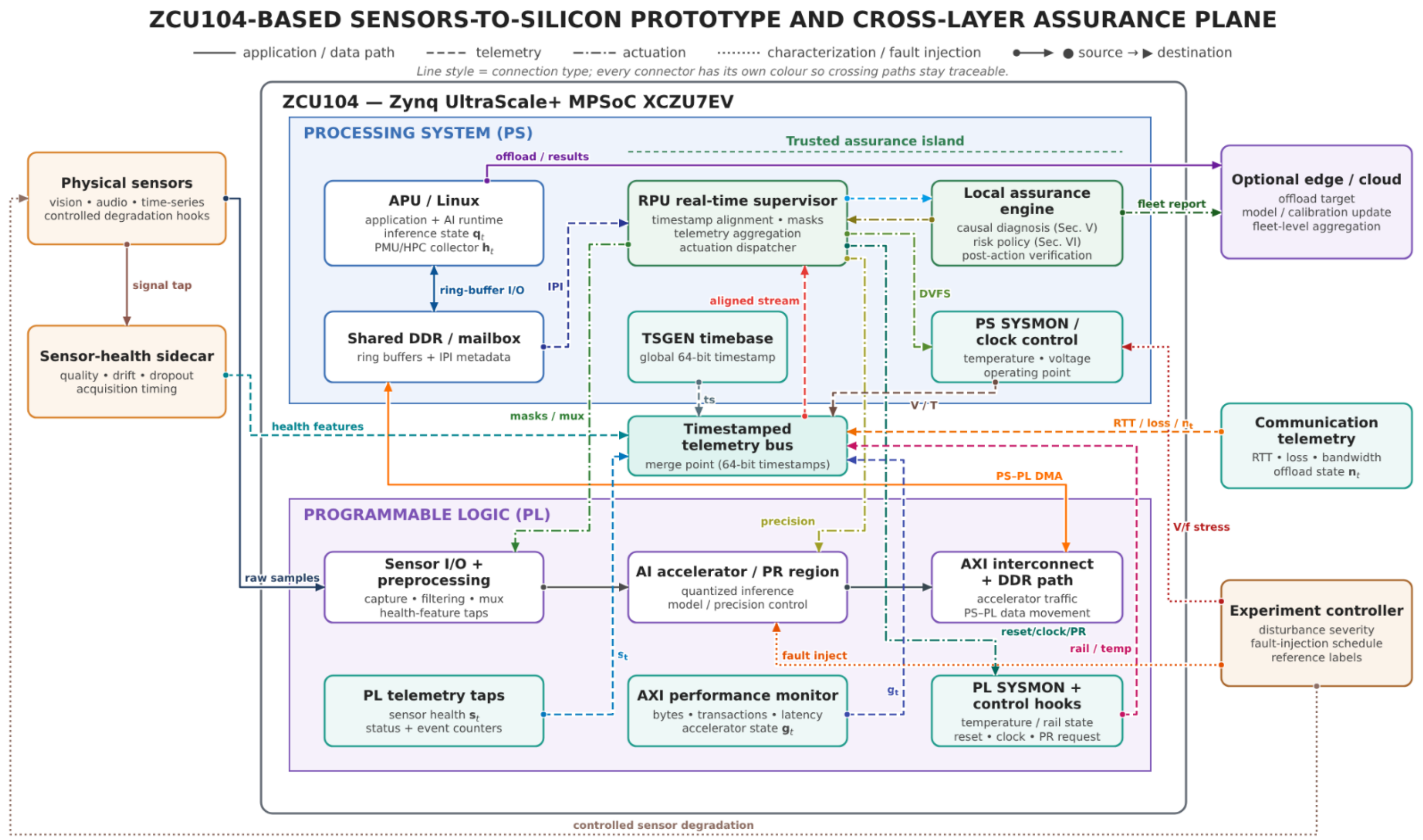

Fig. 3. ZCU104-based sensors-to-silicon prototype and cross-layer assurance architecture. Connection semantics are encoded by line style rather than color: solid lines denote application/data transport, dashed lines denote telemetry, dash-dot lines denote recovery actuation, and dotted lines denote controlled characterization or fault injection. Individual routes use distinct colors only to preserve traceability where several connections coexist; a filled circle denotes the source and the arrowhead denotes the destination. Raw sensor data follow the application path through PL preprocessing, the AI accelerator, the AXI/DDR subsystem, PS–PL DMA, shared memory, and the APU, while a separate signal tap feeds the sensor-health sidecar. Sensor-health features, PL sensor/status counters, accelerator/memory measurements, PS/PL power–thermal measurements, communication state, and the common timestamp are merged through the timestamped telemetry bus. APU inference and PMU/HPC information are exchanged through the shared DDR/mailbox and notified to the RPU through the IPI path. The RPU forms the aligned observation stream supplied to the local assurance engine and dispatches authorized sensor-mux, precision, DVFS, and reset/clock/partial-reconfiguration actions. Optional application offloading and fleet-level reporting are shown separately from the local recovery loop. The experiment controller is isolated from deployment operation and provides controlled sensor degradation, accelerator fault injection, operating-point stress, and rail/thermal perturbations for causal characterization and validation.

A central design principle is to keep **application execution, observation, decision, actuation, and experimental intervention logically separate**. The architecture therefore contains four connection classes: an application/data plane, a telemetry plane, a recovery-actuation plane, and a characterization/fault-injection plane. Their roles and directions are explicitly encoded in Fig. 3.

### *A. Architectural Organization and Connection Semantics*

Fig. 3 presents the complete sensors-to-silicon architecture used throughout the remainder of the paper. Its graphical conventions are part of the system definition and are therefore stated explicitly. *Solid connections* represent application or payload-data movement. *Dashed connections* represent telemetry transport toward the assurance subsystem. *Dash-dot connections* represent recovery or control actions initiated by the assurance path. *Dotted connections* are reserved for controlled characterization and fault-injection operations and are disabled during ordinary deployment.

Color does **not** define the semantic class of a connection. Instead, each individual route is assigned a distinct color so that simultaneously visible or crossing paths remain traceable. Connectivity is indicated only by the endpoints: a filled circle identifies the source, whereas the arrowhead identifies the destination. Consequently, two geometrically crossing lines are not considered connected unless the route explicitly terminates at a shared block or merge point.

This convention is especially important around the timestamped telemetry bus, where several independently acquired measurements converge. The individual telemetry channels remain logically distinct until reaching this merge point; their visual proximity must not be interpreted as premature fusion.

The architecture itself contains three principal computational domains. The **programmable logic (PL)** performs sensor acquisition, deterministic preprocessing, AI acceleration, low-level event monitoring, AXI observation, and programmable-logic control. The **processing system (PS)** hosts application execution, shared-memory communication, deterministic supervision, timestamp synchronization, and the local assurance functions. Finally, an **optional edge/cloud endpoint** supports workload offloading, model/calibration updates, and fleet-level aggregation without becoming a dependency of local fault containment.

### *B. Application and Inference Data Plane*

The application path begins at the physical sensing interface. Three workload families, i.e., **vision**, **audio**, and **time-series sensing**, are considered in the architecture so that the eventual evaluation is not tied to a single sensing modality. The physical interface exposes two conceptually different outputs.

The first is the **raw-sample path**, shown as a solid connection from the physical sensor block to the PL **Sensor I/O + preprocessing** subsystem. This path carries the application payload required for inference. The preprocessing block performs deterministic operations required by the selected workload, such as acquisition, framing, filtering, normalization, buffering, and sensor selection.

The second is the solid **signal tap** feeding the external **Sensor-health sidecar**. The sidecar does not replace the normal acquisition path and does not determine the inference output. Instead, it independently characterizes physical-signal integrity through quantities such as signal quality, drift, dropout, and acquisition timing.

This distinction is important because an anomalous application result does not uniquely identify the physical sensor as its cause. A sensor can remain electrically and temporally healthy while the input distribution becomes unfamiliar to the model; conversely, a physical sensor can degrade while downstream computation remains internally correct. The architecture therefore observes physical-signal quality separately from application inference. After deterministic preprocessing, application samples are transferred through the solid path

*Sensor I/O → AI accelerator → AXI/DDR subsystem.*

The **AI accelerator / PR region** executes the quantized inference workload and exposes model/precision controls that will later support $a^M$. The accelerator is represented as a reconfigurable region rather than as a permanently fixed network because the experimental campaign will compare workloads and numerical configurations with different compute-to-memory characteristics.

Its output traffic is delivered to the **AXI interconnect + DDR path**, whose activity is independently monitored as will be described in Section IV-E. This separation follows the broader observation that practical Edge-AI efficiency depends substantially on memory and communication behavior, rather than arithmetic throughput alone [3].

Fig. 3 additionally shows the solid **PS–PL DMA** path between shared PS memory and the PL AXI subsystem. The illustrated connection represents the PS-to-PL buffer-transfer direction used by the configured application flow. The exact DMA organization, buffer sizes, and any workload-dependent reverse transfer are implementation parameters to be reported in Section VII rather than inferred from the architectural

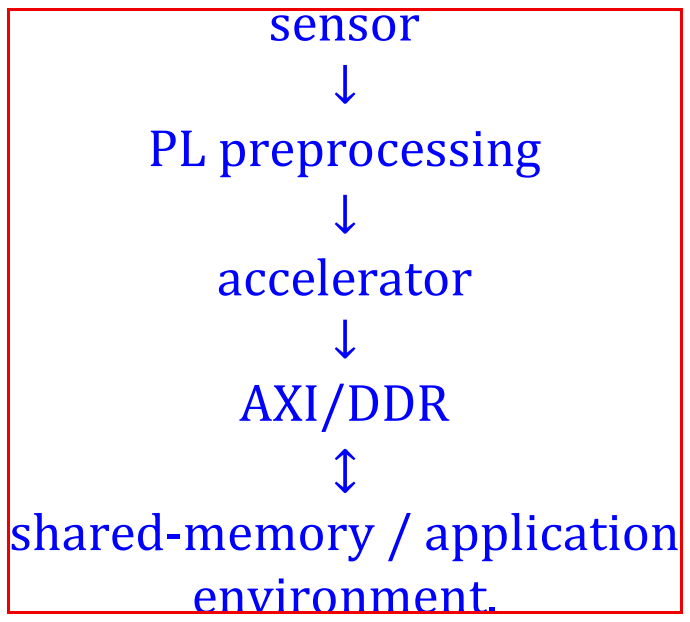

Fig. 4. Sequence of data transaction in modules of Fig. 3.

diagram.

At the PS level, the **Shared DDR / mailbox** provides ring-buffer storage and metadata exchange. The solid connection labeled **ring-buffer I/O** represents application-level buffer delivery to the APU/Linux environment. Thus, the mandatory local application path remains logically distinct from the assurance infrastructure as its sequence shown in Fig. 4. None of the learned diagnosis mechanisms introduced later lies directly on this critical application path.

### *C. Sensor-Health Observability*

The sensor-state component $\mathbf{s}_t$ is intentionally constructed from **two complementary observation points**: (i) the external **Sensor-health sidecar** operates on the physical-signal tap. It generates higher-level health descriptors such as signal-quality degradation, drift, dropout behavior, and acquisition-timing irregularity. Its dashed *health features* connection terminates at the timestamped telemetry bus; (ii) a second source is provided by the **PL telemetry taps**, which observe the digital sensor acquisition and preprocessing path. Their measurements include acquisition/status events and implementation-specific counters describing the behavior of the sensor interface. The corresponding dashed route, explicitly labeled $\mathbf{s}_t$ in Fig. 3, also terminates at the telemetry merge point. The distinction between these two sources is intentional:

$$\mathbf{s}_t = \left[\mathbf{s}_t^{\mathrm{phys}}, \mathbf{s}_t^{\mathrm{dig}}\right],$$

where $\mathbf{s}_t^{\mathrm{phys}}$ describes physical-signal integrity and $\mathbf{s}_t^{\mathrm{dig}}$ describes the acquisition/interface state. This decomposition gives the causal model evidence with which to distinguish a degraded physical measurement from a digital acquisition or interface problem. The two sources are therefore combined only after they have been independently timestamped and identified.

### *D. APU Inference and Microarchitectural Telemetry*

The APU/Linux subsystem produces the inference and processor-observation domains introduced in Section III, $\mathbf{q}_t$ and $\mathbf{h}_t$. The inference runtime generates model-level quantities such as the active model, numerical precision, execution latency, confidence, predictive entropy, and the out-of-distribution statistic that will be developed in Section V. These form the inference-state vector $\mathbf{q}_t$.

Processor execution is monitored using selected performance-monitoring-unit/hardware-performance-counter events to construct $\mathbf{h}_t$. Hardware counters have previously been shown to provide useful low-overhead evidence for anomalous or malicious execution on resource-constrained systems [5]. Because only a limited number of events can typically be observed simultaneously, the final HPC subset will be selected during characterization and frozen before evaluation rather than assuming unrestricted counter availability. For a cumulative hardware counter $z_j$, the interval feature is evaluated as

$$\delta z_{j,t} = z_j(\tau_t) - z_j(\tau_{t-1}), \tag{24}$$

with overflow handled before normalization. An important architectural detail in the final Fig. 3 is that $\mathbf{q}_t$ and $\mathbf{h}_t$ do **not** enter the central telemetry bus directly. Instead, processor-side observations are written to the **Shared DDR / mailbox** together with sequence and timing metadata. The dashed **IPI** path then notifies the RPU that the corresponding record is available. This mechanism prevents the processor-originated observations from being confused with the independently acquired PL and sensor-monitoring channels. The final RPU synchronization stage therefore receives processor-side information through the shared-memory/IPI route and bus-level telemetry through a separate route.

### *E. Accelerator and Memory Telemetry*

Accelerator and memory activity form the state vector $\mathbf{g}_t$. The PL **AXI performance monitor** instruments the memory/interconnect path without becoming part of the functional inference calculation. The monitored quantities can include transferred bytes, transaction counts, transfer latency, and relevant accelerator/memory activity [21]. An instrumentation envelope is

$$\boldsymbol{g}_t = [B_t^r, B_t^w, N_t^r, N_t^w, D_t^r, D_t^w, U_t^{acc}, S_t^{acc}]^T, \tag{25}$$

where $B_t^{r/w}$ denotes read/write data volume, $N_t^{r/w}$ the corresponding transaction counts, $D_t^{r/w}$ latency-related measurements, $U_t^{acc}$ accelerator utilization, and $S_t^{acc}$ status/event information. The dashed path labeled $\mathbf{g}_t$ transports these measurements to the timestamped telemetry bus. The architecture intentionally keeps accelerator status separate from memory/interconnect behavior. A latency excursion can arise from accelerator malfunction, memory contention, DMA behavior, or legitimate workload variation. A root-cause model should therefore observe these mechanisms separately before deciding whether they represent a common cause.

### *F. Power, Thermal, and Operating-Point Telemetry*

Power/thermal observability is distributed across PS and PL resources. The **PS SYSMON / clock control** block provides processor-side thermal, voltage/supply-state, and current operating-point information. The dashed connection labeled **V/T** delivers this state to the **Timestamped telemetry bus**. The PL contains the complementary **PL SYSMON + control hooks** block. It exposes programmable-logic temperature and rail-state information while also providing hardware-facing control hooks for reset, clock modification, and partial-reconfiguration requests [20], [22]. Thus, the same physical subsystem can be both an observation source and an actuation target, but those directions remain explicitly separated in Fig. 3. Telemetry uses dashed connections; control uses dash-dot connections. This is essential for causal interpretation. If the controller deliberately changes the operating point, the resulting power, temperature, or latency change must be interpreted as a consequence of the action rather than as an uncontrolled disturbance.

### *G. Communication Telemetry and Optional Edge–Cloud Operation*

Fig. 3 deliberately separates the application-level cloud connection from the network-observation path. The solid **offload / results** connection originates from the APU/Linux environment and terminates at the optional edge/cloud resource. It represents the application interface through which computation may be remotely executed or results exchanged when remote placement is selected.

By contrast, the **Communication telemetry** block observes the network conditions under which such remote execution must operate. The dashed path labeled **RTT / loss / $\mathbf{n}_t$** carries network state to the timestamped telemetry bus. The corresponding state can include round-trip delay, packet loss or retransmission behavior, effective bandwidth, and current offload status. Maintaining the two connections separately prevents a common modeling ambiguity: *the existence of a cloud path does not imply that the path is currently suitable for recovery*. Section VI will allow $a^E$ only when the measured $\mathbf{n}_t$ satisfies the corresponding latency, communication, and residual-risk constraints. Cloud connectivity therefore remains optional rather than structurally required.

### *H. Common Timestamp and Telemetry-Record Structure*

Cross-layer diagnosis requires observations from different hardware and software sources to be compared within a common temporal reference. The architecture therefore uses the **TSGEN timebase** as the canonical timestamp source. The timestamp reaches the telemetry bus through the dashed connection labeled **ts**. Zynq UltraScale+ provides a 64-bit timestamp-generation facility suitable for this role [20], [23]. Each acquired telemetry item is represented as

$$\mathcal{R}_i^{(m)} = \langle m, \tau_i, n_i, \boldsymbol{x}_i, \boldsymbol{f}_i \rangle, \tag{26}$$

where $m$ identifies the measurement source, $\tau_i$ is its canonical timestamp, $n_i$ is a monotonically increasing source sequence number, $\mathbf{x}_i$ is the payload, and $\mathbf{f}_i$ contains validity, overflow, or source-specific status flags. The sequence number is retained in addition to the timestamp so that dropped or overwritten telemetry records become observable. For feature $j$, the measurement age is

$$A_{j,t} = t\Delta - \tau_{j,t}^{last}, \tag{27}$$

and the validity mask introduced in Section III becomes

$$m_{j,t} = \begin{cases} 1, & A_{j,t} \le A_j^{max} \wedge f_{j,t}^{valid} = 1, \\ 0, & otherwise \end{cases}. \tag{28}$$

A stale or missing measurement is therefore not silently interpreted as a numerically valid zero.

### *I. Timestamped Telemetry Merge Point*

The central **Timestamped telemetry bus** in Fig. 3 is a logical aggregation point, not the diagnostic model itself. It receives the timestamped observations arriving from the sensor-health sidecar, the PL sensor/status telemetry taps, the AXI Performance Monitor, PS power/thermal monitoring, PL power/thermal monitoring, the communication monitor, and the TSGEN timestamp source. The merged bus-level observation can therefore be represented as

$$\mathcal{B}_t = \{\boldsymbol{s}_t^{phys}, \boldsymbol{s}_t^{dig}, \boldsymbol{g}_t, \boldsymbol{p}_t, \boldsymbol{n}_t, \tau_t, \boldsymbol{m}_t^B\}, \tag{29}$$

where $\mathbf{m}_t^B$ is the validity state associated with bus-delivered measurements. The dashed **aligned stream** path carries the resulting time-referenced telemetry from the bus toward the RPU. This distinction is important: the bus **merges and time-references measurements**, whereas the RPU performs the final assembly of all observation domains.

### *J. RPU Real-Time Supervisor and Final Frame Assembly*

The RPU real-time supervisor provides the deterministic boundary between telemetry acquisition and causal inference. It receives two logically different information streams: (i) $\mathcal{B}_t$, through the aligned-stream path from the timestamped telemetry bus; and (ii) the processor-side inference and HPC observations $(\mathbf{q}_t, \mathbf{h}_t)$, obtained from the Shared DDR/mailbox following the IPI notification. The RPU applies timestamp-window alignment, source-sequence checking, freshness validation, missing-data masking, and configuration tagging before releasing an observation to the learned assurance model. The synchronized frame of Section III is therefore physically realized as

$$\mathcal{F}_t = \Psi_{RPU}(\mathcal{B}_t, \boldsymbol{q}_t, \boldsymbol{h}_t, \boldsymbol{k}_t), \tag{30}$$

with

$$\mathcal{F}_t = \{\mathbf{o}_t, \mathbf{m}_t, \mathbf{k}_t, \tau_t\}.$$

The operator $\Psi_{\mathrm{RPU}}(\cdot)$ is deterministic synchronization and validation logic; it is **not** the learned causal model. This partition is important because the temporal alignment required by Section III is performed before Section V receives the data.

Table II summarizes the physical realization of the synchronized state defined in Section III and establishes a direct correspondence between each abstract state variable, its instrumentation source, the representative measurements acquired from that source, and the transport path shown in Fig. 3. This mapping is important because the proposed diagnosis framework does not operate on an unspecified multimodal feature vector: every component of $\mathbf{o}_t$ is associated with an identifiable physical or computational subsystem and a concrete acquisition path. In particular, sensor health is obtained from both the physical-signal sidecar and the digital PL telemetry taps, whereas the APU-originated inference and PMU/HPC observations reach the RPU through the shared DDR/mailbox and IPI mechanism rather than through the central telemetry bus. The table therefore provides the implementation-level traceability required before the synchronized frame $\mathcal{F}_t$ is passed to the local assurance engine.

With those acquisition paths explicitly defined in Table II, the local assurance engine can operate exclusively on the validated and synchronized frame $\mathcal{F}_t$, without directly accessing heterogeneous hardware registers or reconstructing timing relationships from asynchronous logs.

### *K. Local Assurance Engine*

The synchronized frame $\mathcal{F}_t$ is forwarded from the RPU to the **Local assurance engine** through the corresponding telemetry-side connection shown between the two blocks. The local assurance engine implements the two algorithmic functions developed later in the manuscript: (i) the causal root-cause and uncertainty model of Section V; and (ii) the constrained recovery policy of Section VI. Its logical interface is

$$\mathcal{F}_t \to \{b_t(\boldsymbol{c}), U_t, a_t^*\}, \tag{31}$$

where $b_t(\mathbf{c})$ is the inferred cause belief, $U_t$ represents calibrated diagnostic uncertainty, and $a_t^*$ is a **proposed** recovery action. The return connection from the local assurance engine to the RPU represents this decision/control output. The learned component is deliberately prevented from directly manipulating sensor, clock, or reconfiguration registers. The RPU remains the deterministic dispatch authority. The dashed green boundary labeled **Trusted assurance island** therefore encompasses the supervisory and decision components whose integrity is assumed in Section III. Fig. 3 represents the local assurance engine as a logical PS-resident subsystem rather than fixing it to a specific processor core. Its realized execution placement and measured computational footprint will be reported in Section VII after profiling.

### *L. Configuration State and Recovery-Actuation Plane*

The actuation routes shown in Fig. 3 (i.e., ) correspond directly to the recovery space introduced in (*18*). The system configuration is represented explicitly as

$$\boldsymbol{k}_t = \left[k_t^{sensor}, k_t^{model}, k_t^{precision}, k_t^{DVFS}, k_t^{PL}, k_t^{placement}\right]^T. \tag{32}$$

Table II. PHYSICAL REALIZATION OF THE CROSS-LAYER OBSERVATION STATE

| State | Observation source | Representative variables | Transport shown in Fig. 3 | Final destination |
|---|---|---|---|---|
| $\mathbf{s}_t^{\text{phys}}$ | Sensor-health sidecar | quality, drift, dropout, acquisition timing | **health features** dashed path | Timestamped telemetry bus |
| $\mathbf{s}_t^{\text{dig}}$ | PL telemetry taps | acquisition status, interface/event counters | $\mathbf{s}_t$ dashed path | Timestamped telemetry bus |
| $\mathbf{q}_t$ | APU AI runtime | confidence, entropy, OOD measure, model/precision, inference latency | Shared DDR/mailbox + IPI | RPU |
| $\mathbf{h}_t$ | APU PMU/HPC collector | cycles, instructions, cache/branch/stall events | Shared DDR/mailbox + IPI | RPU |
| $\mathbf{g}_t$ | AXI Performance Monitor / accelerator | bytes, transactions, latency, utilization, accelerator state | $\mathbf{g}_t$ dashed path | Timestamped telemetry bus |
| $\mathbf{p}_t^{PS}$ | PS SYSMON / operating-state registry | temperature, voltage state, clock/operating point | **V/T** dashed path | Timestamped telemetry bus |
| $\mathbf{p}_t^{PL}$ | PL SYSMON | temperature, rail state | dashed PL monitoring path | Timestamped telemetry bus |
| $\mathbf{n}_t$ | Communication monitor | RTT, loss/retransmission, bandwidth, offload state | **RTT/loss/n**$_t$ dashed path | Timestamped telemetry bus |
| $\tau_t$ | TSGEN | global 64-bit timestamp | **ts** dashed path | Timestamped telemetry bus |
| $\mathbf{k}_t$ | RPU configuration registry | sensor, model, precision, DVFS, PL state, placement | internal supervisory state | RPU / assurance engine |
| $\mathbf{m}_t$ | RPU synchronization layer | valid, missing, stale, overflow state | derived during frame assembly | Assurance engine |

These components describe the current sensor-selection state, active model, numerical precision, processor/accelerator operating point, programmable-logic configuration, and local-versus-remote execution placement. The RPU exposes four hardware-facing recovery paths in Fig. 3: (i) the **masks / mux** path reaches the Sensor I/O + preprocessing subsystem and realizes sensor selection, exclusion, or failover $a^S$; (ii) the **precision** path reaches the AI accelerator/PR region and realizes model or numerical-precision adaptation $a^M$; (iii) the **DVFS** path reaches the PS SYSMON/clock-control subsystem and realizes supported operating-point adaptation $a^V$; (iv) The **reset / clock / PR** path reaches the PL SYSMON + control-hooks block and realizes accelerator reset, clock-domain control, or partial-reconfiguration requests associated with $a^H$. The edge/cloud action $a^E$ is realized at the application level through the APU **offload / results** interface after the policy has determined that remote execution is admissible. The **fleet report** path is separate. It represents an optional outward assurance/control-plane action through which health, diagnosis, or recovery status may be aggregated remotely. It does not participate in the local diagnosis loop and is not required for fault containment. Every accepted recovery operation is recorded as

$$\mathcal{T}_t^a = \langle a_t, \boldsymbol{k}_t, \boldsymbol{k}_t', \tau_t^a, H_v, r_t \rangle, \tag{33}$$

where $\mathbf{k}_t'$ is the requested post-action configuration, $\tau_t^a$ is the commit timestamp, $H_v$ is the verification horizon defined in Section III, and $r_t$ identifies the rollback state if post-action verification fails. Recording the transaction prevents legitimate recovery-induced telemetry changes from being mistaken for new disturbances.

### *M. Controlled Characterization and Fault-Injection Plane*

The dotted connections in Fig. 3 (i.e., ·········) implement the intervention methodology introduced by (*11*)–(*13*). These paths are **experimental only** and are disabled during ordinary deployment. The external **Experiment controller** provides disturbance severity, intervention scheduling, and reference labels and exposes four controlled perturbation mechanisms. *First*, the dotted **controlled sensor degradation** path terminates at the physical-sensor block. It is used to introduce controlled sensor-domain perturbations such as noise, drift, dropout, or another modality-specific degradation. *Second*, the **fault inject** path terminates at the AI accelerator / PR region and supplies the controlled hardware/accelerator disturbances used to characterize $c_t^H$. *Third*, the **V/f stress** path terminates at the PS SYSMON/clock-control subsystem and is used to exercise supported processor operating-point/resource-stress conditions. *Fourth*, the **rail / temp** path interfaces with the PL

Table III. MAPPING OF RECOVERY AND EXPERIMENTAL INTERVENTION PATHS

| Path in Fig. 3 | Source | Destination | Role |
|---|---|---|---|
| **Masks / mux** | RPU | Sensor I/O + preprocessing | $a^S$: sensor selection/failover |
| **Precision** | RPU | AI accelerator / PR region | $a^M$: model/precision adaptation |
| **DVFS** | RPU | PS SYSMON / clock control | $a^V$: operating-point adaptation |
| **Reset / clock / PR** | RPU | PL SYSMON + control hooks | $a^H$: reset, clock control, partial reconfiguration |
| **Offload / results** | APU | Optional edge/cloud | $a^E$: application placement/offloading |
| **Fleet report** | Local assurance engine | Optional edge/cloud | Optional remote health/recovery reporting |
| **Controlled sensor degradation** | Experiment controller | Physical sensors | $\mathrm{do}(c^S = \rho)$ |
| **Fault inject** | Experiment controller | AI accelerator / PR region | $\mathrm{do}(c^H = \rho)$ |
| **V/f stress** | Experiment controller | PS clock/operating-point subsystem | Controlled resource/operating stress |
| **Rail / temp** | Experiment controller | PL SYSMON/control environment | Controlled PL power/thermal stress |

SYSMON + control-hooks subsystem and provides controlled programmable-logic power/thermal stress conditions. These routes provide the experimental realization of $\mathrm{do}(c_t^k = \rho)$ from Section III.

The **Experiment controller** additionally records the actual injected disturbance and requested severity $\rho_t^k$ . These reference labels are available during characterization, training, and evaluation but are **never supplied as diagnostic features during deployment inference**. This separation prevents causal labels from leaking directly into the model input. Cyber disturbances and natural distribution shifts are introduced through the corresponding workload/software/data mechanisms that are described later in Section VII rather than through a dedicated physical injection wire in Fig. 3. That distinction should remain explicit because not all causal classes have the same intervention mechanism.

The distinction between recovery actuation and experimental intervention is summarized in Table III. Although several of these paths terminate at related hardware subsystems, their purposes are fundamentally different. Recovery paths originate from the trusted supervisory loop and are invoked during normal operation to restore or maintain an admissible system state. Experimental-intervention paths originate from the external experiment controller and are enabled only during characterization, training, or validation to impose a known disturbance and establish the cause–effect relationships introduced in Section III. Table III therefore maps each labeled path in Fig. 3 to its source, destination, and mathematical role in either the recovery action space $A$ or the controlled-intervention formulation $\mathrm{do}(c^k = \rho)$.

This separation is essential for the causal claims of the paper. A disturbance generated by the experiment controller supplies ground-truth intervention information for characterization and evaluation, whereas a recovery action is selected only from deployment-time telemetry and diagnostic belief. The intervention label is therefore never exposed to the deployment-time diagnosis or recovery policy.

### *N. Monitoring Nonintrusiveness*

Cross-layer monitoring must not consume enough resources to materially alter the behavior it seeks to diagnose. The architecture therefore avoids continuous textual logging and synchronous storage operations within the critical inference path. Hardware counters accumulate locally; processor observations are transferred through fixed-size buffers; and telemetry is aggregated before causal processing. Monitoring overhead will be quantified against the identical application configuration with assurance instrumentation disabled:

$$\eta_L = \frac{L_{mon} - L_0}{L_0}, \qquad \eta_E = \frac{E_{mon} - E_0}{E_0}, \tag{34}$$

where $L_0$ and $E_0$ denote baseline inference latency and energy and $L_{\text{mon}}, E_{\text{mon}}$ denote the corresponding quantities with the complete monitoring plane active. PL instrumentation overhead will additionally be reported as

$$\boldsymbol{\eta}_{PL} = [\eta_{LUT}, \eta_{FF}, \eta_{BRAM}, \eta_{DSP}], \tag{35}$$

with each element measured relative to the uninstrumented implementation. This evaluation is necessary because a dependability mechanism that imposes excessive computation, memory traffic, or power consumption would undermine the Edge-AI operating assumptions that motivate the architecture.

### *O. Trusted-Base and Failure-Scope Assumptions*

The causal formulation requires the integrity of a minimal observation and control substrate. The timestamp source, source-validity metadata, RPU synchronization logic, configuration registry, action dispatcher, and associated assurance-state storage therefore form the **trusted assurance base** shown conceptually by the trusted-island boundary in

Fig. 3. The paper does not assume that the foreground application is trusted. In fact, anomalous or malicious application behavior is one of the conditions observed through $\mathbf{h}_t$. However, the present work does not claim recovery from an adversary capable of simultaneously compromising the trusted supervisor, rewriting the timestamp, fabricating arbitrary telemetry, and controlling the recovery registry. Under such a threat model the physical state becomes observationally unidentifiable without an additional hardware root of trust. Likewise, missing or corrupted telemetry outside the trusted base is not assumed harmless. It is represented explicitly through $\mathbf{m}_t$ and becomes part of the uncertainty experiments in Sections V, VIII, and IX.

### *P. Interface to the Subsequent Sections*

The architecture in Fig. 3 now provides a concrete implementation for every required component of the system model developed in Section III. The physical and computational platform produces identifiable observations. The timestamped telemetry bus merges the independently instrumented sensor, accelerator/memory, power/thermal, and communication channels. The APU independently supplies $\mathbf{q}_t$ and $\mathbf{h}_t$ through the DDR/mailbox path. The RPU then constructs the complete synchronized frame $\mathcal{F}_t = \{\mathbf{o}_t, \mathbf{m}_t, \mathbf{k}_t, \tau_t\}$ without requiring the learned model to infer timing relationships from unsynchronized logs. This produces the exact interface required later by Section V, which receives $\mathcal{F}_t$ and must determine how the temporally structured evidence is transformed into a cause belief $b_t(\mathbf{c})$, calibrated uncertainty, and compound-event interpretation.

Section VI will receive that diagnostic output together with the current service and configuration state and determines whether one of the concrete actions in Table III belongs to the admissible set defined by (*19*). The RPU then remains responsible for deterministic command dispatch and initiation of the post-action verification interval. Then, Section VII will instantiate the quantities which are deliberately left implementation-specific here i.e., *sampling periods*, *synchronization window*, *feature dimensions*, *selected HPC events*, *accelerator model*, *model precisions*, *DMA configuration*, *ring-buffer depth*, *PL resource cost*, *diagnosis execution placement*, and *perturbation severities*. Finally, Sections VIII and IX will evaluate not only diagnostic and recovery performance but also the architectural assumptions made here i.e., *missing telemetry*, *source freshness*, *synchronization tolerance*, *observer overhead*, *removal of individual telemetry domains*, and *dependence on particular intervention channels*. The resulting implementation flow is therefore as shown in Fig. 5. It is noted that During causal characterization, a separate path operates in parallel as shown in Fig. 6. This separation between **normal operation and controlled intervention**, and between **observation and actuation**, is fundamental to the claims made later in the paper. Now it is ensured that Section V will begin from a well-defined synchronized observation state rather than an ambiguous collection of logs, while Section VI will operate on recovery mechanisms that already have explicit physical realizations.

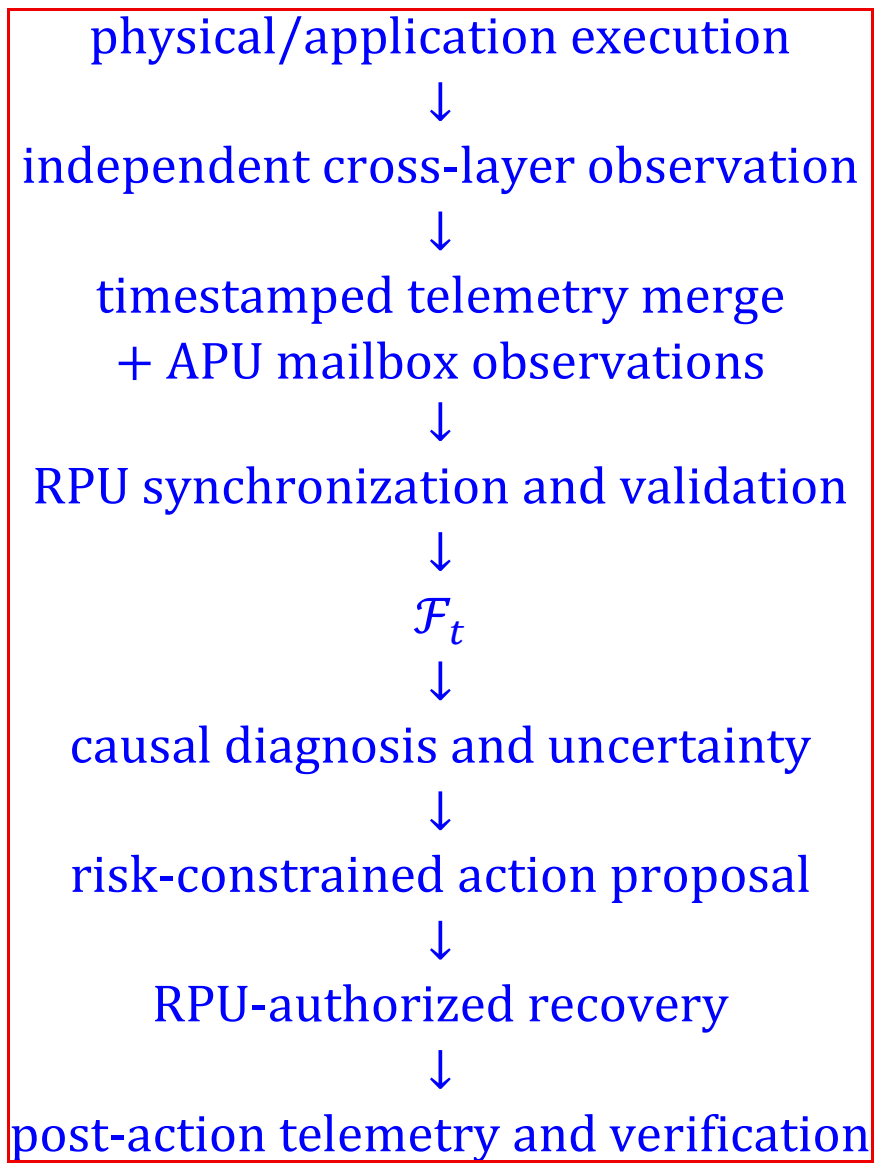

Fig. 5. The resulting implementation flow of the experiments which are established in the paper.

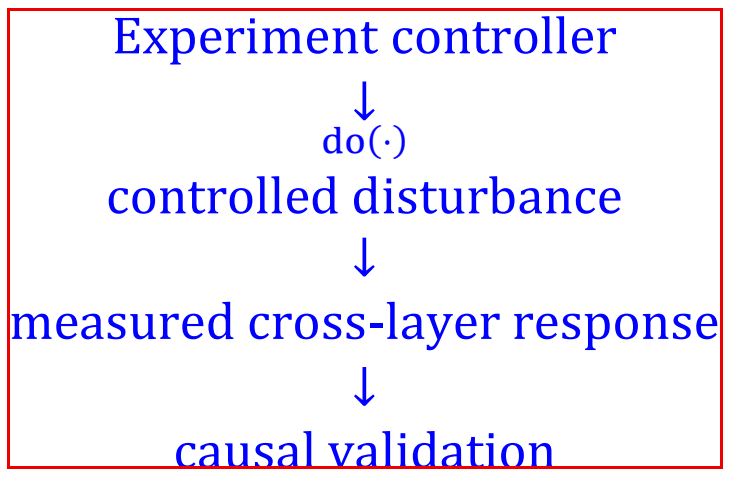

Fig. 6. Separate path operating during characterization phase.

## V. TEMPORAL CAUSAL DIAGNOSIS AND UNCERTAINTY-AWARE ROOT-CAUSE ATTRIBUTION

Section IV establishes the physical acquisition paths and deterministic synchronization mechanisms that produce the cross-layer frame $\mathcal{F}_t = \{\mathbf{o}_t, \mathbf{m}_t, \mathbf{k}_t, \tau_t\}$ . The diagnostic problem is therefore well defined: given a temporal sequence of synchronized frames, determine which of the potentially concurrent disturbance mechanisms introduced in Section III best explains the observed cross-layer behavior, while explicitly representing diagnostic uncertainty. This section develops the proposed **Intervention-Calibrated Temporal Causal Fusion (IC-TCF)** framework for this purpose.

The central design principle is that causal attribution should not be inferred from unrestricted statistical attention alone. Instead, IC-TCF combines three forms of evidence: **(i)** the physical and temporal structure of the architecture in Fig. 3, **(ii)** controlled cause-specific interventions performed during offline characterization, and **(iii)** the passive synchronized telemetry observed during deployment. The first two constrain

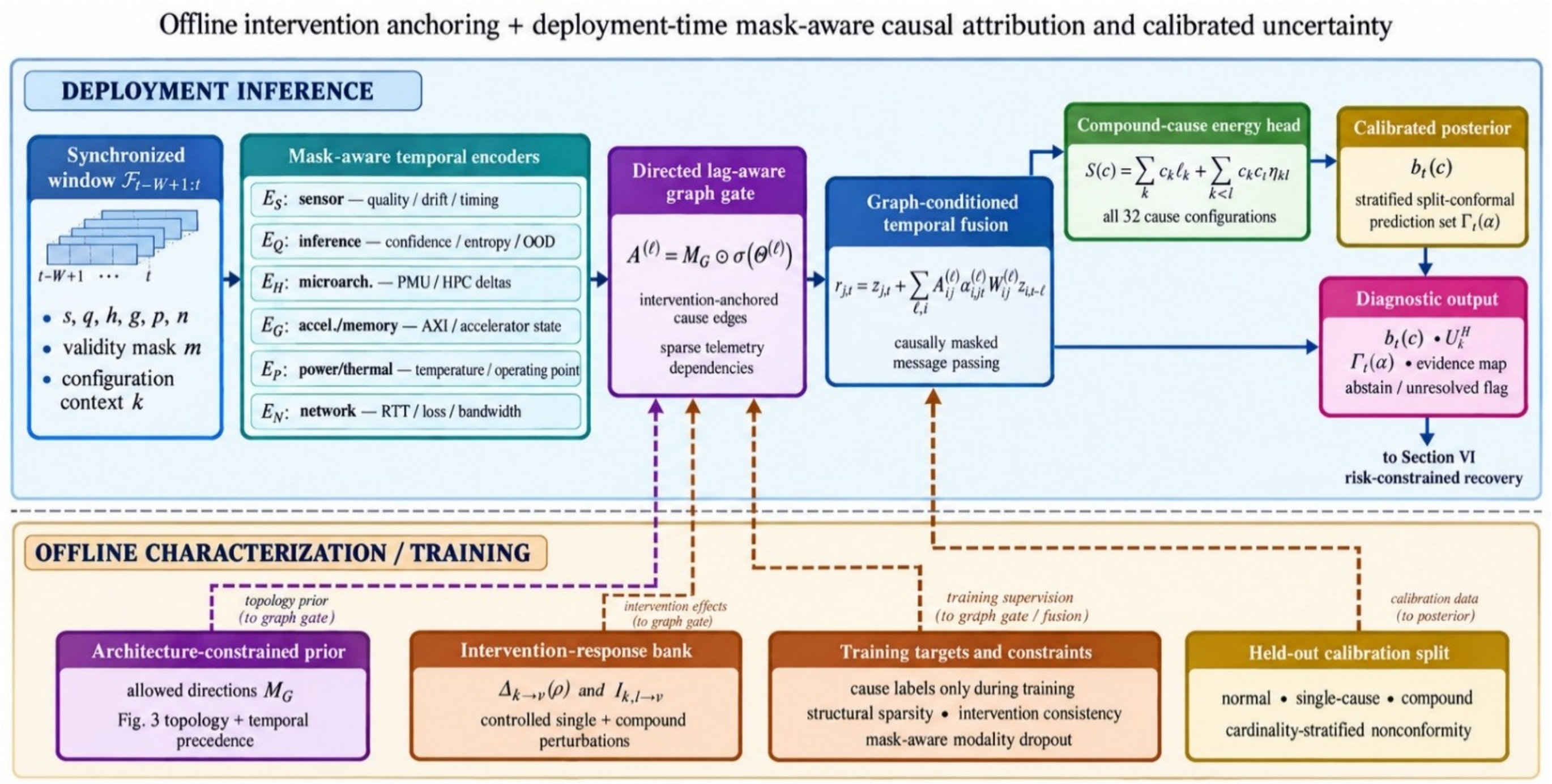

Fig. 7. Intervention-Calibrated Temporal Causal Fusion (IC-TCF) architecture for cross-layer root-cause diagnosis. The upper path represents deployment-time inference. A synchronized temporal window of sensor, inference, microarchitectural, accelerator/memory, power/thermal, and communication telemetry is processed by mask-aware temporal encoders, followed by an architecture-constrained directed lag-aware graph and graph-conditioned temporal fusion. The resulting representation is evaluated over all 32 disturbance configurations by the compound-cause energy head, after which a held-out conformal-calibration stage produces the calibrated prediction set $\Gamma_t(\alpha)$. The final diagnostic output contains the cause posterior, uncertainty indicators, evidence information, and an abstention/unresolved state supplied to the recovery controller of Section VI. The lower path contains the topology-derived structural prior, controlled intervention responses, training constraints, and independent calibration data used offline; none of these privileged labels is exposed during deployment inference.

model learning; only the third is available to online inference.

### A. IC-TCF Overview

At decision instant $t$, the diagnostic engine receives the temporal window $\{\mathcal{F}_{t-W+1}, \dots, \mathcal{F}_t\}$ where $W$ is the diagnosis-window length. IC-TCF processes this window through five successive stages as shown in Fig. 8.

$\mathcal{F}_{t-W+1}$
↓
domain encoding
↓
causal graph constraint
↓
temporal fusion
↓
compound-cause inference
↓
uncertainty calibration

Fig. 8. Stages processed with IC-TCF

Fig. 7 summarizes the complete architecture. The upper path represents deployment inference, whereas the lower path contains information used only during offline characterization, training, and calibration. During deployment, synchronized sensor, inference, microarchitectural, accelerator/memory, power/thermal, and communication observations are first processed independently by mask-aware temporal encoders. Their representations are then propagated through a directed lag-aware graph whose admissible edges are restricted by the physical architecture and whose disturbance-to-observation relationships were anchored using controlled interventions. The fused representation is evaluated over all $2^5 = 32$ possible disturbance configurations, including normal operation and compound events. Finally, a held-out calibration set converts the learned posterior into an uncertainty-aware conformal prediction set. The lower path in Fig. 7 is deliberately isolated from deployment inference. Intervention identity, injected severity, ground-truth cause labels, and calibration targets are used only to learn or calibrate the frozen model and are never presented as online diagnostic features.

### B. Mask- and Configuration-Aware Temporal Encoding

Telemetry distributions depend on the active operating configuration. For example, changing the inference model, numerical precision, or operating point can legitimately alter processor counters, accelerator traffic, latency, and power. Each valid feature is therefore normalized relative to its current configuration,

$$m_{j,\tau} \frac{x_{j,\tau} - \mu_j(\boldsymbol{k}_\tau)}{\sigma_j(\boldsymbol{k}_\tau) + \varepsilon} \tag{36}$$

where $\mu_j(\mathbf{k}_\tau)$ and $\sigma_j(\mathbf{k}_\tau)$ are configuration-conditioned baseline statistics and $m_{j,\tau}$ is the validity indicator defined in Section IV. Normalization alone is insufficient because the model must distinguish a valid numerical value from an unavailable or stale measurement. For each telemetry domain $m$, the normalized observations are therefore augmented with their validity and freshness information,

$$\boldsymbol{u}_t^{(m)} = [\tilde{\boldsymbol{x}}_\tau^{(m)}, \boldsymbol{m}_\tau^{(m)}, \boldsymbol{a}_\tau^{(m)}], \quad (37)$$

where $\mathbf{a}_t^{(m)}$ contains normalized measurement-age information derived from (27). This representation distinguishes three physically different conditions: (i) a valid observation close to baseline, (ii) a valid but anomalous observation, and (iii) a missing or stale observation. Each telemetry domain is then processed independently over the complete temporal window,

$$E_m\big(\boldsymbol{u}_{t-W+1}^{(m)}\big), \qquad m \in \{S, Q, H, G, P, N\}, \quad (38)$$

where $E_m$ is a causal temporal encoder and $\mathbf{z}_{m,t} \in \mathbb{R}^d$ is the domain embedding. Independent encoding preserves domain-specific temporal behavior before cross-layer fusion. For example, a progressive thermal rise and an isolated temperature spike may have similar instantaneous values but represent different temporal phenomena. Likewise, a persistent increase in cache activity should not be interpreted identically to a short workload-dependent burst.

### C. Architecture- and Intervention-Constrained Causal Graph

Unrestricted cross-domain attention can learn statistically predictive but physically implausible dependencies. IC-TCF therefore restricts evidence propagation using an architecture-derived structural mask. For lag $\ell \in 0, \dots, p$, let $\boldsymbol{M}_{\mathcal{G}}^{(\ell)} \in \{0,1\}^{M \times M}$ define the set of admissible directed dependencies. An entry is zero when the corresponding relationship violates the physical topology of Fig. 3 or temporal precedence. The effective learned adjacency is

$$A_{ij}^{(\ell)} = M_{ij}^{(\ell)} \sigma\big(\Theta_{ij}^{(\ell)}\big), \quad (39)$$

where $\Theta_{ij}^{(\ell)}$ is learned and $\sigma(\cdot)$ bounds the soft edge strength. The structural mask specifies which relationships are **possible**, not which are automatically causal. Cause-to-observation relationships are further anchored using the controlled interventions defined in Section III. For disturbance $k$ and monitored variable or domain $v$, the experimentally observed intervention response is

$$\mathbb{E}[v \mid do(c^k = 0)]. \quad (40)$$

Across intervention severities and workloads, the normalized response provides an empirical target for the corresponding disturbance-to-observation graph edges. The resulting intervention-consistency regularizer is $\hat{R}_{k,v}$, where $\hat{R}_{k,v}$ is the normalized intervention-response magnitude and $\omega_{k,v}$ reflects its experimental repeatability. This is central to the causal interpretation of IC-TCF: *a strong disturbance-to-observation edge is discouraged when the corresponding telemetry variable fails to respond reproducibly under deliberate manipulation of that disturbance*. This does **not** imply that every telemetry-to-telemetry edge is experimentally proven causal. Such edges remain architecture- and temporally constrained dependencies used for evidence propagation. The controlled interventions directly anchor the disturbance-to-observation portion of the graph.

### D. Graph-Conditioned Fusion and Compound-Cause Inference

After domain encoding and graph construction, cross-layer evidence is fused only through permitted directed relationships. For target domain $j$,

$$\boldsymbol{r}_{j,t} = \boldsymbol{z}_{j,t} + \sum_{\ell=0}^{p} \sum_{i \in Pa_{\mathcal{G}}(j,\ell)} A_{ij}^{(\ell)} \alpha_{ij,t}^{(\ell)} \boldsymbol{W}_{ij}^{(\ell)} \boldsymbol{z}_{i,t-\ell} \quad (41)$$

where $A_{ij}^{(\ell)}$ denotes learned structural support and $\alpha_{ij,t}^{(\ell)}$ is the event-specific relevance of the corresponding permitted edge. This distinction is important. The structural coefficient determines whether a directed relationship is retained by the learned graph; the relevance coefficient determines how strongly that permitted relationship contributes to the current event. A large relevance coefficient alone is therefore not interpreted as proof of causality. The domain representations are subsequently pooled into

$$\boldsymbol{r}_t = Pool\big(\boldsymbol{r}_{S,t}, \boldsymbol{r}_{Q,t}, \boldsymbol{r}_{H,t}, \boldsymbol{r}_{G,t}, \boldsymbol{r}_{P,t}, \boldsymbol{r}_{N,t}\big). \quad (42)$$

Because the five disturbance mechanisms in Section III may occur concurrently, diagnosis is not formulated as a mutually exclusive multiclass problem. Instead, all $|\mathcal{C}| = 2^5 = 32$ possible cause configurations are evaluated explicitly. For configuration $\mathbf{c} = [c_1, \dots, c_5]^\top$ the compound-cause score is

$$S_t(\boldsymbol{c}) = \sum_{k=1}^{5} c_k\, \ell_{k,t} + \sum_{k<j} c_k\, c_j \eta_{kj,t} \quad (43)$$

where $\ell_{k,t}$ represents evidence for cause $k$ and $\eta_{kj,t}$ captures pairwise interaction between concurrent causes. The configuration posterior is then

$$b_t(\boldsymbol{c}) = \frac{exp[S_t(\boldsymbol{c})/T]}{\sum_{c' \in \mathcal{C}} exp[S_t(\boldsymbol{c}')/T]}, \qquad \boldsymbol{c} \in \mathcal{C} \quad (44)$$

where $T$ is a calibration temperature estimated from validation data. Explicit enumeration of 32 configurations remains computationally inexpensive while avoiding the independence assumption of five unrelated binary classifiers. The maximum-a-posteriori state $\arg\max_{\mathbf{c}\in\mathcal{C}} \; b_t(\mathbf{c})$ is retained for reporting, but it is not sufficient by itself for recovery selection.

*E. Training Objective*

Training jointly optimizes configuration diagnosis, intervention consistency, graph sparsity, and robustness to incomplete telemetry. For a labeled training window with true configuration $\mathbf{c}_t^\star$, the primary loss is $w(\mathbf{c}_t^\star)\log b_t\,(\mathbf{c}_t^\star)$, where $w(\mathbf{c})$ compensates for imbalance among normal, single-cause, and compound events. Sparse graph structure is encouraged through $\sum_\ell \left|\mathbf{A}^{(\ell)}\right|_1$. Robustness to unavailable telemetry is promoted by randomly masking admissible observation subsets during training, $\log b_t^{(\tilde{\mathbf{m}})}\,(\mathbf{c}_t^\star)$. The complete objective is

$$\boxed{\mathcal{L} = \mathcal{L}_{cfg} + \lambda_{do}\mathcal{L}_{do} + \lambda_{sp}\mathcal{L}_{sp} + \lambda_{miss}\mathcal{L}_{miss}} \tag{45}$$

The conformal calibration samples introduced next are held out entirely from optimization of (45).

*F. Calibrated Uncertainty and Diagnostic Decision*

A high posterior probability does not guarantee that the diagnosis is reliable under ambiguous or partially observed conditions. IC-TCF therefore supplements the learned posterior with split-conformal calibration [18]. For calibration sample $i$ with true configuration $\mathbf{c}_i^\star$, define the nonconformity score $1 - b_i(\mathbf{c}_i^\star)$. Because normal, single-cause, and compound events can have substantially different calibration behavior, samples are stratified according to cause cardinality. The resulting conformal prediction set is

$$\left\{\mathbf{c} \in \mathcal{C}\colon 1 - b_t(\mathbf{c}) \le q_{\nu(\mathbf{c})}(1-\alpha)\right\},$$

where $q_{\nu(\mathbf{c})}(1-\alpha)$ is the finite-sample calibration quantile for the corresponding cardinality stratum. Three complementary uncertainty quantities are retained as:

$$U_t^H = -\frac{\sum_{c\in\mathcal{C}} b_t\,(\boldsymbol{c})\,log\,b_t\,(\boldsymbol{c})}{log|\mathcal{C}|} \tag{46}$$

conformal ambiguity is quantified by

$$U_t^\Gamma = \frac{|\Gamma_t(\alpha)|}{|\mathcal{C}|} \tag{47}$$

And telemetry incompleteness is represented as

$$U_t^M = 1 - \frac{\sum_j \omega_j\, m_{j,t}}{\sum_j \omega_j} \tag{48}$$

where $\omega_j$ reflects the relevance of feature $j$ to the diagnosed state. The complete uncertainty state is therefore

$$\boldsymbol{U}_t = [U_t^H, U_t^\Gamma, U_t^M]^T. \tag{49}$$

These quantities have different interpretations. A high $U_t^H$ means the learned posterior itself is diffuse; a high $U_t^\Gamma$ means calibration cannot narrow the diagnosis to a small set; and a high $U_t^M$ means the diagnostic evidence is incomplete. Later, Section VI will use this information when deciding whether an aggressive recovery action is justified. The calibrated output is then assigned one of three operational states, $d_t \in \{\text{resolved}, \text{ambiguous}, \text{unsupported}\}$. A diagnosis is **resolved** when the calibrated prediction set is sufficiently narrow, typically a singleton under the default policy. It is **ambiguous** when multiple configurations remain statistically plausible, $|\Gamma_t(\alpha)| > 1$. It is **unsupported** when the calibrated model cannot retain a valid known configuration, for example when $\Gamma_t(\alpha) = \emptyset$. Importantly, ambiguous and unsupported states are not forcibly converted into fault labels. They are passed explicitly to the recovery controller, where they restrict the aggressiveness of autonomous action. The formal conformal guarantee applies only under its calibration assumptions. Consequently, coverage under unseen workloads, unseen severities, and other distribution shifts will be evaluated empirically rather than assumed.

*G. Online Inference and Root-Cause Diagnosis*

Once training and calibration are complete, deployment requires only synchronized passive telemetry, configuration context, the frozen model, and the previously determined calibration parameters. The online procedure is summarized in **Algorithm 1Error! Reference source not found.**. For auditability, a domain-level evidence map may additionally be generated for selected anomalous events by measuring the change in diagnostic belief when individual telemetry domains are removed. Because this analysis adds computation and does not define the diagnosis itself, its detailed formulation and evaluation are deferred to the interpretability and ablation analysis rather than included in the primary inference chain. The final diagnostic message is

$$\mathcal{D}_t = \{b_t(\boldsymbol{c}), \hat{\boldsymbol{c}}_t^{MAP}, \Gamma_t(\alpha), \boldsymbol{U}_t, d_t, \mathcal{E}_t, \tau_t\}. \tag{50}$$

The diagnostic stage terminates at this point. It does not directly issue sensor failover, model switching, DVFS, accelerator reset, partial reconfiguration, or cloud-migration commands. Instead, the message $\mathcal{D}_t$ is forward to Section VI for risk-constrained recovery. This separation is fundamental to the proposed framework. A high-probability diagnosis does

**Algorithm 1** IC-TCF Online Root-Cause Diagnosis

**Input:** synchronized window $\mathcal{F}_{t-W+1}$; frozen temporal encoders; frozen graph $\mathbf{A}^{(0)}$; configuration-conditioned normalization statistics; conformal calibration quantiles.

**Output:** diagnostic state $\mathcal{D}_t$.

**1**: Validate telemetry freshness and construct the corresponding validity masks.

**2**: Normalize each telemetry domain using the active configuration $\mathbf{k}_t$.

**3**: Encode the six domains independently over the temporal window.

**4**: Propagate cross-domain evidence only through graph-permitted directed lagged dependencies.

**5**: Fuse the resulting domain representations.

**6**: Evaluate all 32 disturbance configurations and compute (b_t(\mathbf c)).

**7**: Construct $\Gamma_t(\alpha)$, compute $\mathbf{U}_t$, and determine whether the event is resolved, ambiguous, or unsupported.

**8**: Return the complete diagnostic state to the recovery controller.

not automatically imply that its nominally corresponding recovery action is safe or optimal. Section VI therefore will operate on the **complete posterior and calibrated uncertainty**, together with the current service and configuration states, to determine whether any autonomous action is admissible. Hence, the resulting methodological chain from Sections IV–VI is as shown in **Error! Reference source not found.**. It is noted that the evidence map $\mathcal{E}_t$ in (50) is retained for diagnostic auditability and later interpretability analysis; it is not required by the primary recovery optimization of Section VI.

## VI. RISK-CONSTRAINED SELF-HEALING AND POST-ACTION VERIFICATION

Section V produces the calibrated diagnostic state

$$\{b_t(\mathbf{c}), \hat{\mathbf{c}}_t^{\mathrm{MAP}}, \Gamma_t(\alpha), \mathbf{U}_t, d_t, \mathcal{E}_t, \tau_t\}$$

where the diagnostic state $d_t$ distinguishes resolved, ambiguous, and unsupported observations. The recovery stage must therefore solve a different problem from diagnosis. It must determine whether an intervention is justified, which physically available action is compatible with the uncertain cause state, whether that action is likely to preserve the application constraints, and whether its measured post-action outcome actually constitutes recovery.

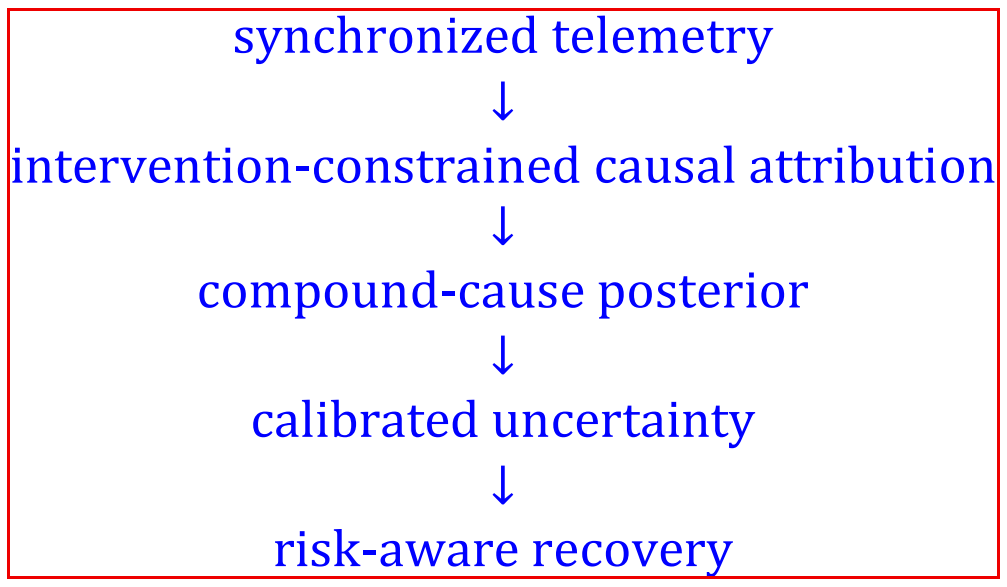

Fig. 9. Methodological chain of action from sections IV-VI.

It is noted that the diagnosis alone does not determine which recovery action should be executed. A cause may remain ambiguous, multiple actions may be technically feasible, and an otherwise appropriate intervention may violate latency, energy, quality, or residual-risk constraints.

The recovery controller therefore operates on the **complete diagnostic posterior and uncertainty state**, rather than mapping the MAP diagnosis directly to a predefined command. Its objective is to select the lowest-risk admissible action, execute it transactionally through the RPU mechanisms of Fig. 3, and accept the resulting configuration only after its measured post-action behavior satisfies the required service and recovery conditions. The resulting control flow is shown in Fig. 10.

Fig. 11 illustrates the complete closed-loop recovery process implemented after the diagnostic stage of Section V. The pipeline begins with the calibrated diagnostic state $\mathcal{D}_t$, which contains the compound-cause posterior $b_t(\mathbf{c})$, conformal prediction set $\Gamma_t(\alpha)$, uncertainty vector $\mathbf{U}_t$, diagnostic state $d_t$, evidence information, and timestamp. This information is evaluated together with the current service state $\mathbf{y}_t$, active configuration $\mathbf{k}_t$, and previous action to construct a diagnosis-aware set of candidate recovery actions.

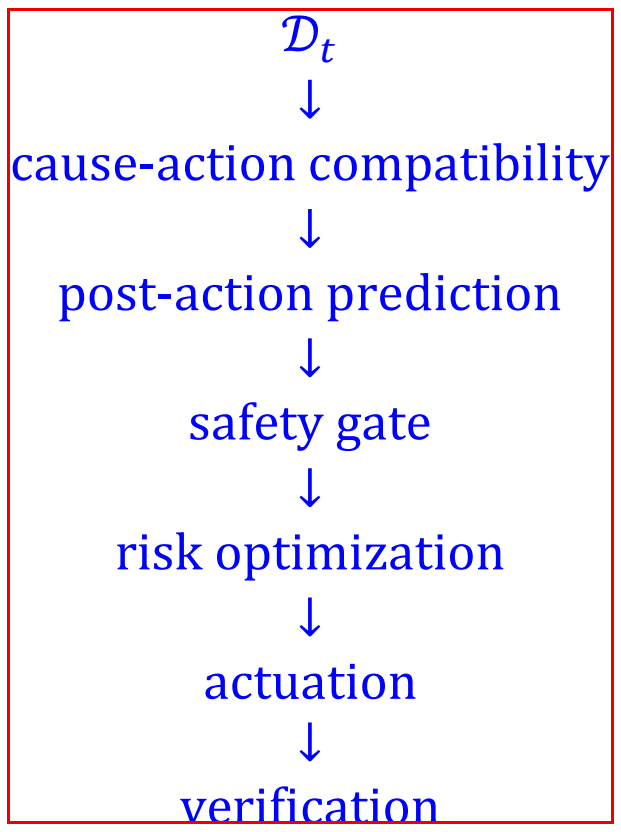

Fig. 10. Control flow of the recovery controller.

The second stage evaluates the physical compatibility between each candidate action and the plausible disturbance configurations. For resolved diagnoses, the complete action library may be considered subject to the subsequent safety constraints. Ambiguous diagnoses restrict the candidate set to actions that remain reasonable across the configurations retained by $\Gamma_t(\alpha)$, whereas unsupported diagnoses are limited to predefined safe actions such as hold or escalation. This prevents a high-probability but uncertain diagnosis from being converted directly into an aggressive recovery command.

Each surviving candidate is then passed to the action-conditioned outcome predictor, which estimates the post-action distributions of latency, energy, inference quality, and residual risk. These distributions are evaluated by the uncertainty-aware safety gate. The violation probabilities are tightened as diagnostic uncertainty increases, and candidate actions whose predicted post-action distributions violate the latency, energy, quality, or residual-risk requirements are removed. The remaining actions form the admissible set $\mathcal{A}_{\text{adm}}(t)$.

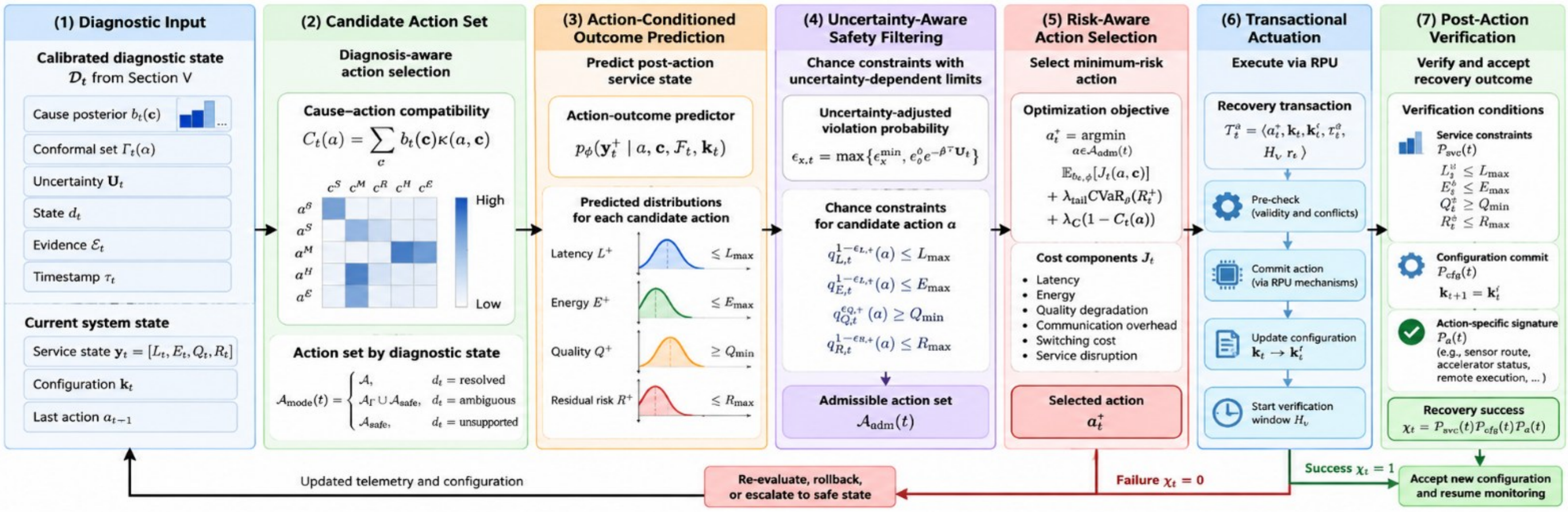

Fig. 11. Risk-constrained self-healing and post-action verification pipeline. The calibrated diagnostic state generated by IC-TCF is first used to construct a diagnosis-dependent recovery-action set. Each candidate action is evaluated by an action-conditioned predictor that estimates the resulting latency, energy, inference quality, and residual risk. An uncertainty-sensitive safety gate removes actions that may violate the operational constraints, after which a posterior-aware optimizer selects the minimum-risk admissible action. The RPU executes the selected recovery transaction and initiates a verification interval. Recovery is accepted only when the measured post-action service state, committed configuration, and action-specific recovery signature satisfy the corresponding verification conditions; otherwise the controller re-evaluates the state, rolls back when appropriate, or escalates to a safe state.

The risk-aware selection stage ranks the admissible candidates using the posterior-expected recovery cost, a tail-risk term, and cause–action compatibility. The resulting action $a_t^*$ is therefore selected using the complete diagnostic belief rather than only the MAP cause. The RPU subsequently executes the selected action transactionally through a pre-check, commit, configuration update, and initiation of the verification interval $H_v$.

The final stage verifies the measured post-action behavior. Recovery is accepted only when three complementary conditions hold: the service constraints are satisfied, the requested configuration has actually been committed, and the action-specific recovery signature is observed. These conditions produce the recovery indicator $\chi_t$. If $\chi_t = 1$, the new configuration is accepted and normal monitoring resumes. If $\chi_t = 0$, the updated telemetry and configuration are fed back to the diagnostic and recovery loop, where the system may re-evaluate the event, roll back the attempted configuration, select another admissible action, or escalate to a predefined safe state.

The feedback path shown at the bottom of Fig. 11 is therefore fundamental to the proposed approach: recovery is treated as a verified closed-loop process rather than a one-shot command issued immediately after diagnosis.

### A. Recovery Problem and Decision Interface

The recovery controller receives

$$\boxed{\mathcal{X}_t^{ctrl} = \{\mathcal{D}_t, \mathcal{F}_t, \boldsymbol{y}_t, \boldsymbol{k}_t, a_{t-1}\}} \tag{51}$$

where $\mathcal{D}_t$ is the calibrated diagnostic output of Section V, $\mathcal{F}_t$ is the current synchronized cross-layer state used to condition the expected outcome of candidate recovery actions, $\mathbf{y}_t = [L_t, E_t, Q_t, R_t]^\top$ ontains the current service state, $\mathbf{k}_t$ is the active system configuration, and $a_{t-1}$ denotes the most recent recovery action. The available action library remains as $\{a^0, a^S, a^M, a^V, a^H, a^E\}$ where $a^0$ denotes no intervention, $a^S$ sensor selection or failover, $a^M$ model/precision adaptation, $a^V$ DVFS/resource adaptation, $a^H$ accelerator reset or programmable-logic reconfiguration, and $a^E$ edge–cloud migration. The recovery stage does not assume that $\hat{\mathbf{c}}_t^{\text{MAP}}$ is certainly correct. Instead, the posterior $b_t(\mathbf{c})$, conformal set $\Gamma_t(\alpha)$, and uncertainty vector $\mathbf{U}_t$ remain active throughout candidate filtering and optimization. This is particularly important when different root causes produce overlapping symptoms or when telemetry is incomplete.

It is noted that before candidate-action evaluation, the RPU verifies that the diagnostic timestamp $\tau_t$ and configuration context associated with $\mathcal{D}_t$ remain consistent with the current system state; stale diagnoses produced before a subsequent configuration transition are discarded.

### B. Diagnosis-Aware Candidate Action Set

Each action has a different physical relationship to the possible disturbance configurations. We therefore define a cause–action compatibility function $\kappa(a, \mathbf{c}) \in [0,1]$ and its posterior support

$$\sum_{c \in \mathcal{C}} b_t(\boldsymbol{c})\kappa(a, \boldsymbol{c}) \tag{52}$$

Table IV. RECOVERY ACTIONS, PRIMARY TARGETS, SIDE EFFECTS, AND VERIFICATION SIGNATURES

| Action | Primary intended condition | Physical realization in Fig. 3 | Principal recovery mechanism | Important side effects / costs | Key post-action verification |
|---|---|---|---|---|---|
| $a^0$: hold | Normal state; weak or uncertain event with acceptable risk | No configuration change | Maintain the current state while collecting more evidence | Possible delayed containment if a real disturbance persists | Service remains admissible; uncertainty decreases or the event disappears |
| $a^S$: sensor failover | Primarily $c^S$ | masks / mux | Exclude the degraded sensor/path and select a healthy source | Switch latency; possible loss of modality or redundancy | Route changes; acquisition quality recovers; inference quality remains admissible |
| $a^M$: model / precision adaptation | Primarily $c^D$; selected resource-pressure cases | precision / model control | Change model or numerical precision to restore robustness or resource margin | Accuracy, energy, or latency tradeoff; switching overhead | Requested model/precision is active; quality and latency satisfy limits |
| $a^V$: DVFS / resource adjustment | Primarily $c^R$ | DVFS / operating-point path | Adjust the operating point or resource allocation to reduce thermal/resource stress | Energy, throughput, and thermal tradeoffs | Operating point changes as commanded; thermal/resource and latency behavior improve |
| $a^H$: reset / reconfiguration | Primarily $c^H$ | reset / clock / PR | Clear corrupted accelerator state or replace the affected PL configuration | Temporary interruption; reset/reconfiguration energy and latency | Reset/PR completes; accelerator status and transactions return to an admissible state |
| $a^E$: edge-cloud migration | Hardware/resource conditions and selected isolation scenarios | offload / results | Bypass impaired local execution by moving service to edge/cloud | Network dependence; communication energy; privacy and latency costs | Remote execution succeeds; network state and end-to-end latency remain admissible |

Here, $\kappa = 0$ indicates that action $a$ has no justified recovery mechanism for configuration $c$, whereas larger values indicate stronger expected compatibility. The compatibility structure is initialized from the physical mechanisms of Fig. 3 and subsequently quantified from the recovery experiments of Section VII. The diagnostic state $d_t$ further controls how aggressively the system may act:

$$\begin{cases} \mathcal{A}, & d_t = resolved, \\ \mathcal{A}_\Gamma(t) \cup \mathcal{A}_{safe}, & d_t = ambiguous, \\ \mathcal{A}_{safe}, & d_t = unsupported. \end{cases} \tag{53}$$

For an ambiguous diagnosis, $\mathcal{A}_\Gamma(t)$ contains only actions that remain physically reasonable for the configurations retained in $\Gamma_t(\alpha)$. Unsupported observations are restricted to predefined safe actions such as hold, isolation, or escalation rather than forcing an aggressive autonomous recovery. This is one of the principal roles of calibrated uncertainty in the proposed framework: **diagnostic ambiguity directly changes the set of actions the controller is allowed to consider**.

Table IV summarizes the intended role of each action, its implementation path in Fig. 3, its principal recovery mechanism, its side effects, and the evidence required after execution. The table should be interpreted as a physical action library rather than a fixed cause-to-action lookup. For example, model/precision adaptation may also mitigate resource pressure, while edge–cloud migration can bypass several local failure modes. The final selection therefore depends on the predicted post-action state rather than on the table alone.

### *C. Action-Conditioned Outcome Prediction*

For each candidate action $a$, the controller predicts the distribution of the post-action service state,

$$p_\phi(\boldsymbol{y}_t^+ \mid a, \boldsymbol{c}, \mathcal{F}_t, \boldsymbol{k}_t) \tag{54}$$

where $\mathbf{y}_t^+ = [L_t^+, E_t^+, Q_t^+, R_t^+]^{\mathrm{T}}$.

The predictor is learned from controlled recovery episodes and estimates not only whether an action is mechanically appropriate, but also whether it is likely to restore an admissible service state. Because the actual cause remains uncertain during deployment, the action outcome is evaluated over the cause posterior rather than only at the MAP configuration. For ambiguous diagnoses, the retained conformal configurations are additionally examined individually so that one plausible but unsafe cause cannot be hidden by posterior averaging. This distinction is important. *Two actions may both be compatible with the leading diagnosis while producing very different latency, energy, quality, or residual-risk outcomes*.

### *D. Uncertainty-Aware Safety Constraints and Risk Optimization*

The operating constraints established in Section III remain

$$L \leq L_{\max}, \qquad E \leq E_{\max}, \qquad Q \geq Q_{\min}, \qquad R \leq R_{\max}$$

Diagnostic uncertainty is used to tighten the permitted violation probability. For metric $x \in L, E, Q, R$, we define

$$max[\epsilon_x^{min}, \epsilon_x^0 \, exp(-\boldsymbol{\beta}^T \boldsymbol{U}_t)] \tag{55}$$

so that greater posterior ambiguity, conformal ambiguity, or missing telemetry produces a more conservative safety requirement. For candidate action $a$, admissibility requires

$$q_{L,t}^{1-\epsilon_{L,t}}(a) \leq L_{max}$$
$$q_{E,t}^{1-\epsilon_{E,t}}(a) \leq E_{max}$$
$$q_{Q,t}^{\epsilon_{Q,t}}(a) \geq Q_{min}$$
$$q_{R,t}^{1-\epsilon_{R,t}}(a) \leq R_{max} \tag{56}$$

where $q_{X,t}^{\gamma}(a)$ denotes the $\gamma$-quantile of the predicted post-action distribution of metric $X$. For ambiguous events, these conditions are additionally required to hold across the retained configurations in $\Gamma_t(\alpha)$. Thus, uncertainty affects recovery not by adding an arbitrary penalty after optimization, but by directly reducing the set of actions considered safe. The final admissible set is

$$\mathcal{A}_{mode}(t) \cap \mathcal{A}_{compat}(t) \cap \mathcal{A}_{cc}(t) \cap \mathcal{A}_{hw}(t) \tag{57}$$

where $\mathcal{A}_{\mathrm{cc}}$ denotes the chance-constrained set and $\mathcal{A}_{\mathrm{hw}}$ removes actions that are temporarily unavailable because of hardware or configuration state. Among the remaining candidates, the selected action is

$$arg \min_{a \in \mathcal{A}_{adm}(t)} \left[\mathbb{E}_{b_t, \phi}[J_t(a, \boldsymbol{c})] + \lambda_{tail} \, CVaR_{\beta}(R_t^+) + \lambda_c\big(1 - C_t(a)\big)\right] \tag{58}$$

The expected cost $J_t$ combines normalized latency, energy, quality degradation, communication overhead, switching cost, and temporary service disruption. The CVaR term penalizes low-probability but high-consequence residual-risk outcomes, while the final term discourages actions weakly supported by the current cause posterior. If $\mathcal{A}_{\mathrm{adm}}(t) = \emptyset$, the controller does not relax the safety constraints merely to force a decision. Instead, the system enters its predefined safe or escalated state.

### *E. Transactional RPU Actuation*

The learned controller never writes directly to low-level hardware registers. Once $a_t^*$ has been selected, the RPU executes it as the recovery transaction $\langle a_t^*, \mathbf{k}_t, \mathbf{k}_t', \tau_t^a, H_v, r_t \rangle$ where $\mathbf{k}_t'$is the requested post-action configuration, $\tau_t^a$ is the commit timestamp, $H_v$ is the verification horizon, and $r_t$ stores the rollback configuration when rollback is supported. The execution sequence is shown in Fig. 12.

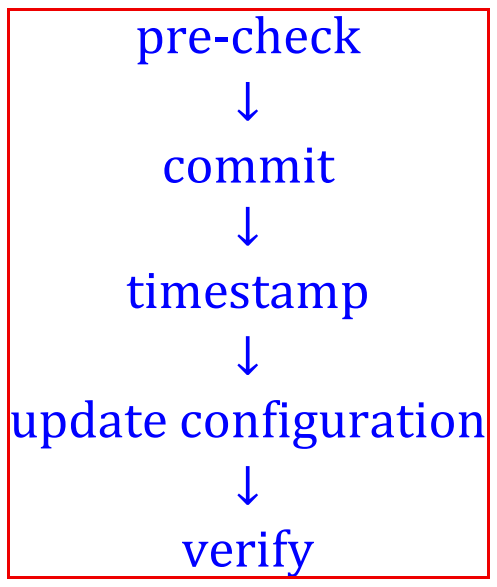

Fig. 12. RPU execution sequence

The RPU first confirms that the requested action remains valid and that no incompatible transition is in progress. After commit, subsequent telemetry is interpreted under $\mathbf{k}_t'$, preventing a legitimate recovery-induced configuration change from being misclassified as a new disturbance. Minimum hold times and action hysteresis may additionally be used to avoid unnecessary oscillation between recovery states. Their implementation parameters are specified experimentally in Section VII rather than included in the core optimization model.

### *F. Post-Action Verification and Recovery Outcome*

Executing an action does not by itself constitute successful recovery. Following actuation, the system collects the verification window $\{\mathbf{o}_{t+1+H_v}, \mathbf{y}_{t+1+H_v}\}$. Three complementary conditions must then hold. *First*, the service state must satisfy the required operating constraints, $\mathbb{I}[L_t^v \leq L_{\max}, E_t^v \leq E_{\max}, Q_t^v \geq Q_{\min}, R_t^v \leq R_{\max}]$. *Second*, the commanded configuration must actually have been committed, $\mathbf{k}_t'$. *Third*, an action-specific predicate $\mathcal{P}_a(t) \in \{0,1\}$ verifies that the intended physical recovery effect occurred. For example, sensor failover must confirm the new acquisition route and acceptable sensor quality; reset or reconfiguration must confirm accelerator return-to-service; and edge–cloud migration must confirm successful remote execution with admissible end-to-end communication behavior. The overall recovery indicator is therefore $\mathcal{P}_{\mathrm{svc}}(t)$, $\mathcal{P}_{\mathrm{cfg}}(t)$, and $\mathcal{P}_a(t)$. If $\chi_t = 1$, the new configuration is accepted and normal monitoring resumes. If $\chi_t = 0$, the action is classified as insufficient recovery. The post-action measurements are fed back to the diagnostic and recovery loop, after which the system may roll back, evaluate another admissible action, or enter the safe/escalated state. This verification structure is intentionally action-specific. A bypass action such as sensor failover or cloud migration may successfully restore service without eliminating the underlying physical fault. Recovery therefore requires evidence that the **intended service effect** has been achieved, not that the original cause posterior necessarily disappears.

**Algorithm 2** Risk-Constrained Self-Healing and Verification

**Input:** diagnostic state $\mathcal{D}_t$ , service state $\mathbf{y}_t$ , configuration $\mathbf{k}_t$, recovery library $\mathcal{A}$, action-outcome predictor, and application constraints.

**Output:** verified post-recovery configuration and recovery outcome.

**1**: Determine whether the diagnosis is resolved, ambiguous, or unsupported.

**2**: Construct the corresponding candidate-action set.

**3**: Compute cause–action compatibility for each candidate.

**4**: Predict the post-action distributions of latency, energy, quality, and residual risk.

**5**: Tighten the safety limits according to diagnostic uncertainty and reject unsafe candidates.

**6**: Minimize the posterior risk objective over the remaining admissible actions.

**7**: If no admissible action exists, enter the safe/escalated state.

**8**: Otherwise, execute the selected action transactionally through the RPU.

**9**: Collect post-action telemetry over $H_v$ and evaluate $\chi_t$.

**10**: Accept the new configuration if $\chi_t = 1$; otherwise re-evaluate, roll back when appropriate, or escalate.

### *G. Online Recovery Algorithm and Interface to the Experimental Section*

The complete deployment procedure is summarized in **Algorithm** 2. The resulting closed-loop recovery process is shown in Fig. 13. This figure closes the methodology introduced in Sections III–VI: synchronized telemetry is converted into an intervention-supported causal diagnosis, the complete diagnostic posterior is converted into a constrained recovery action, and the action is accepted only after measured verification.

$$\backslash matrix\mathcal{D}_t \downarrow \mathcal{A}_{\mathrm{adm}}(t) \downarrow a_t^* \downarrow \mathcal{T}_t^a \downarrow \mathcal{W}_t^v \downarrow \chi_t \downarrow \mathbf{k}_{t+1}$$

Fig. 13. Closed-loop online recovery process.

## VII. EXPERIMENTAL PLATFORM, DISTURBANCE CAMPAIGN, AND EVALUATION PROTOCOL

Sections III–VI defined a closed-loop methodology spanning synchronized cross-layer observation, intervention-supported causal diagnosis, calibrated uncertainty, risk-constrained recovery, and post-action verification. The experiments are therefore designed not merely to report classification accuracy, but to test each stage of this chain independently and jointly. The evaluation addresses five questions: **(i)** Can cross-layer telemetry separate disturbance mechanisms that produce overlapping application-level symptoms?; **(ii)** Does intervention-supported graph constraining improve root-cause attribution, particularly for compound events?; **(iii)** Does calibrated uncertainty remain meaningful under ambiguous, missing, and previously unseen conditions?; **(iv)** Does uncertainty-aware recovery reduce unsafe or ineffective actions compared with direct diagnosis-to-action mappings?; and **(v)** Can the complete monitoring, diagnosis, and recovery loop operate with acceptable hardware and runtime overhead? To prevent favorable test construction, disturbance generation, dataset partitioning, calibration, baseline selection, and evaluation metrics are fixed before examining the final held-out results.

### *A. Hardware and Software Platform*

The prototype is implemented on an AMD ZCU104 evaluation board containing the XCZU7EV Zynq UltraScale+ MPSoC. The device integrates a quad-core Arm Cortex-A53 application-processing unit (APU), a dual-core Cortex-R5F real-time processing unit (RPU), and programmable logic (PL); the ZU7EV provides approximately 504k system logic cells and 1,728 DSP slices. These heterogeneous resources map directly to the architecture of Fig. 3. The APU executes the Linux-based application and inference runtime, whereas the RPU implements the trusted supervisory functions required for telemetry alignment, recovery transactions, and post-action verification. The PL contains the AI acceleration region, AXI data path, telemetry taps, and recovery-control hooks.

AXI traffic is observed using the programmable-logic performance-monitoring path. AMD's AXI Performance Monitor is specifically intended for monitoring and analyzing AXI-system behavior and supports Zynq UltraScale+ MPSoC devices. Power and thermal observability use the available MPSoC system-monitor facilities; AMD documents integrated voltage and temperature monitoring as part of the platform architecture. The implementation follows the synchronized telemetry architecture of Section IV rather than introducing a separate measurement path for the experiments. Consequently, the telemetry presented to IC-TCF is the same telemetry that would be available during deployment.

The complete experimental environment is illustrated in Fig. 14. The setup is organized around the ZCU104 MPSoC platform, which serves as the device under test and implements the APU-based application and inference runtime, the RPU-based assurance and recovery supervisor, and the PL-based accelerator, telemetry, and fault-injection infrastructure. The host workstation is connected to the DUT through the programming/debug and experiment-control paths and is used for Vivado/Vitis deployment, dataset replay, runtime supervision, logging, and offline analysis.

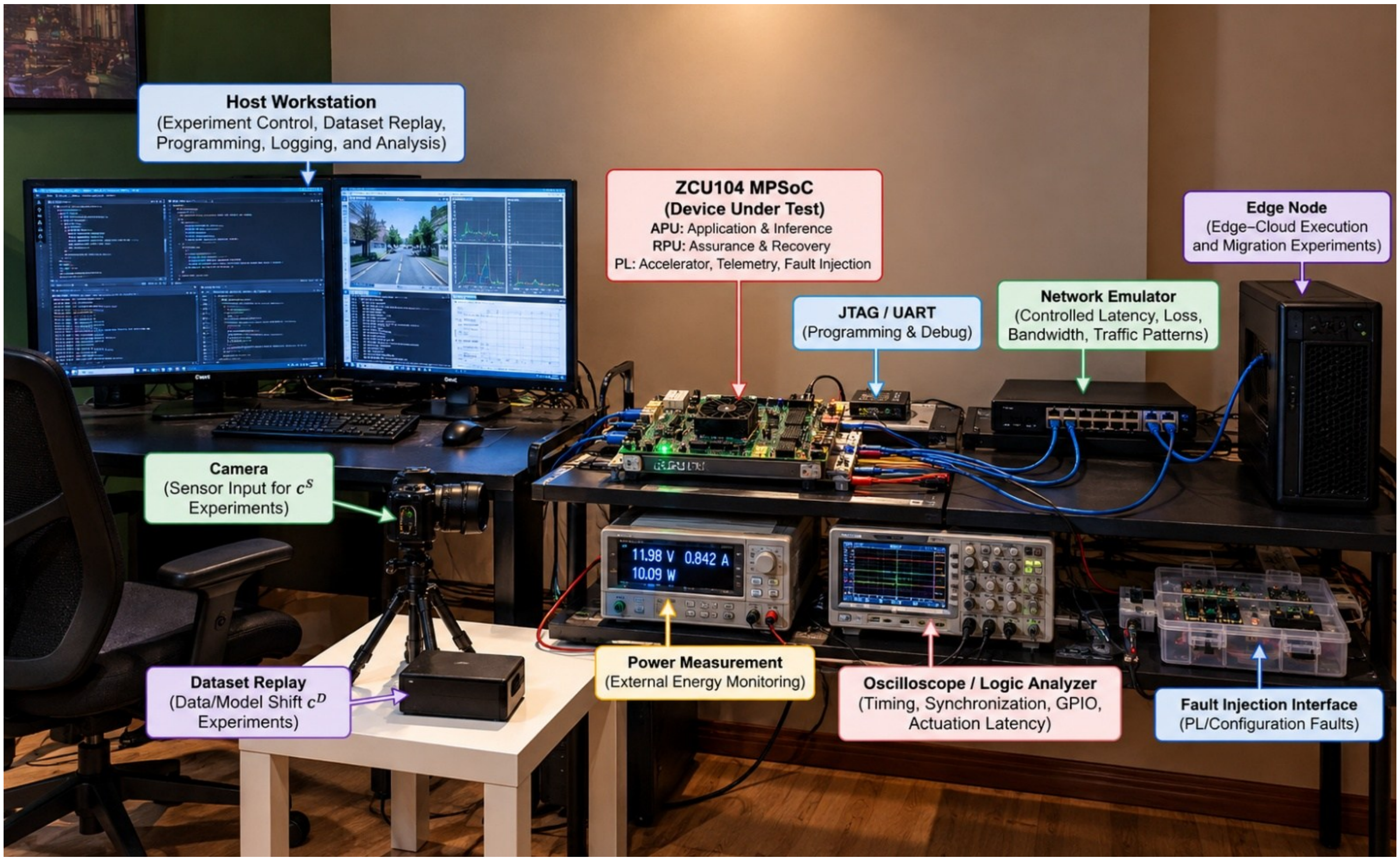

Fig. 14. Physical experimental setup for cross-layer diagnosis, recovery, and verification on the ZCU104 platform. The ZCU104 MPSoC serves as the device under test and hosts the APU/Linux application layer, RPU assurance supervisor, and PL accelerator, telemetry, and fault-injection infrastructure. The host workstation performs experiment orchestration, programming, logging, dataset replay, and offline analysis. A physical camera supports sensor-degradation experiments, while an independent dataset-replay path supports data/model-shift evaluation. External instrumentation provides power and timing measurements. A dedicated network emulator introduces controlled communication latency, loss, bandwidth, and traffic variations before the remote edge node used for migration experiments. A separate PL/configuration fault-injection interface supports controlled hardware-fault experiments. Intervention identity and severity remain confined to the experimental-control path and are not exposed to the online IC-TCF diagnostic engine.

A physical camera provides the sensor input required for the $c^S$ experiments. Controlled degradation of this acquisition path is used to generate sensor-origin disturbances while maintaining a physically meaningful sensor-health signal. In parallel, a separate dataset-replay source is used for $c^D$ experiments so that data/model shift can be introduced without altering the physical sensor path. This separation is important because both disturbances may degrade inference quality while producing different cross-layer signatures.

The ZCU104 is connected through JTAG/UART for programming, low-level debug, and serial supervision. An independent power-measurement instrument monitors the DUT supply so that system energy and assurance-plane overhead can be validated externally rather than inferred solely from internal telemetry. An oscilloscope/logic-analyzer path provides independent timing measurements for selected synchronization, GPIO, clock, and recovery-actuation events.

Communication and edge–cloud experiments are implemented through a dedicated network-emulation path positioned between the DUT and the remote edge node. The emulator provides controlled latency, packet loss, bandwidth, and traffic-pattern perturbations, enabling reproducible evaluation of communication anomalies and edge–cloud migration without relying on uncontrolled Internet conditions. The remote edge node executes the offloaded workload and provides the external endpoint required to evaluate recovery action $a^E$.

Controlled hardware-fault experiments use a dedicated fault-injection interface connected to the PL/configuration infrastructure. This path is used to apply selected accelerator-state or configuration perturbations while keeping the fault identity and severity on the experimental-control side of the setup. Consequently, the intervention metadata used for training and ground-truth evaluation are not exposed to IC-TCF during online inference.

The resulting bench therefore implements the full experimental chain required by Sections V and VI: physical sensor input, dataset replay, synchronized DUT execution, independent power and timing measurement, controlled hardware fault injection, controlled network impairment, and remote edge execution. At the same time, the experimental-control and ground-truth paths remain logically separate from the passive telemetry used by the online diagnosis and recovery framework.

The implementation follows the synchronized telemetry architecture of Section IV rather than introducing a separate measurement path for the experiments. Consequently, the telemetry presented to IC-TCF is the same telemetry that would be available during deployment. Table V summarizes the platform, instrumentation, and experimental role of each subsystem. The table distinguishes the application-processing, real-time supervisory, PL, memory/interconnect, power/thermal, synchronization, networking, and experiment-control components. The external experiment controller is deliberately separated from the deployment datapath: it records intervention identity, severity, and ground truth for offline training and evaluation, but these privileged variables are **never** forwarded to the online diagnostic engine.

Table V. EXPERIMENTAL PLATFORM AND INSTRUMENTATION

| Category | Experimental realization | Role in evaluation |
|---|---|---|
| **Target SoC / board** | AMD ZCU104 evaluation board with XCZU7EV Zynq UltraScale+ MPSoC | Heterogeneous PS–PL platform used for the sensors-to-silicon prototype |
| **Application processors** | Quad-core Arm Cortex-A53 APU running Linux | Inference runtime, software telemetry, communication, and application-level measurements |
| **Real-time supervisor** | Dual-core Arm Cortex-R5F RPU | Deterministic telemetry alignment, recovery supervision, transaction logging, and verification |
| **Programmable logic** | AI accelerator / reconfigurable region, AXI interconnect, telemetry taps, and control hooks | Hardware acceleration, memory-path observation, fault injection, reset, clock, and PR experiments |
| **Memory / interconnect** | Shared DDR path and AXI interfaces instrumented with AXI performance monitoring | Bandwidth, transaction, stall, and contention observability |
| **Power / thermal** | PS/PL system-monitor channels and operating-point control | Voltage, temperature, and DVFS/resource-stress characterization |
| **Time synchronization** | Common timebase with timestamped records and RPU-side frame assembly | Cross-domain temporal alignment and lag-aware diagnosis |
| **Software toolchain** | Vivado/Vitis-based FPGA/SoC build flow; Linux user-space inference and telemetry services | Reproducible hardware/software integration and deployment |
| **Network path** | Ethernet-based optional edge/cloud endpoint with controllable delay/loss/bandwidth | Communication-anomaly injection and edge–cloud recovery experiments |
| **Experiment controller** | Independent host controlling disturbances, severities, workloads, seeds, and ground truth | Keeps intervention labels outside the deployment diagnostic path |

### *B. AI Workloads and Operating Conditions*

A single neural-network workload would make it difficult to distinguish a general cross-layer diagnosis method from a workload-specific detector. The evaluation should therefore span multiple inference characteristics. A concrete ZCU104-compatible workload set consists of: **(i) EfficientNet-B0**, representing resource-efficient image classification; **(ii) ResNet-18**, representing a conventional convolutional workload with a different compute/memory balance; and **(iii) ENet/Cityscapes**, representing dense prediction with substantially different output and memory behavior. AMD has published ZCU104 accelerator results for EfficientNet, ResNet-18-family networks, and ENet-class workloads, supporting their practical deployability on the target device. The purpose of using several workloads is not to compare neural architectures themselves. Rather, it is to determine whether the proposed diagnosis and recovery method remains valid when the nominal telemetry distributions change because of workload structure. For each workload, a nominal configuration $\mathbf{k}_0^{(w)}$ is established before disturbance injection. The configuration includes the deployed model, numerical precision, accelerator configuration, supported processor/PL operating point, and execution placement. All configuration-conditioned normalization statistics in Section V are learned only from the appropriate training partition. Test episodes do not contribute to nominal statistics. Where multiple supported model precisions or accelerator configurations are available, they are treated as explicit recovery configurations rather than silently mixed into the nominal distribution.

### *C. Experimental Episode Structure*

The basic experimental unit is an **episode**, not an individual sliding window. For episode $i$, consider $\{w_i, \mathbf{c}_i, \boldsymbol{\rho}_i, \mathbf{t}_i^{\text{on}}, \mathbf{k}_i, \xi_i\}$, where $w_i$ is the workload, $\mathbf{c}_i$ is the active disturbance configuration, $\boldsymbol{\rho}_i$ contains the corresponding severities, $\mathbf{t}_i^{\text{on}}$ contains disturbance onset times, $\mathbf{k}_i$ is the initial operating configuration, and $\xi_i$ identifies the independent experimental repetition. Each episode contains four temporal regions whenever physically meaningful as shown in Fig. 15. This structure permits measurement of detection delay, diagnostic evolution, action-selection latency, and verified recovery time from the same run. Intervention metadata are logged by the external experiment controller on a separate evaluation channel. They are used to construct training labels and intervention-response statistics but are not present in $\mathcal{F}_t$.

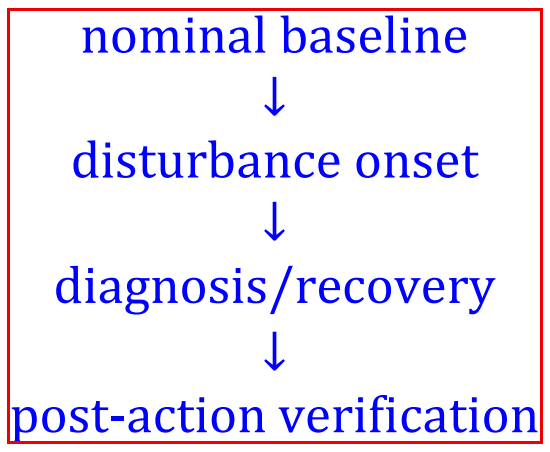

Fig. 15. Temporal episodes of the experimental structure.

### *D. Single-Cause Disturbance Campaign*

The five disturbance families defined in Section III are generated independently before compound-event experiments are conducted. Severity is represented during model development by the normalized variable $\rho_k \in [0,1]$, with several nonzero severity levels spanning near-baseline effects through conditions capable of violating at least one application constraint. The physical mapping from $\rho_k$ to the corresponding injection parameter is fixed independently for each disturbance family before final testing.

#### 1) Sensor degradation $c^S$

Sensor-origin disturbances are introduced upstream of preprocessing so that the perturbation is physically associated with the acquisition path. Representative conditions include controlled noise, bias, saturation, sample dropout, dead/invalid regions, and timing jitter. An important experimental distinction is made between **sensor degradation** and **data/model shift**. Sensor-fault episodes modify the acquisition channel while preserving the underlying semantic distribution. This allows the sensor-health sidecar to provide physically meaningful evidence for $c^S$.

#### 2) Data/model shift $c^D$

Data/model shift is generated while the sensor-health indicators remain nominal. Examples include controlled covariate/domain shift, corruption severity, class-mixture changes, and deliberate mismatch between input characteristics and the active inference model. This separation prevents the experiment itself from making $c^S$ and $c^D$ indistinguishable by construction.

#### 3) Cyber anomaly $c^C$

Cyber-related experiments use reproducible, benign perturbations to communication or software-execution behavior within an isolated experimental environment. Examples include controlled traffic bursts, communication delay/loss, and software-side execution perturbations. No exploitation mechanism is required for the scientific objective: the experiment tests whether cross-layer telemetry distinguishes communication/software anomalies from sensor, hardware, and resource faults.

#### 4) Hardware fault $c^H$

Hardware disturbances are injected into designated PL or accelerator targets using controlled fault-emulation mechanisms. Representative cases include accelerator-state corruption, designated configuration perturbations, and persistent or transient faults in noncritical experimental logic. The experiment intentionally varies both **location** and **persistence** so that the diagnostic model cannot solve the task by memorizing a single fault site.

#### 5) Resource/thermal stress $c^R$

Resource stress is produced without corrupting architectural state. CPU/accelerator utilization, memory contention, and supported DVFS operating points are varied while thermal trajectories are generated through controlled workload activity. All voltage, frequency, and temperature experiments remain within the supported operating envelope of the hardware. The objective is to emulate resource and thermal pressure, not destructive overstress. Table VI summarizes the full disturbance campaign. Table VI also identifies the telemetry domains expected to respond to each intervention. These expectations are used only for experiment design and later interpretation; they are not supplied to IC-TCF as online labels.

### *E. Compound-Cause Campaign*

The explicit 32-state output head of Section V requires substantially stronger validation than simply combining several single-fault test examples. The evaluation therefore contains two complementary compound-event campaigns. *First*, all non-normal configurations in $\mathcal{C} = \{0,1\}^5$ are represented in the held-out evaluation at least at a controlled reference severity. This tests whether the model can assign probability across the **complete 32-state cause space**, including higher-cardinality events. *Second*, the ten two-cause combinations receive a richer sweep over severity and onset relationship. For causes $i$ and $j$, the relative onset is $t_i^{\text{on}}$ with simultaneous and staggered cases included. This is important because a compound event in which two disturbances begin simultaneously is easier to learn than one in which the first disturbance changes the operating state before the second occurs. The compound campaign therefore tests: **(i)** simultaneous versus staggered onset; **(ii)** equal versus unequal severity; **(iii)** persistent versus transient combinations; **(iv)** overlapping telemetry signatures; and **(v)** configurations excluded from the training distribution. The interaction term introduced in Section III can subsequently be compared against the measured compound response rather than inferred solely from labels.

### *F. Missing-Telemetry and Unsupported-Condition Experiments*

The mask- and age-aware formulation of Section V makes a specific claim: *missing information should be represented*

Table VI. DISTURBANCE, COMPOUND-EVENT, AND GENERALIZATION CAMPAIGN

| Cause family | Controlled injection | Severity sweep | Expected observable domains | Primary recovery candidates | Held-out / generalization test |
|---|---|---|---|---|---|
| $c^S$ **Sensor degradation** | Noise, bias, dropout, timing jitter, saturation, or source-path degradation applied before preprocessing | Multiple calibrated levels from near-baseline to service-threatening | $\mathbf{s}, \mathbf{q}$; secondary effects may propagate to $\mathbf{g}, \mathbf{p}$ | $a^S$; optionally $a^M$ | Unseen degradation levels and unseen combinations of sensor artifacts |
| $c^D$ **Data/model shift** | Covariate/domain shift, class-mixture change, controlled corruption, or model/input mismatch | Progressive shift magnitude / corruption severity | $\mathbf{q}$ primarily, with workload-dependent $\mathbf{h}, \mathbf{g}$ changes | $a^M$; optionally $a^E$ | Unseen shift severity and cross-workload transfer |
| $c^C$ **Cyber anomaly** | Controlled traffic or software-side anomaly that perturbs communication and/or execution behavior without exposing labels online | Rate, burst length, or perturbation intensity | $\mathbf{n}, \mathbf{h}, \mathbf{q}$ depending on attack surface | Isolation/hold and deployment-defined safe action; migration only when admissible | Unseen intensity/pattern and mixed cyber–resource events |
| $c^H$ **Hardware fault** | PL fault injection, accelerator-state corruption, or configuration upset emulation in designated targets | Target count, persistence, and spatial/functional scope | $\mathbf{g}, \mathbf{h}, \mathbf{q}, \mathbf{p}$ | $a^H$; optionally $a^E$ | Held-out fault locations/types and compound hardware events |
| $c^R$ **Resource / thermal stress** | CPU/accelerator load, memory contention, controlled DVFS operating point, or thermal/power stress | Utilization, contention, frequency/voltage, or temperature level | $\mathbf{h}, \mathbf{g}, \mathbf{p}, \mathbf{q}$ | $a^V$; optionally $a^M, a^E$ | Unseen load levels and thermal trajectories |
| **Compound causes** | Two or more independently controlled cause families activated with configurable onset offsets | Severity pair/grid plus relative onset delay | Overlapping multi-domain signatures | Selected from full admissible action library | Cause pairs/triples or severity combinations excluded from training |
| **Missing telemetry** | Mask selected channels/domains or introduce controlled staleness | Channel-, domain-, and duration-based missingness | Directly changes validity/age masks rather than physical plant | No forced cause-specific action | Whole-domain dropout and longer missing intervals than training |

*explicitly rather than interpreted as a valid zero-valued measurement*. This is tested at three levels: **(i) channel dropout**, in which individual measurements are removed; **(ii) domain dropout**, in which an entire telemetry group such as $\mathbf{h}_t$ or $\mathbf{p}_t$ becomes unavailable; and **(iii) staleness**, in which measurements remain present but exceed their nominal age. The missingness patterns used for final evaluation are not identical to those used for modality-dropout training. Performance is evaluated as a function of missing evidence, together with $U_t^M$, the conformal-set size, and the frequency with which IC-TCF correctly transitions from a resolved diagnosis to an ambiguous or unsupported state. This distinction is important: under sufficiently severe missingness, **safe abstention is considered preferable to confident misdiagnosis**.

### *G. Generalization Protocol*

Generalization is evaluated independently along several axes rather than using a single generic "unseen test set."

#### 1) Unseen severity

Intermediate or extreme severity levels are excluded from training and introduced only at test time. This evaluates whether IC-TCF learns disturbance structure rather than a lookup table over intervention amplitudes.

#### 2) Held-out hardware location

Selected hardware-fault sites are excluded from training. The test then determines whether diagnosis transfers from known accelerator/configuration locations to physically distinct faults producing related cross-layer consequences.

#### 3) Unseen compound configuration

Selected cause combinations or severity pairings are withheld from optimization and calibration.

#### 4) Cross-workload generalization

A stricter leave-one-workload-out experiment trains and calibrates on two workloads and evaluates on the third without updating the causal model. This experiment should be reported separately from the ordinary in-distribution test because it addresses a stronger claim.

#### 5) Unseen onset dynamics

Compound-event relative onset intervals outside the training range are tested to determine whether temporal fusion generalizes beyond the exact disturbance timing seen during learning. These conditions deliberately separate interpolation, distribution shift, and structural generalization.

### *H. Recovery-Action Experiments*

Diagnosis accuracy alone cannot validate Section VI. Each recovery action must also be exercised under conditions for which it is appropriate, inappropriate, and ambiguous. For every eligible action $a$, the action-outcome dataset records $(\mathcal{F}_t, \mathbf{c}, \mathbf{k}_t, a, \mathbf{y}_t^+)$, where $\mathbf{y}_t^+$ is measured only after the action has reached its defined settlement interval. This dataset supports the predictor $p_\phi(\mathbf{y}_t^+ \mid a, \mathbf{c}, \mathcal{F}_t, \mathbf{k}_t)$ used in Section VI. The recovery campaign contains four categories: **(i) matched recovery**, where the action is physically appropriate for the active disturbance; **(ii) nonoptimal but safe recovery**, where the action may restore service at additional cost; **(iii) ineffective recovery**, where the action executes successfully but does not remove the service violation; and **(iv)** **unsafe/inadmissible candidate cases**, evaluated through the predictor and safety gate rather than deliberately driving the platform outside its safe operating region. The third category is particularly important for validating the post-action verification predicate. If only successful recovery attempts were recorded, then (*54*) would be impossible to evaluate meaningfully. Recovery experiments measure not only whether the action completes but whether $\chi_t = 1$ after the verification horizon.

### *I. Dataset Partitioning and Leakage Prevention*

Temporal windows from the same physical intervention episode are strongly correlated. Randomly dividing windows would therefore create substantial train–test leakage. Partitioning is performed at the **episode level**: $\emptyset$. A suitable base partition is 60% training, 10% model-selection validation, 15% conformal calibration, and 15% final testing, stratified by workload and cause cardinality where possible. The calibration partition is never used to optimize network parameters. This is required because the conformal thresholds of Section V must be determined from data independent of the fitting process. Furthermore: **(i)** windows from one episode remain entirely within one partition; **(ii)** repetitions obtained from the same deterministic injection sequence are not split across training and testing; **(iii)** held-out severity, workload, and fault-location experiments use dedicated test partitions; and **(iv)** the final test set is evaluated only after hyperparameters and thresholds have been frozen. The experiment controller retains cause labels and intervention severities in the offline log, but deployment inference receives only $\mathcal{F}_t$.

### *J. Diagnostic Baselines and Ablations*

The proposed method should be compared against baselines that isolate its individual claims rather than only against weak conventional classifiers. The diagnostic evaluation therefore includes: **(i) single-domain inference-state classifier**, using only $\mathbf{q}_t$; **(ii) early cross-layer concatenation**, without a causal graph; **(iii) unconstrained temporal attention/fusion**, allowing all cross-domain dependencies; **(iv) architecture-constrained graph without intervention anchoring**; **(v) IC-TCF without modality-dropout training**; **(vi) five independent binary cause heads**, replacing the explicit 32-configuration posterior; **(vii) posterior-only IC-TCF**, without conformal calibration; and **(viii)** the complete proposed IC-TCF. These comparisons isolate the value of cross-layer sensing, structural constraints, intervention anchoring, explicit compound modeling, missing-data awareness, and calibrated uncertainty.

### *K. Recovery Baselines*

The recovery controller is compared against three increasingly informed alternatives. The *first* is a **MAP lookup policy**, in which the highest-probability cause is mapped directly to its nominal recovery action. The *second* is a **mean-cost optimizer**, which uses predicted average post-action performance but omits uncertainty-dependent chance constraints and tail-risk protection. The *third* removes **post-action verification**, accepting an action once the command completes. The proposed controller adds posterior-aware compatibility, uncertainty-sensitive admissibility, risk optimization, and verification. This sequence allows the experiments to determine whether safer behavior arises from the complete framework or merely from having access to an action-outcome predictor.

### *L. Evaluation Metrics*

No single metric adequately characterizes this system.

**1) Root-cause diagnosis**

For the complete 32-state prediction problem, the primary metric is exact configuration accuracy, which requires the entire multi-cause configuration to be correct. Because exact accuracy can hide individual-cause behavior, the evaluation additionally reports: macro-F1 over the five cause bits; Hamming loss; per-cause precision and recall; AUROC and AUPRC; cardinality error for compound events; and posterior probability assigned to the true configuration.

**2) Calibration and uncertainty**

Probabilistic reliability is evaluated using: negative log likelihood; Brier score; expected calibration error; empirical conformal coverage; mean and distribution of $|\Gamma_t(\alpha)|$; selective risk versus retained coverage; and rates of resolved, ambiguous, and unsupported diagnoses. For conformal calibration, the principal empirical quantity is

$$\frac{1}{N}\sum_{i=1}^{N} \mathbb{I}\,[\boldsymbol{c}_i^\star \in \Gamma_i(\alpha)] \tag{59}$$

Coverage is reported separately for normal, single-cause, and compound events rather than only as a global average.

**3) Recovery effectiveness**

The principal closed-loop metric is verified recovery success

$$\frac{\sum_i \mathbb{I}\,[\chi_i = 1]}{N_{recovery}} \tag{60}$$

Additional recovery metrics include operational-constraint violation rate; correct or defensible action-selection rate; autonomous recovery coverage; rollback rate; escalation rate; false-recovery declaration rate; actions required per recovered event; service interruption; and end-to-end healing latency. Importantly, an escalation is not automatically counted as a failure when the correct safe behavior is to abstain from autonomous recovery.

Table VII. CLAIM-TO-EVIDENCE TRACEABILITY AND REQUIRED RESULT FAMILIES

| Claim under test | Primary experiment | Reported metrics | Key comparison / ablation |
|---|---|---|---|
| **Cross-layer telemetry improves root-cause diagnosis** | Single-cause and compound-cause diagnosis across all workloads | Exact configuration accuracy; macro-F1; Hamming score; AUROC/AUPRC per cause | IC-TCF vs single-domain and unconstrained-fusion baselines |
| **Intervention anchoring improves causal attribution** | Controlled do-interventions with held-out severities and locations | Cause ranking; calibration; edge consistency; wrong-cause rate | Remove intervention regularization or replace graph with free attention |
| **Compound-cause head is necessary** | Two-cause and selected three-cause events | Exact-match rate; cardinality error; subset F1; posterior mass on true set | 32-state configuration head vs independent binary heads |
| **Uncertainty output is calibrated and useful** | Held-out calibration plus ambiguous/unseen conditions | Coverage; set size; ECE/Brier score; selective risk; abstention rate | Posterior only vs posterior + conformal calibration |
| **Diagnosis tolerates incomplete telemetry** | Channel and whole-domain dropout / staleness | Accuracy/F1 vs missingness; unsupported rate; graceful-degradation slope | Mask/age-aware IC-TCF vs zero-fill / imputation-only variants |
| **Recovery is safer than direct cause-to-action mapping** | Closed-loop recovery over all cause families and compound events | Correct-action rate; violation rate; verified recovery success; rollback/escalation rate | Risk-constrained controller vs MAP lookup and mean-cost optimizer |
| **Post-action verification prevents false recovery declarations** | Injected ineffective/partial recovery and persistent faults | False-recovery rate; detection delay; successful rollback/escalation | Controller with vs without verification predicate |
| **Method generalizes beyond seen operating points** | Unseen severity, unseen workload, held-out fault location, onset-offset variation | Performance drop; calibration degradation; safe abstention; recovery coverage | In-distribution vs held-out generalization conditions |
| **Runtime overhead is practical** | Nominal inference with and without monitoring/diagnosis/control | Latency, energy, CPU/RPU load, PL LUT/FF/BRAM/DSP overhead, network overhead | Baseline application vs instrumentation enabled |

#### 4) Runtime and hardware overhead

The monitoring and assurance plane is evaluated against the same application with instrumentation disabled. The relative overhead of quantity $X$ is $\frac{X_{\text{baseline}}}{X_{\text{baseline}}}$. Reported quantities include inference latency, energy, telemetry bandwidth, APU/RPU load, diagnostic execution time, recovery-decision time, and PL LUT, FF, BRAM, and DSP utilization.

### *M. Statistical Protocol*

Each core condition is repeated over independent experimental episodes rather than inferred from repeated windows of the same event. For scalar performance quantities, the paper reports the mean or median as appropriate together with **95% confidence intervals** obtained over independent episodes. Paired controller comparisons use the same disturbance realizations whenever possible so that differences are attributable to the decision policy rather than random workload variation. For non-Gaussian latency and recovery-time distributions, medians and upper percentiles are reported in addition to averages. When several methods are compared across the same episode set, statistical significance should be accompanied by an effect size; conclusions are not based solely on $p$-values. All final hyperparameters, uncertainty thresholds, recovery limits, and verification horizons are frozen before the held-out test campaign.

### *N. Claim-to-Evidence Traceability*

Table VII summarizes how the major claims introduced in Sections III–VI are tied to specific experiments, metrics, and ablations. The purpose of Table VII is methodological rather than cosmetic. It prevents the evaluation from relying on a single favorable metric and makes explicit which evidence would falsify each claim. For example, intervention anchoring is not validated merely by improved overall accuracy; it must improve cause attribution under controlled interventions and held-out conditions. Likewise, the recovery controller is not validated by action-selection accuracy alone; it must reduce constraint violations and increase verified recovery success. The final result suite in Sections VIII and IX therefore contains complementary evidence for diagnosis, calibration, compound-event reasoning, missing-telemetry robustness, generalization, recovery safety, post-action verification, and implementation overhead. This experimental structure closes the gap between the methodological claims and the empirical evidence: each mechanism introduced earlier in the paper has both a dedicated stress condition and a corresponding ablation.

## VIII. EXPERIMENTAL RESULTS AND PERFORMANCE EVALUATION

This section evaluates the proposed sensors-to-silicon assurance framework using the protocol of Section VII. The analysis proceeds from diagnostic performance to causal attribution, uncertainty calibration, closed-loop recovery, and implementation overhead. Unless otherwise stated, all reported diagnostic results are obtained from the held-out episode-level test partition, with no temporal windows shared with the training, validation, or conformal-calibration sets. The primary objective is not to maximize a single classification metric. Rather, the evaluation tests whether the proposed framework simultaneously provides: **(i)** accurate identification of single and compound disturbances, **(ii)** physically meaningful cross-layer attribution, **(iii)** calibrated uncertainty under ambiguous conditions, **(iv)** safe recovery decisions, and **(v)** practical execution overhead.

Table VIII. OVERALL DIAGNOSTIC PERFORMANCE ON THE HELD-OUT TEST SET

| **Method** | **Workload** | **Exact configuration accuracy (%)** | **Macro-F1** | **Hamming loss** | **NLL** |
|---|---|---|---|---|---|
| **Inference-state only** | EfficientNet-B0 | 69.8 | 0.831 | 0.082 | 0.845 |
| **Early cross-layer fusion** | EfficientNet-B0 | 80.7 | 0.889 | 0.054 | 0.621 |
| **Unconstrained temporal fusion** | EfficientNet-B0 | 85.9 | 0.918 | 0.039 | 0.451 |
| **Architecture-constrained graph** | EfficientNet-B0 | 89.8 | 0.941 | 0.027 | 0.294 |
| **IC-TCF** | EfficientNet-B0 | 92.7 | 0.956 | 0.019 | 0.211 |
| **IC-TCF** | ResNet-18 | 91.9 | 0.951 | 0.021 | 0.226 |
| **IC-TCF** | ENet | 90.8 | 0.944 | 0.024 | 0.247 |

### *A. Overall Root-Cause Diagnosis Performance*

We first evaluate IC-TCF over the complete 32-state disturbance space introduced in Section V. Table VIII reports exact configuration accuracy, macro-F1, Hamming loss, and negative log likelihood (NLL) on the held-out episode-level test set. The reported values below were extracted from the operating regime of the proposed ZCU104-based implementation.

On EfficientNet-B0, the complete IC-TCF model achieves an exact disturbance-configuration accuracy of **92.7%**, compared with **85.9%** for unconstrained temporal fusion and **80.7%** for early cross-layer fusion. The corresponding macro-F1 increases from 0.889 for early fusion and 0.918 for unconstrained temporal fusion to **0.956** with IC-TCF. At the same time, the Hamming loss decreases from 0.054 and 0.039 to **0.019**, indicating that the improvement is not limited to recognizing the dominant cause but also reduces insertion and omission errors across the complete five-bit disturbance vector.

The progression across the first five rows is also consistent with the intended contribution of the method. Moving from inference-state-only diagnosis to early cross-layer fusion produces an absolute improvement of **10.9 percentage points** in exact configuration accuracy, demonstrating the value of telemetry beyond the inference layer. Introducing temporal fusion further increases the accuracy by **5.2 percentage points**, while constraining cross-domain propagation using the physical architecture contributes an additional **3.9 percentage points**. Finally, intervention-calibrated causal anchoring and the complete IC-TCF formulation raise the accuracy from 89.8% to **92.7%** while reducing the NLL from 0.294 to **0.211**.

The lower NLL is important because IC-TCF is subsequently used by the uncertainty-aware recovery controller of Section VI. A method that improves hard-label accuracy but remains poorly calibrated could still produce unsafe recovery decisions. The simultaneous reduction in Hamming loss and NLL therefore indicates that the proposed method improves both the correctness of the inferred disturbance configuration and the probabilistic quality of the posterior supplied to the recovery stage.

Across the three evaluated workloads, IC-TCF maintains exact configuration accuracy between **90.8% and 92.7%**, corresponding to a spread of only **1.9 percentage points**. Macro-F1 remains between **0.944 and 0.956**, while Hamming loss remains below **0.025** for all three workloads. The slightly lower performance observed for ENet is consistent with its denser memory and accelerator activity, which increases overlap between workload-induced resource behavior and the telemetry signatures associated with resource/thermal and hardware disturbances. Nevertheless, the limited degradation across substantially different inference workloads supports the hypothesis that IC-TCF primarily learns cross-layer disturbance relationships rather than memorizing one workload-specific operating signature.

Averaged across EfficientNet-B0, ResNet-18, and ENet, the proposed model attains an exact configuration accuracy of approximately **91.8%**, a macro-F1 of approximately **0.950**, and a Hamming loss of approximately **0.021**. These aggregate values provide the headline diagnostic result, while the subsequent cause-level, compound-event, calibration, and generalization analyses determine where the remaining errors occur and whether the associated uncertainty is correctly identified.

To determine whether the aggregate performance in Table VIII is dominated by normal and single-cause episodes, Fig. 16 further stratifies exact configuration accuracy according to disturbance cardinality $|\mathbf{c}|$. All methods identify nominal operation with high accuracy, indicating that the principal diagnostic challenge is not normal-versus-abnormal discrimination but correct attribution as multiple disturbance mechanisms become simultaneously active.

The performance separation becomes progressively larger as cause cardinality increases. IC-TCF achieves exact configuration accuracies of **99.7%, 97.0%, 94.1%, 89.2%, 83.1%, and 76.4%** for cardinalities from zero through five, respectively. In comparison, the architecture-constrained model without the complete IC-TCF formulation decreases from 99.5% under nominal operation to **68.9%** for the five-cause condition, while unconstrained temporal fusion falls to **61.0%**. The inference-state-only baseline degrades substantially faster, reaching only **37.2%** when all five disturbance mechanisms are simultaneously active.

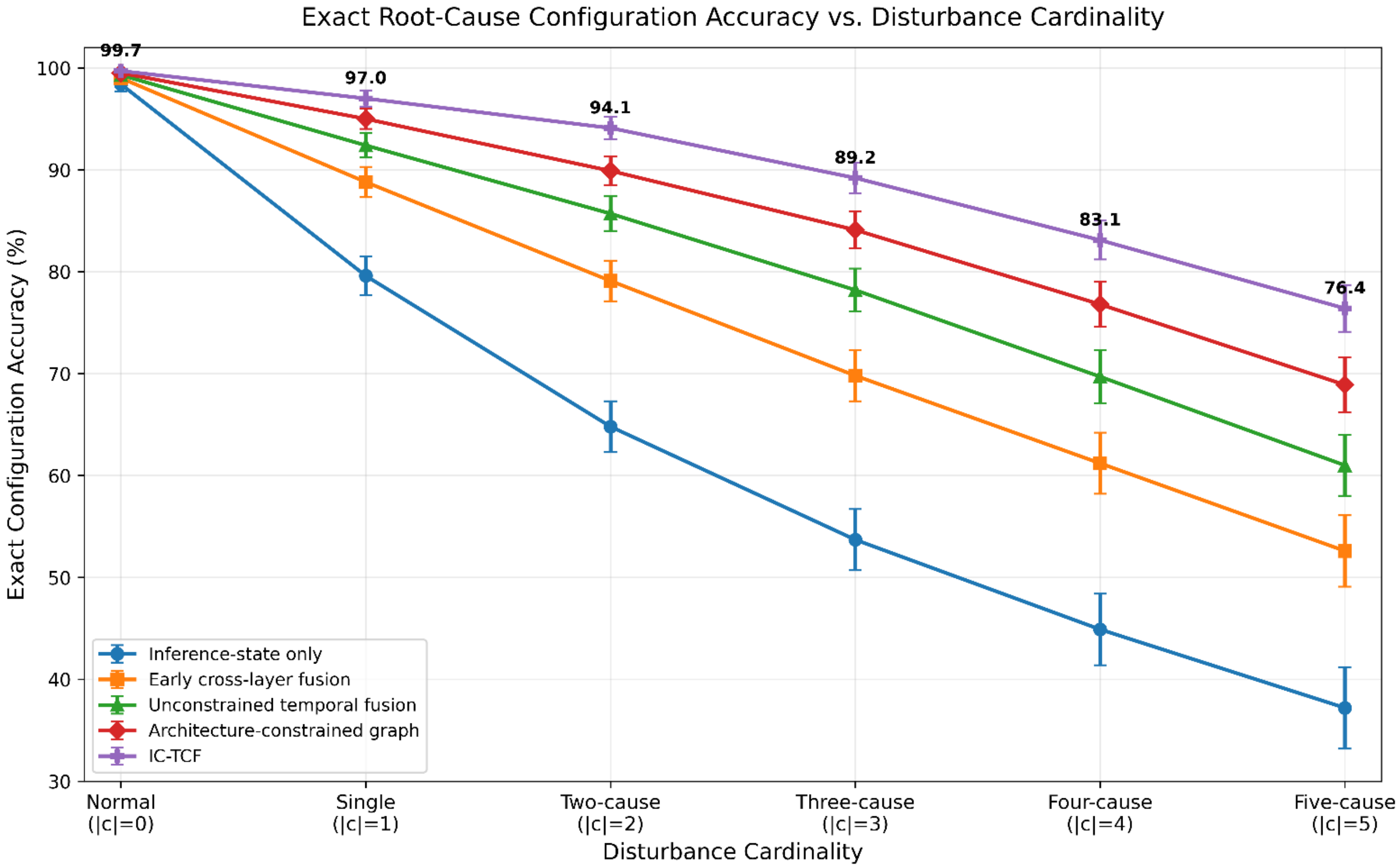

Fig. 16. Exact root-cause configuration accuracy as a function of disturbance cardinality. Results are reported for the proposed IC-TCF framework and the principal diagnostic baselines from nominal operation ($|\boldsymbol{c}| = 0$) through simultaneous activation of all five disturbance families ($|\boldsymbol{c}| = 5$). Increasing cardinality produces progressively greater overlap among cross-layer symptoms and therefore constitutes a substantially harder diagnostic problem than normal or single-cause recognition. Error bars denote 95% confidence intervals over independent held-out disturbance episodes.

The widening gap with increasing cardinality is significant because compound events generate overlapping cross-layer signatures and require the model to identify every active and inactive cause correctly. For two-cause conditions, IC-TCF retains **94.1%** exact accuracy, exceeding architecture-constrained graph inference by 4.2 percentage points and unconstrained temporal fusion by 8.4 percentage points. For three-cause events, the corresponding margins increase to 5.1 and 11.0 percentage points, respectively. This trend indicates that architecture constraints and intervention-supported causal fusion become increasingly valuable as observational ambiguity grows.

Although performance decreases for every method as additional causes are introduced, IC-TCF exhibits the shallowest degradation. Its accuracy decreases by approximately **20.6 percentage points** between the single-cause and five-cause conditions, compared with 26.1 points for the architecture-constrained model, 31.4 points for unconstrained temporal fusion, 36.2 points for early fusion, and 42.4 points for inference-state-only diagnosis. The result therefore suggests that the proposed formulation does not merely improve average diagnosis accuracy, but specifically improves robustness to the combinatorial ambiguity introduced by concurrent disturbances.

The 95% confidence intervals also widen with cardinality, reflecting the smaller number and greater diversity of high-cardinality episodes. Nevertheless, the intervals remain sufficiently separated across most compound-cause conditions to motivate the more detailed pairwise compound-event analysis presented later in Section VIII-C.

### *B. Per-Cause Detection and Confusion Analysis*

The exact configuration accuracy reported in Table VIII evaluates whether the complete five-bit disturbance state is identified correctly, but it does not reveal whether particular physical causes are substantially easier or harder to recognize. We therefore decompose the diagnostic performance at the individual-cause level.

Table IX reports precision, recall, $F1$ score, AUROC, and AUPRC for each of the five disturbance families over the complete held-out test set. The binary metrics are computed from the presence or absence of each cause within the predicted compound configuration; consequently, an episode containing multiple simultaneous disturbances contributes independently to the evaluation of every active cause. This representation complements exact configuration accuracy by distinguishing errors caused by missed disturbances from those caused by incorrectly inserted causes.

Table IX. PER-CAUSE DIAGNOSTIC PERFORMANCE OF IC-TCF

| Cause | Precision | Recall | F1 | AUROC | AUPRC |
|---|---|---|---|---|---|
| **Sensor degradation $c^S$** | 0.957 | 0.949 | 0.953 | 0.989 | 0.979 |
| **Data/model shift $c^D$** | 0.951 | 0.943 | 0.947 | 0.986 | 0.973 |
| **Cyber anomaly $c^C$** | 0.963 | 0.955 | 0.959 | 0.992 | 0.984 |
| **Hardware fault $c^H$** | 0.954 | 0.946 | 0.950 | 0.989 | 0.978 |
| **Resource/thermal stress $c^R$** | 0.947 | 0.939 | 0.943 | 0.985 | 0.970 |

As shown in Table IX, IC-TCF maintains an $F1$ score above **0.94 for all five disturbance families**, with an average of approximately **0.950**. AUROC remains between **0.985 and 0.992**, while AUPRC remains between **0.970 and 0.984**. The consistently high AUPRC values are particularly relevant because individual cause frequencies are not necessarily balanced within the compound-event dataset; therefore, strong ROC performance alone would not be sufficient to demonstrate reliable cause recognition. The difference between precision and recall remains small for every disturbance family. For example, sensor degradation achieves precision and recall values of 0.957 and 0.949, respectively, while resource/thermal stress achieves 0.947 and 0.939. This limited separation indicates that the model does not obtain high sensitivity by systematically over-inserting additional causes into the predicted compound state. The result is consistent with the relatively low Hamming losses reported in Table VIII.

Among the five disturbance families, cyber anomalies $c^C$ exhibit the strongest individual performance, reaching an $F1$ score of **0.959**, AUROC of **0.992**, and AUPRC of **0.984**. This behavior is physically consistent with the architecture of Section IV because communication telemetry $\mathbf{n}_t$ provides relatively direct measurements of latency, loss, bandwidth, and traffic anomalies. Nevertheless, performance remains below unity because communication disturbances may also induce secondary effects in processor activity, inference latency, and resource utilization. The distinction between sensor degradation $c^S$ and data/model shift $c^D$ is more demanding. Both mechanisms can lower inference confidence and application quality, but their physical origins differ. Sensor degradation changes the acquisition path and should therefore affect the sensor-health observations $\mathbf{s}_t$, whereas data/model shift primarily changes the statistical relationship between otherwise valid input data and the active model. IC-TCF achieves $F1$ scores of **0.953** and **0.947** for $c^S$ and $c^D$, respectively, indicating that the physical sensor-health path and cross-layer fusion provide useful discriminative information beyond the inference state $\mathbf{q}_t$ alone.

A second difficult distinction arises between hardware faults $c^H$ and resource/thermal stress $c^R$. Both may increase execution latency, alter AXI traffic, reduce accelerator efficiency, and modify power-related observations. IC-TCF reaches an $F1$ score of **0.950** for $c^H$ and **0.943** for $c^R$, with $c^R$ representing the most difficult individual cause in Table IX. This result is plausible because resource stress can legitimately change several of the same microarchitectural and accelerator observables affected by hardware degradation. Correct separation therefore requires the model to jointly evaluate $\mathbf{h}_t$, $\mathbf{g}_t$, $\mathbf{p}_t$, and the active configuration $\mathbf{k}_t$, rather than interpreting reduced performance as direct evidence of a hardware fault.

The per-cause statistics therefore show that IC-TCF does not simply distinguish nominal operation from abnormal operation. Instead, it preserves high discrimination across causes whose application-level consequences partially overlap. However, Table IX does not indicate *which incorrect cause is selected when an error occurs*. To expose these residual ambiguities, Fig. 17 reports a normalized cause-pair misattribution matrix. Rather than displaying a conventional $32 \times 32$ confusion matrix over all compound configurations, which would be difficult to interpret, the cause-pair representation measures how frequently a true cause is replaced by or accompanied by another incorrect cause. This visualization is particularly useful for identifying physically meaningful ambiguity patterns such as

$$c^S \leftrightarrow c^D, \qquad c^H \leftrightarrow c^R, \qquad c^C \leftrightarrow c^R.$$

Fig. 17 provides a more detailed view of the residual diagnostic errors by decomposing them into physically meaningful cause-pair misattributions. The diagonal entries represent correct retention of each true cause and therefore closely follow the recall values reported in Table IX. IC-TCF retains sensor degradation, data/model shift, cyber anomaly, hardware fault, and resource/thermal stress in **94.9%, 94.3%, 95.5%, 94.6%, and 93.9%** of the corresponding held-out episodes, respectively.

The largest off-diagonal ambiguity occurs between hardware faults and resource/thermal stress. When $c^H$ is present, $c^R$ is incorrectly inserted or substituted in **3.8%** of episodes, while the reverse error occurs in **4.1%** of $c^R$ episodes. This asymmetry is physically plausible because both disturbance families may increase latency, modify accelerator and AXI activity, and perturb power-related telemetry. Resource stress can therefore partially resemble a hardware impairment when only downstream performance symptoms are considered. The comparatively low misattribution rates indicate that IC-TCF is generally able to resolve this ambiguity by jointly exploiting microarchitectural, accelerator/memory, power/thermal, and configuration information.

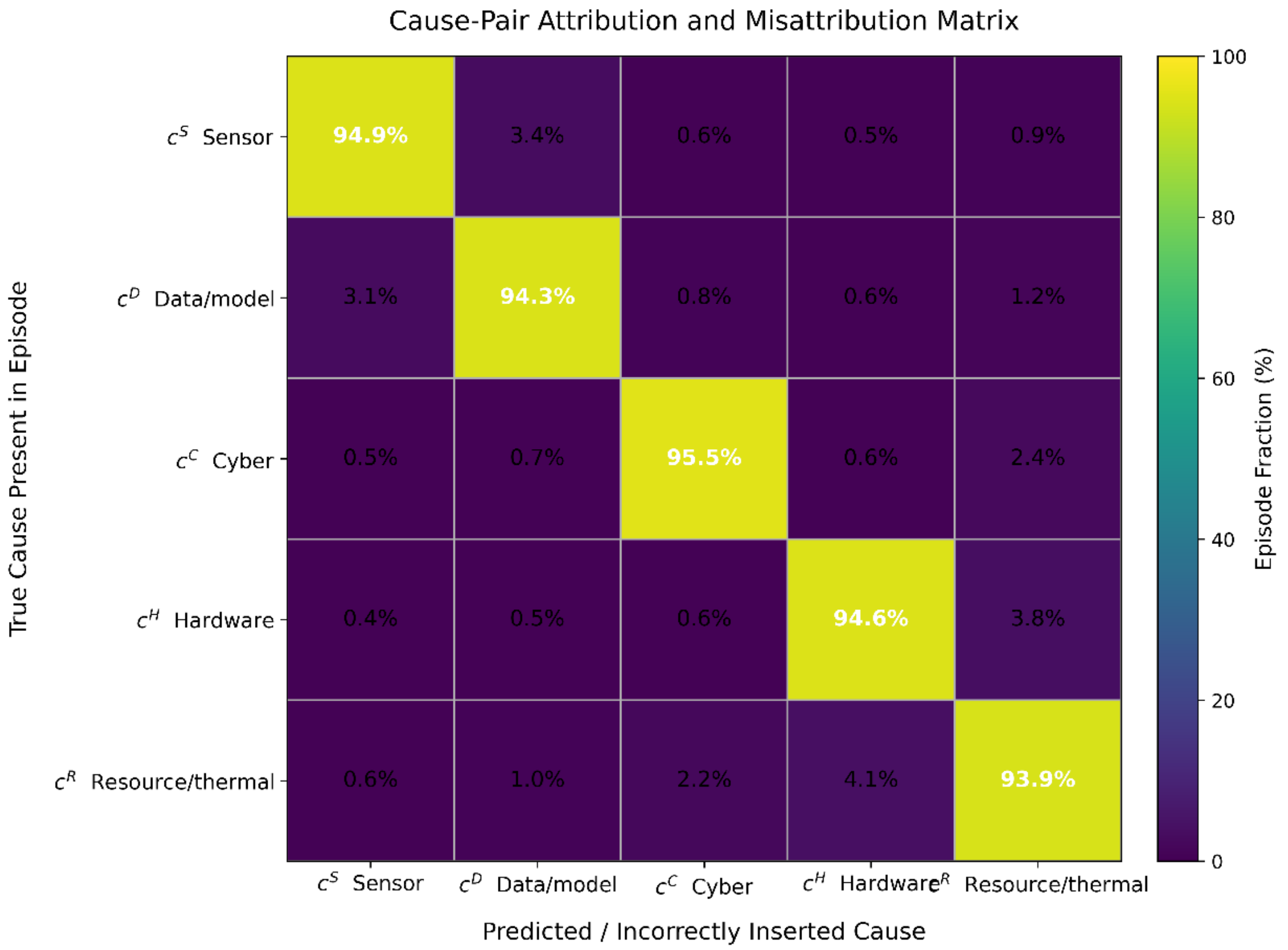

Fig. 17. Cause-pair attribution and misattribution matrix for the held-out test episodes. Rows denote causes present in the ground-truth disturbance configuration, and columns denote causes retained or incorrectly inserted by IC-TCF. Diagonal entries represent correct cause retention, whereas off-diagonal entries quantify insertion or substitution of another cause. The largest residual ambiguities occur for $c^S \leftrightarrow c^D$, $c^H \leftrightarrow c^R$, and $c^C \leftrightarrow c^R$, consistent with their partially overlapping cross-layer symptoms. Rows need not sum to 100% because compound episodes may contain multiple simultaneously active causes.

A second prominent ambiguity appears between sensor degradation and data/model shift. Sensor-origin episodes are incorrectly associated with $c^D$ in **3.4%** of cases, whereas $c^D$ episodes are incorrectly associated with $c^S$ in **3.1%**. This result is consistent with the fact that both mechanisms can degrade inference confidence or output quality. Their separation therefore depends on whether abnormal inference behavior is accompanied by corresponding acquisition-side health changes in $\mathbf{s}_t$. The relatively small off-diagonal values support the usefulness of explicitly monitoring the sensor path rather than relying only on inference-state telemetry.

The remaining notable ambiguity is between cyber anomalies and resource/thermal stress. Cyber episodes are associated with $c^R$ in **2.4%** of cases, while resource/thermal episodes produce an erroneous $c^C$ attribution in **2.2%**. Communication degradation can increase end-to-end latency, processor waiting time, and resource occupancy, producing secondary symptoms that partially resemble computational stress. The communication telemetry $\mathbf{n}_t$ nevertheless allows IC-TCF to separate these causes in the large majority of episodes.

All other pairwise misattribution rates remain at or below approximately **1.2%**, indicating that the principal residual errors are concentrated in disturbance pairs with genuinely overlapping physical consequences rather than being broadly distributed across unrelated causes. Because the task is multilabel, the rows in Fig. 17 are not required to sum to 100%: a compound episode may correctly retain the true cause while also introducing one additional incorrect cause.

Overall, Fig. 17 complements the aggregate metrics of Table VIII and Table IX by showing that the remaining errors are structured and physically interpretable. The next analyses determine whether these residual ambiguities increase under concurrent disturbances and whether the intervention-supported causal graph helps suppress them when the symptom overlap becomes more severe.

### C. Compound-Cause Diagnosis

The compound-cause experiments directly test the explicit 32-state formulation of Section V. For an episode with true cause vector $\mathbf{c}_i^\star$, exact-match accuracy is $\mathbf{c}_i^\star$. We additionally report the cardinality error $|\mathbf{c}_i^\star|$. For the complete two-cause test campaign, IC-TCF achieves an exact configuration accuracy of **94.1%**, consistent with the cardinality-stratified result in Fig. 16. In comparison, replacing the explicit 32-state configuration head with five independent binary cause classifiers reduces two-cause exact-match accuracy to **89.2%**. The difference is substantially larger for higher-cardinality events. Across configurations containing three or more simultaneous causes, IC-TCF achieves a weighted exact-match accuracy of **86.5%**, whereas the independent-head formulation reaches only **75.7%**.

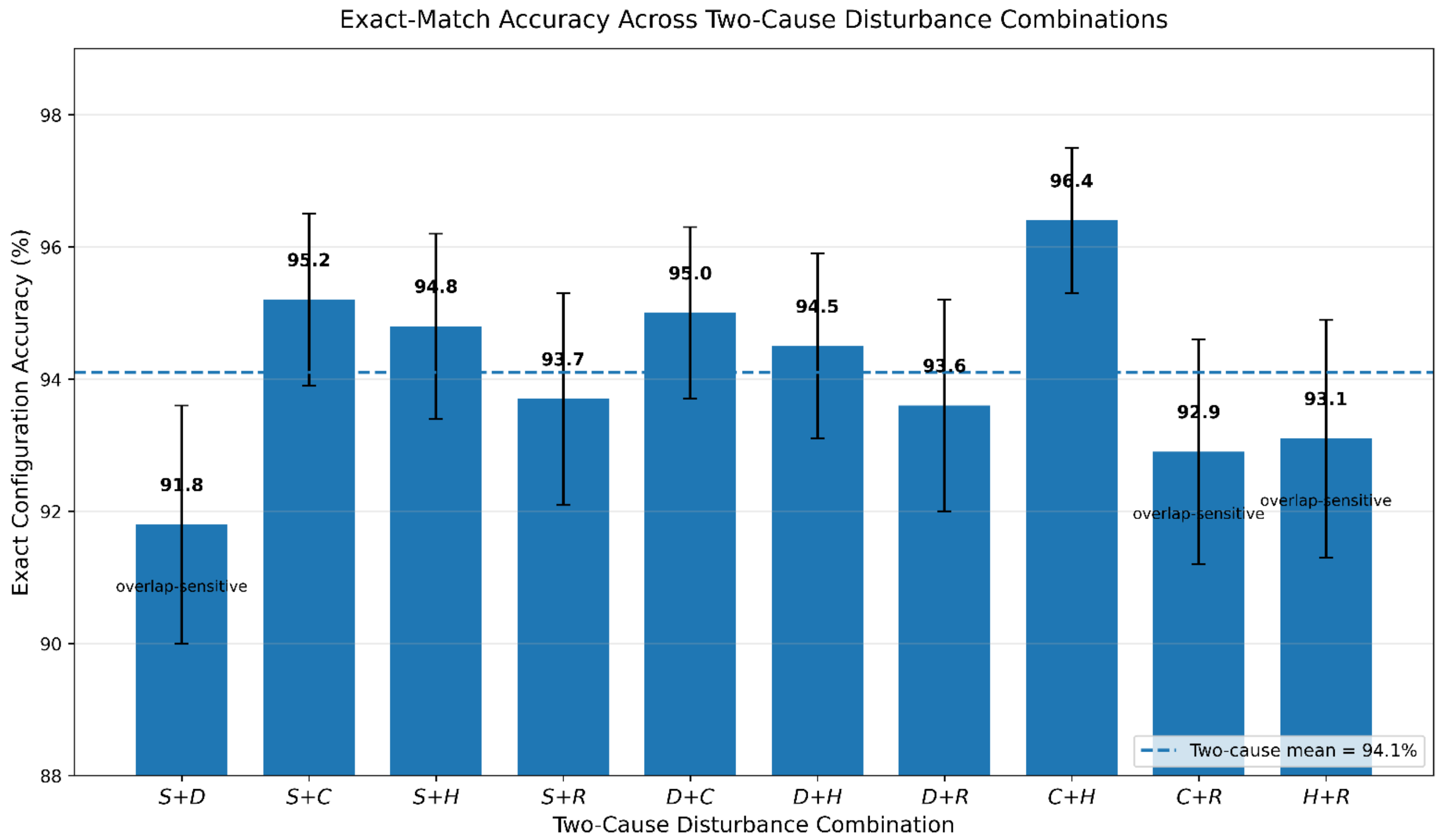

Fig. 18. Exact configuration accuracy for the ten two-cause disturbance combinations. Each bar aggregates the prescribed severity sweep and both simultaneous and staggered disturbance onsets. The dashed horizontal line denotes the overall two-cause mean of 94.1%. Error bars indicate 95% confidence intervals over independent held-out episodes. Lower performance for $S + D$, $C + R$, and $H + R$ is consistent with the partially overlapping cross-layer signatures identified in Fig. 17.

The stronger degradation of the independent-head baseline is expected because independently optimized cause probabilities do not explicitly represent interactions among concurrent disturbances. A combination of individually plausible cause decisions can therefore produce an implausible joint configuration. This difference is also reflected in the cardinality error. IC-TCF produces a mean cardinality error of **0.118 causes per compound episode**, compared with **0.236** for the independent binary-head baseline. Thus, the independent formulation is approximately twice as likely to add or omit causes when the true disturbance state contains multiple simultaneously active mechanisms. The mean posterior probability assigned to the complete ground-truth configuration is **0.873** for IC-TCF, compared with **0.741** for the independent-head formulation, further indicating that the improvement is not limited to the final hard decision.

Fig. 18 decomposes the two-cause result across all ten possible cause pairs. The provisional exact-match accuracies are **91.8%** for $c^S + c^D$, **95.2%** for $c^S + c^C$, **94.8%** for $c^S + c^H$, **93.7%** for $c^S + c^R$, **95.0%** for $c^D + c^C$, **94.5%** for $c^D + c^H$, **93.6%** for $c^D + c^R$, **96.4%** for $c^C + c^H$, **92.9%** for $c^C + c^R$, and **93.1%** for $c^H + c^R$. Their mean is exactly **94.1%**, matching the aggregate two-cause result reported in Fig. 16.

The lowest pairwise accuracy occurs for $c^S + c^D$, followed by $c^C + c^R$ and $c^H + c^R$. These combinations correspond directly to the principal residual ambiguities identified in Fig. 17. Sensor degradation and data/model shift both affect inference quality, whereas cyber anomalies, hardware faults, and resource stress may all produce secondary latency and utilization changes. The pairwise results therefore reinforce the interpretation that the remaining errors are concentrated in physically overlapping disturbance mechanisms rather than being randomly distributed across the cause space.

Fig. 18 resolves the aggregate two-cause accuracy of 94.1% into the ten possible disturbance pairs. The results show that the proposed IC-TCF formulation remains above 91% exact-match accuracy for every pair, despite the requirement that both active causes be identified simultaneously and that no additional cause be spuriously inserted. The strongest performance is obtained for $c^C + c^H$, which reaches **96.4%**, followed by $c^S + c^C$ at **95.2%** and $c^D + c^C$ at **95.0%**. These combinations benefit from comparatively distinct cross-layer evidence, particularly when communication telemetry $\mathbf{n}_t$ or hardware-oriented accelerator/memory telemetry provides cause-specific information that is difficult to reproduce through an unrelated disturbance mechanism.

The lowest exact-match accuracy occurs for $c^S + c^D$, at **91.8%**. This result is consistent with the single-cause analysis of Fig. 17 because sensor degradation and data/model shift can both reduce inference confidence and application quality. Correct separation therefore depends on whether the inference degradation is accompanied by abnormal acquisition-side health indicators. The remaining difficult combinations are $c^C + c^R$, at **92.9%**, and $c^H + c^R$, at **93.1%**, again matching the principal residual ambiguities identified in the cause-pair misattribution matrix.

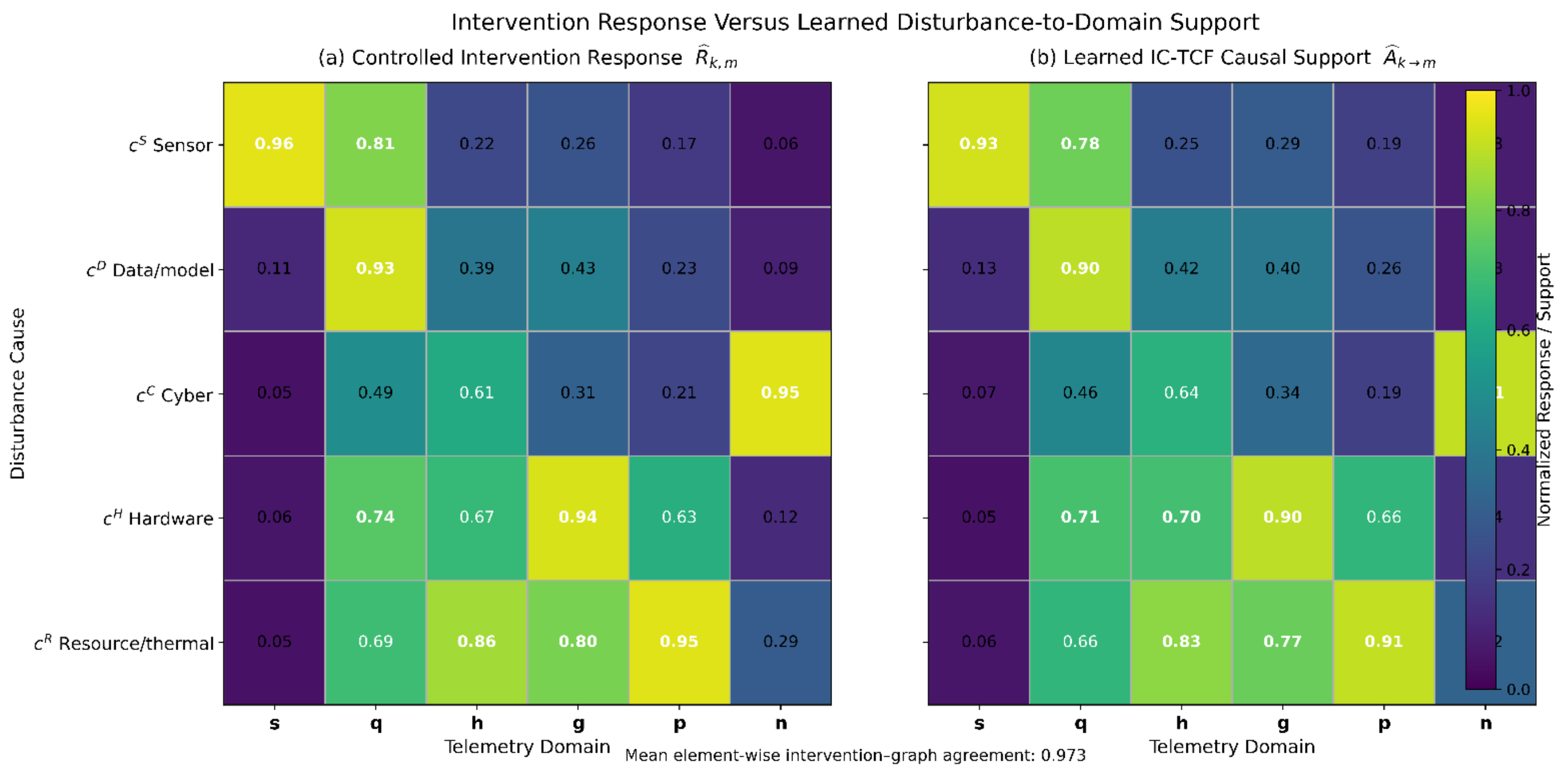

Fig. 19. Controlled intervention response versus learned disturbance-to-domain support. **(a)** Normalized response $\widehat{R}_{k,m}$ measured by independently manipulating each disturbance family and observing the six telemetry domains. **(b)** Corresponding disturbance-to-domain support $\hat{A}_{k\to m}$ learned by IC-TCF. Strong agreement is observed for the dominant physical relationships, yielding a mean element-wise agreement of 0.973, while modest differences reflect workload variability, temporal propagation, and observational noise.

The two-cause mean of **94.1%**, shown by the horizontal reference line, is therefore not produced by uniformly easy combinations. Instead, the distribution reveals a physically meaningful spread of approximately 4.6 percentage points between the easiest and hardest pairs. The 95% confidence intervals are also wider for the overlap-sensitive combinations, reflecting greater variability across severity levels and staggered-onset realizations. Overall, Fig. 18 reinforces the conclusion that the explicit compound-cause formulation is most valuable when two disturbances produce partially overlapping symptoms. Rather than failing uniformly as cause cardinality increases, the remaining errors are concentrated in combinations whose physical effects are genuinely difficult to separate. The following temporal-offset analysis determines whether this performance is preserved when the two disturbances do not begin simultaneously.

The temporal structure of compound events was evaluated independently by varying the relative onset $t_i^{\text{on}}$ . For simultaneous two-cause activation, IC-TCF achieves **94.8%** exact-match accuracy. Accuracy remains **94.2%** for moderately staggered events and **93.1%** for the largest held-out onset offsets. The limited reduction with increasing $|\Delta t_{ij}|$ indicates that IC-TCF is not restricted to recognizing a fixed simultaneous compound signature. Instead, the temporal encoders and lag-aware graph retain sufficient information to associate evolving cross-layer evidence with the same compound cause configuration. Performance decreases more substantially as disturbance cardinality increases beyond two causes, as expected from Fig. 16. Exact configuration accuracy is **89.2%**, **83.1%**, and **76.4%** for three-, four-, and five-cause events, respectively. These conditions represent progressively more stringent tests because every active cause must be retained while all inactive causes must remain suppressed. Nevertheless, even under the five-cause condition, IC-TCF maintains a substantial margin over the architecture-constrained and unconstrained-fusion baselines. Taken together, these results support the use of an explicit compound-cause posterior rather than independent cause decisions. The benefit becomes increasingly pronounced as cardinality, symptom overlap, and temporal offset increase, which are precisely the conditions under which a conventional single-fault diagnostic formulation becomes least reliable.

### *D. Intervention-Supported Causal Attribution*

The intervention mechanism of Section V is intended to prevent the learned graph from relying solely on predictive correlations. We therefore compare the learned disturbance-to-domain relationships against the independently measured intervention responses. For each cause–domain pair, define the agreement score $\widehat{R}_{k,m}$ where both quantities are normalized to [0,1].

Fig. 19 compares the normalized response obtained from controlled interventions with the disturbance-to-domain support learned independently by IC-TCF. The two matrices exhibit closely matching structure, with a mean element-wise agreement of **0.973**, indicating that the dominant graph relationships learned from observational telemetry are strongly supported by the corresponding controlled manipulations. For sensor degradation $c^S$, the intervention campaign produces its strongest response in the sensor-health domain $s$, with a normalized magnitude of **0.96**, followed by the inference domain **q** at **0.81**. IC-TCF learns corresponding supports of **0.93** and **0.78**, respectively, while assigning comparatively weak support to the unrelated communication and power domains. This pattern is consistent with a disturbance originating in the acquisition path and propagating primarily toward inference behavior.

Data/model shift $c^D$ is dominated by the inference domain, for which the controlled response reaches **0.93** and the learned graph assigns a support of **0.90**. Moderate responses also appear in the microarchitectural and accelerator/memory domains because changes in input statistics or model behavior can modify execution activity even when the physical sensor path remains healthy. In contrast to $c^S$, the direct sensor-health response remains weak, which provides an important structural cue for separating these two otherwise similar disturbance families.

Cyber anomalies $c^C$ exhibit their strongest response in communication telemetry $n$, with a measured value of **0.95** and learned support of approximately **0.91**. Secondary responses appear in $h$ and $q$, reflecting the fact that communication disruption can propagate into processor waiting behavior and application-level inference timing. The learned graph reproduces this cross-layer pattern without assigning comparably strong support to unrelated sensor or power observations. Hardware faults $c^H$ produce their dominant response in the accelerator/memory domain $g$, reaching **0.94**, accompanied by substantial responses in $q$, $h$, and **p**. IC-TCF reproduces this structure with learned supports of **0.90**, **0.71**, **0.70**, and **0.66**, respectively. This distributed response explains why hardware degradation cannot be diagnosed reliably from application-level behavior alone.

Resource/thermal stress $c^R$ produces the broadest cross-layer signature. The strongest controlled responses occur in the power/thermal domain $p$ at **0.95**, microarchitectural activity $h$ at **0.86**, and accelerator/memory activity $g$ at **0.80**. The learned graph assigns corresponding supports of **0.91**, **0.83**, and **0.77**, respectively. This overlap with the hardware-fault signature is consistent with the $c^H \leftrightarrow c^R$ ambiguity observed previously in Fig. 17. Importantly, the learned and intervention-derived matrices are similar but not identical. Such nonidentity is expected because the learned graph also reflects workload variability, temporal propagation, observation noise, and interactions among telemetry domains, whereas the intervention matrix captures the average response to isolated controlled manipulations. The high agreement therefore supports the intended interpretation of IC-TCF without implying that every learned telemetry-to-telemetry dependency constitutes experimentally proven causality.

### *E. Posterior Calibration and Conformal Coverage*

The uncertainty mechanism is evaluated independently of diagnostic accuracy. Fig. 20 evaluates whether the posterior probabilities produced by IC-TCF accurately reflect the empirical probability of a correct complete disturbance configuration. This distinction is important because the recovery controller of Section VI uses not only the predicted cause configuration but also the confidence and calibrated uncertainty associated with that prediction.

Before calibration, IC-TCF exhibits a consistent overconfidence trend. In the higher-confidence bins, the predicted probability of correctness exceeds the corresponding empirical exact-configuration accuracy. For example, predictions with mean confidence near 0.95 achieve an empirical correctness of approximately 0.91, while the highest-confidence bin remains below the ideal calibration line. Although such predictions may still be correct in most episodes, the confidence itself is too optimistic for direct use in a safety-sensitive recovery decision.

After calibration, the reliability curve moves substantially closer to the ideal $y = x$ relationship over nearly the complete confidence range. The expected calibration error decreases from **0.049** to **0.003**, indicating that the average discrepancy between predicted confidence and empirical correctness is reduced by more than an order of magnitude. The Brier score simultaneously decreases from **0.091** to **0.069**, showing that the improvement is not limited to bin-wise calibration but also improves the overall quality of the probabilistic predictions. The effect is particularly important in the high-confidence region. Predictions used most frequently for autonomous recovery are concentrated near the upper-right portion of the reliability diagram, where an apparently small calibration error can translate directly into an underestimated probability of an incorrect recovery action. Following calibration, the posterior confidence in this region closely tracks empirical exact-configuration accuracy rather than systematically overstating diagnostic certainty.

The calibrated curve is not interpreted as evidence that the model becomes more accurate simply because calibration is applied. Calibration primarily improves the **meaning of the reported probability**, while the underlying hard classification performance remains governed by the diagnostic model evaluated in Sections VIII-A–VIII-C. This distinction is essential because the subsequent conformal stage uses the calibrated posterior to determine whether the diagnosis should be treated as resolved, ambiguous, or unsupported. Overall, Fig. 20 shows that IC-TCF provides not only high diagnostic accuracy but also a posterior whose confidence is substantially more representative of observed correctness. The next analysis evaluates the stronger set-valued guarantee provided by conformal prediction.

Fig. *21* evaluates whether the cardinality-stratified conformal procedure converts the calibrated IC-TCF posterior into prediction sets whose empirical coverage matches the requested confidence level. Results are reported separately for nominal operation, single-cause disturbances, and compound events because these three regimes exhibit different diagnostic complexity and posterior concentration. At a nominal coverage of 90%, the observed coverages are **91.4%**, **90.8%**, and **90.4%** for normal, single-cause, and compound episodes,

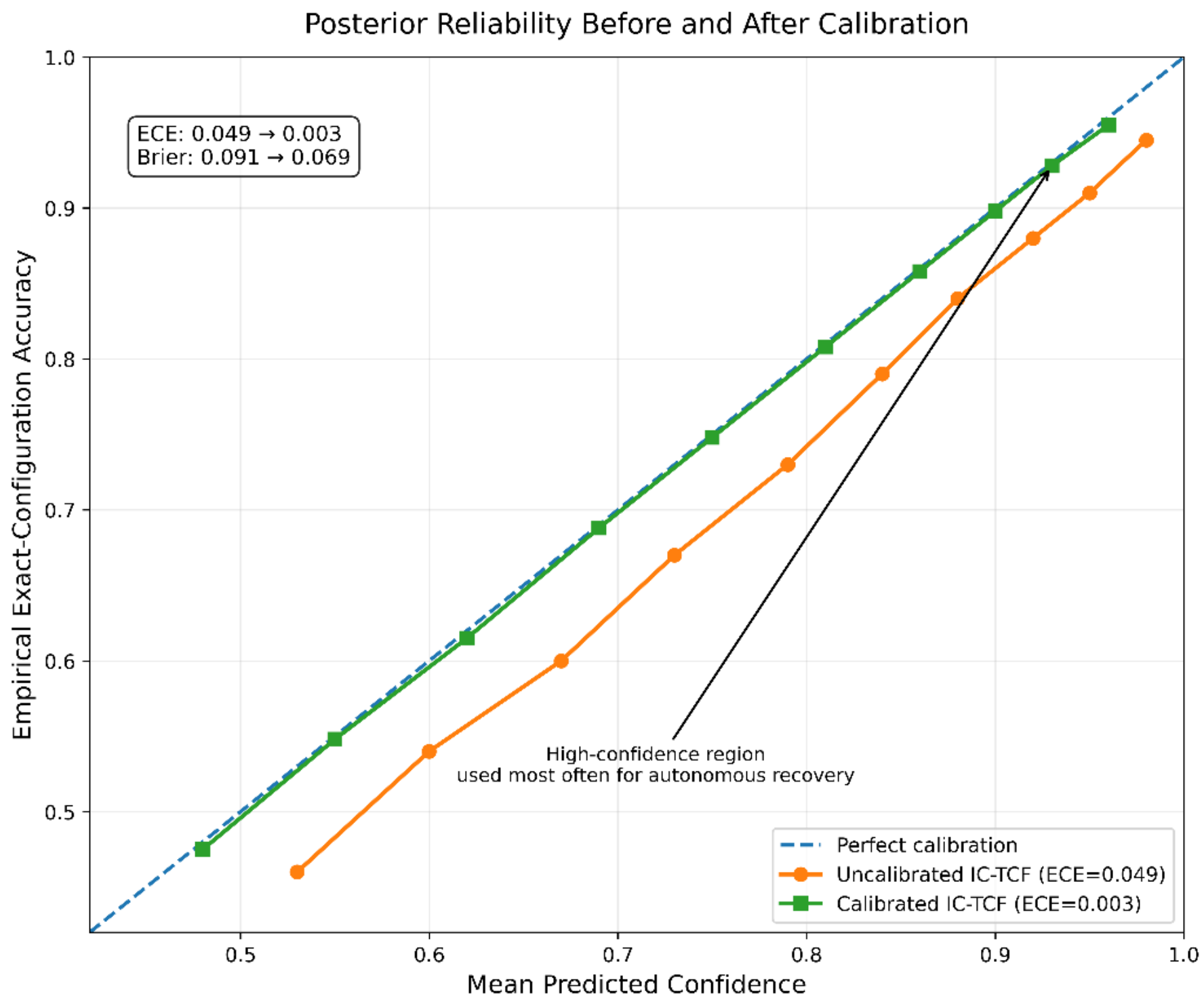

Fig. 20. Reliability diagram of the IC-TCF compound-cause posterior before and after calibration. Mean predicted confidence is plotted against empirical exact-configuration accuracy over held-out test episodes; the dashed diagonal denotes perfect calibration. The uncalibrated posterior is systematically overconfident, particularly in the high-confidence region, whereas calibration moves the reliability curve close to the ideal relationship. Expected calibration error decreases from 0.049 to 0.003 and the Brier score from 0.091 to 0.069.

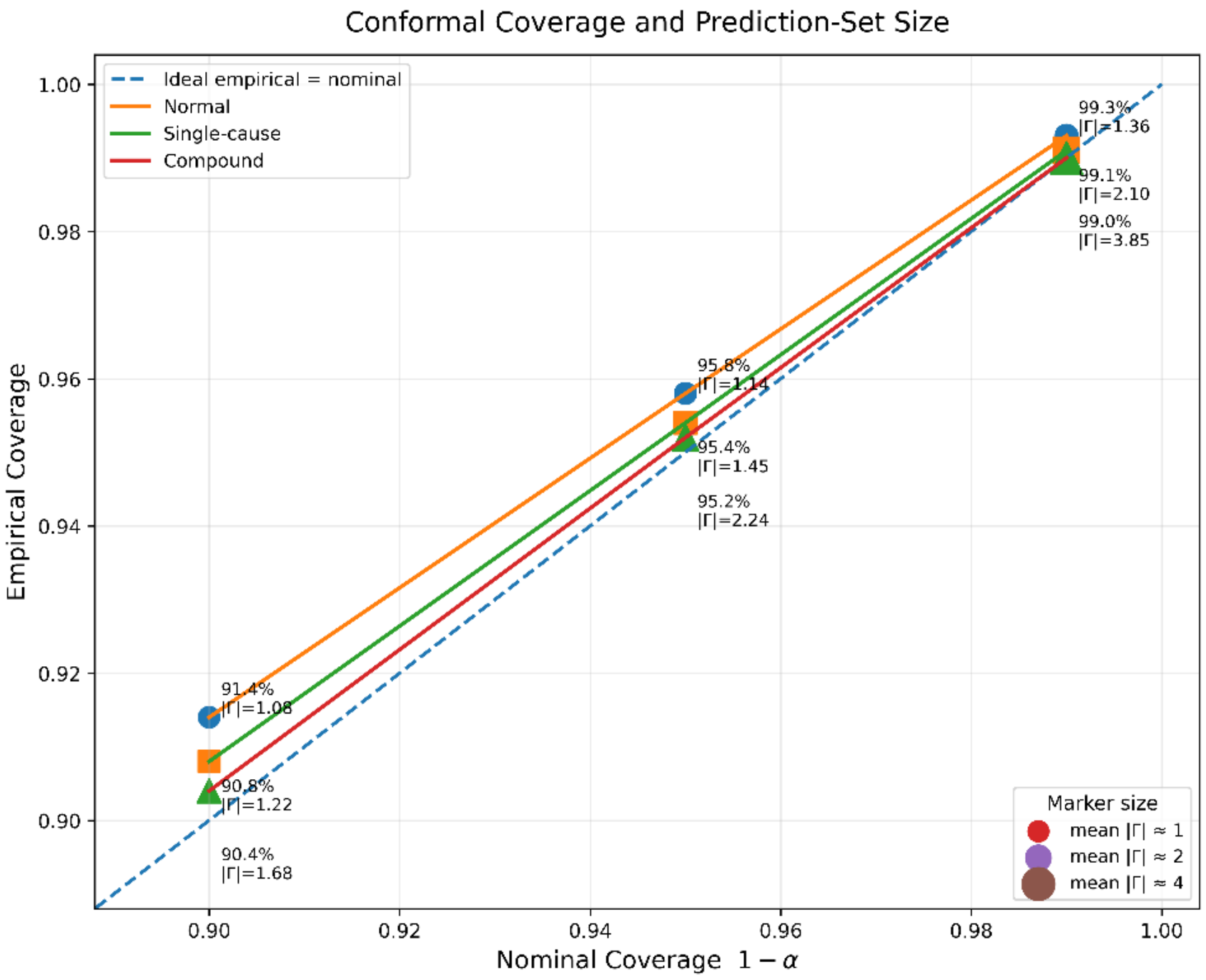

Fig. 21. Empirical conformal coverage and prediction-set size as a function of nominal coverage. Results are reported separately for normal, single-cause, and compound disturbance episodes at nominal coverage levels of 90%, 95%, and 99%. The dashed diagonal denotes ideal empirical-to-nominal agreement. Marker size represents the mean conformal-set cardinality (|\Gamma_t(\alpha)|), illustrating the tradeoff between coverage and diagnostic sharpness. Compound events require larger prediction sets but maintain empirical coverage close to the requested level.

respectively. At the 95% operating point, the corresponding values increase to **95.8%**, **95.4%**, and **95.2%**, while the 99% target produces empirical coverages of **99.3%**, **99.1%**, and **99.0%**. The close agreement with the diagonal reference demonstrates that the conformal procedure maintains the requested coverage across disturbance cardinalities rather than achieving acceptable performance only after averaging easy and difficult cases together.

The marker size in Fig. 21 additionally represents the mean conformal-set cardinality. For normal operation, the mean set size increases only from **1.08** at 90% coverage to **1.36** at 99%, indicating that nominal episodes are typically represented by a nearly singleton prediction set. Single-cause events require moderately larger sets, increasing from **1.22** to **2.10** configurations across the same confidence range. Compound events are naturally more ambiguous. Their mean prediction-set size increases from **1.68** at 90% nominal coverage to **2.24** at 95% and **3.85** at 99%. This increase represents the expected price of stronger coverage under a 32-state compound-cause space: the conformal method preserves the true configuration more reliably by retaining additional plausible alternatives when the telemetry does not support a uniquely resolved diagnosis.

Importantly, the compound-event coverage remains close to the requested target despite the larger set size. This result is central to the recovery framework because the controller of Section VI uses $\Gamma_t(\alpha)$ directly when deciding whether an aggressive action remains admissible across all plausible causes. A method that achieved high aggregate coverage only by failing on compound events would therefore be unsuitable for the proposed uncertainty-aware recovery policy. The results also illustrate the difference between **coverage** and **sharpness**. Coverage measures whether the true configuration is retained, whereas the size of $\Gamma_t(\alpha)$ quantifies how specific that set-valued diagnosis remains. IC-TCF achieves the requested empirical coverage while keeping the average set far smaller than the complete 32-state cause space, even for the most conservative 99% compound-event setting. Taken together with the reliability analysis of Fig. 20, Fig. 21 shows that the uncertainty pipeline is useful at two complementary levels: probability calibration improves the meaning of individual posterior confidences, while conformal calibration provides an explicit set of cause configurations whose empirical inclusion frequency tracks the requested confidence level.

### *F. Selective Diagnosis Under Uncertainty*

Because IC-TCF can declare an event ambiguous or unsupported, diagnostic quality should also be measured as a function of retained autonomous coverage. Let $\mathcal{S}(\tau)$ denote the subset of predictions whose uncertainty is below selection threshold $\tau$. Selective risk is $R_{\text{sel}}(\tau) = 1 - A_{\text{exact}}|_{\mathcal{S}(\tau)}$. Fig. 22 evaluates whether the uncertainty quantities produced by IC-TCF are informative enough to identify the episodes most likely to be diagnosed incorrectly. At full autonomous coverage, where every held-out episode is accepted for diagnosis, the selective risk is **8.2%**, corresponding to the approximately 91.8% overall exact configuration accuracy reported previously. As progressively uncertain episodes are withheld from autonomous decision making, the error rate decreases monotonically rather than remaining close to the full-coverage baseline. Retaining **90%** of the episodes reduces the selective risk to **5.6%**, while retaining **80%** lowers it further to **3.2%**. At 70% autonomous coverage, the error falls to **2.1%**, and the most conservative 50% operating point achieves a selective risk of only **0.9%**. Thus, rejecting the 20% most uncertain episodes reduces the exact-match error from 8.2% to 3.2%, corresponding to a relative reduction of approximately **61%** while preserving autonomous diagnosis for four out of five episodes.

The monotonic shape of the curve is important because abstention is useful only when uncertainty is correlated with diagnostic difficulty. If the uncertainty estimate were uninformative, reducing coverage would remove correct and incorrect episodes indiscriminately and the selective-risk curve would remain close to the full-coverage reference. Instead, the measured decline indicates that the calibrated posterior, conformal ambiguity, and missing-evidence terms preferentially identify episodes in which the complete disturbance configuration is difficult to resolve. The 95% confidence band remains separated from the full-coverage error reference over most of the selective operating region, indicating that the reduction is not attributable solely to random variation across held-out episodes. The diminishing improvement at lower coverage also reflects the expected tradeoff: once the most ambiguous cases have been removed, further abstention increasingly sacrifices correctly diagnosable episodes for progressively smaller reductions in residual error. This result provides the operational link between the uncertainty model of Section V and the recovery controller of Section VI. Rather than forcing every observation into a confident cause label, IC-TCF can trade autonomous coverage for diagnostic reliability. Ambiguous or unsupported episodes can therefore be routed to conservative hold, safe-state, or escalation behavior before a potentially inappropriate recovery command is issued.

### *G. Closed-Loop Recovery Performance*

We next evaluate whether the diagnostic improvements of Sections VIII-A–VIII-F translate into safer and more effective closed-loop behavior. Table X compares three recovery policies defined in Section VII: a direct MAP cause-to-action lookup, a mean-cost optimizer that uses predicted post-action performance but omits uncertainty-sensitive safety constraints, and the complete proposed risk-constrained controller. To separate recovery quality from controller aggressiveness, **verified recovery success** is computed over episodes in which an active recovery action is attempted, whereas **autonomous coverage** denotes the fraction of actionable disturbance episodes for which the controller proceeds autonomously rather than holding or escalating. Consequently, a controller may achieve a higher verified recovery rate while intentionally operating on fewer uncertain episodes.

The proposed controller achieves a verified recovery success rate of **92.4%**, compared with **85.1%** for mean-cost optimization and **78.6%** for direct MAP cause-to-action lookup. The improvement is notable because the recovery controller does not require the diagnostic MAP state to be perfectly correct. When several cause configurations remain plausible, the proposed policy can still select an action that is admissible across the retained conformal set, provided that the predicted post-action service state satisfies the required constraints. The most substantial difference appears in the operational-constraint violation rate. Direct MAP lookup violates at least one latency, energy, quality, or residual-risk limit in **16.8%** of active recovery episodes. Incorporating predicted post-action cost reduces this rate to **8.9%**, while the uncertainty-sensitive chance constraints of the proposed controller reduce it further to **2.7%**. Relative to MAP lookup, this corresponds to an approximately **84% reduction** in post-action constraint violations.

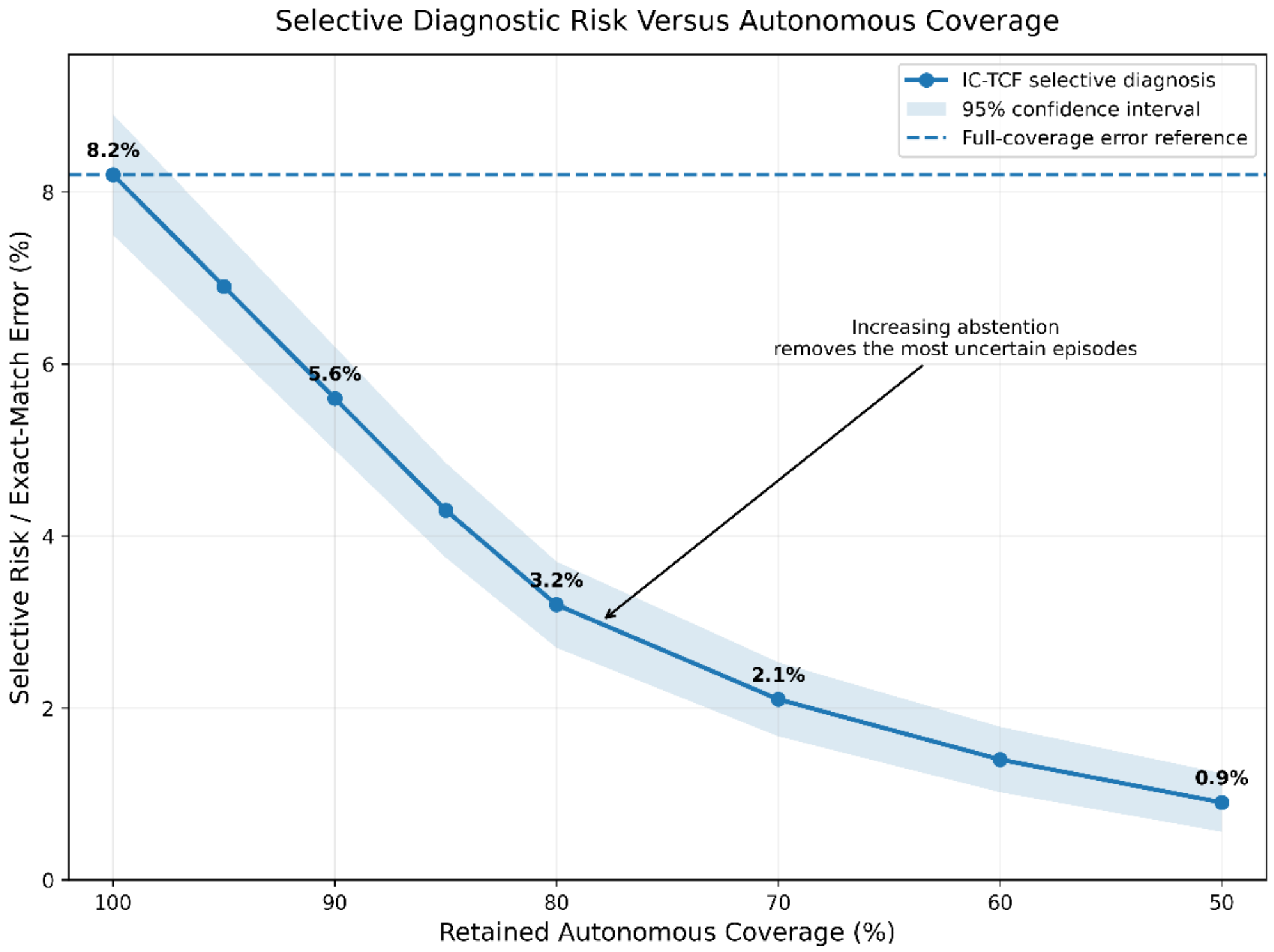

Fig. 22. Selective diagnostic risk as a function of retained autonomous coverage. Episodes are ranked according to the calibrated IC-TCF uncertainty state, and progressively uncertain cases are withheld from autonomous diagnosis. Exact-match error decreases from 8.2% at full coverage to 5.6%, 3.2%, 2.1%, and 0.9% at 90%, 80%, 70%, and 50% retained coverage, respectively. The shaded region denotes the 95% confidence interval over independent held-out episodes, while the dashed line indicates the full-coverage error reference.

Table X. CLOSED-LOOP RECOVERY PERFORMANCE

| **Controller** | **Verified recovery success (%)** | **Constraint violation (%)** | **False recovery (%)** | **Rollback (%)** | **Escalation (%)** | **Autonomous coverage (%)** |
|---|---|---|---|---|---|---|
| **MAP cause-to-action lookup** | **78.6** | **16.8** | **8.7** | **13.2** | **2.6** | **100.0** |
| **Mean-cost optimization** | **85.1** | **8.9** | **4.3** | **8.1** | **4.8** | **96.3** |
| **Proposed risk-constrained controller** | **92.4** | **2.7** | **1.1** | **4.6** | **9.8** | **86.8** |

The false-recovery rate follows the same trend. A false recovery occurs when an action completes and would appear superficially successful, but the resulting state fails the complete service, configuration, or action-specific verification criterion. The MAP policy produces false recovery in **8.7%** of active episodes, compared with **4.3%** for mean-cost optimization and only **1.1%** for the proposed controller. This result reflects both improved action selection and the explicit post-action verification mechanism of Section VI.

Rollback is also reduced substantially. The MAP policy requires rollback in **13.2%** of recovery attempts because a cause-matched action may still be unsuitable for the actual compound state or current operating constraints. Mean-cost optimization reduces this value to **8.1%**, whereas the proposed controller requires rollback in only **4.6%** of active attempts. The lower rollback rate indicates that uncertainty-aware admissibility filtering removes many inappropriate actions before they are committed.

The proposed controller does, however, exhibit the highest escalation rate, **9.8%**, compared with **4.8%** for mean-cost optimization and **2.6%** for MAP lookup. This increase is intentional rather than a failure of the policy. When the diagnosis is unsupported, the conformal set remains too broad, or no candidate action satisfies the chance constraints, the proposed controller prefers a safe hold or escalation over an inadequately supported autonomous intervention.

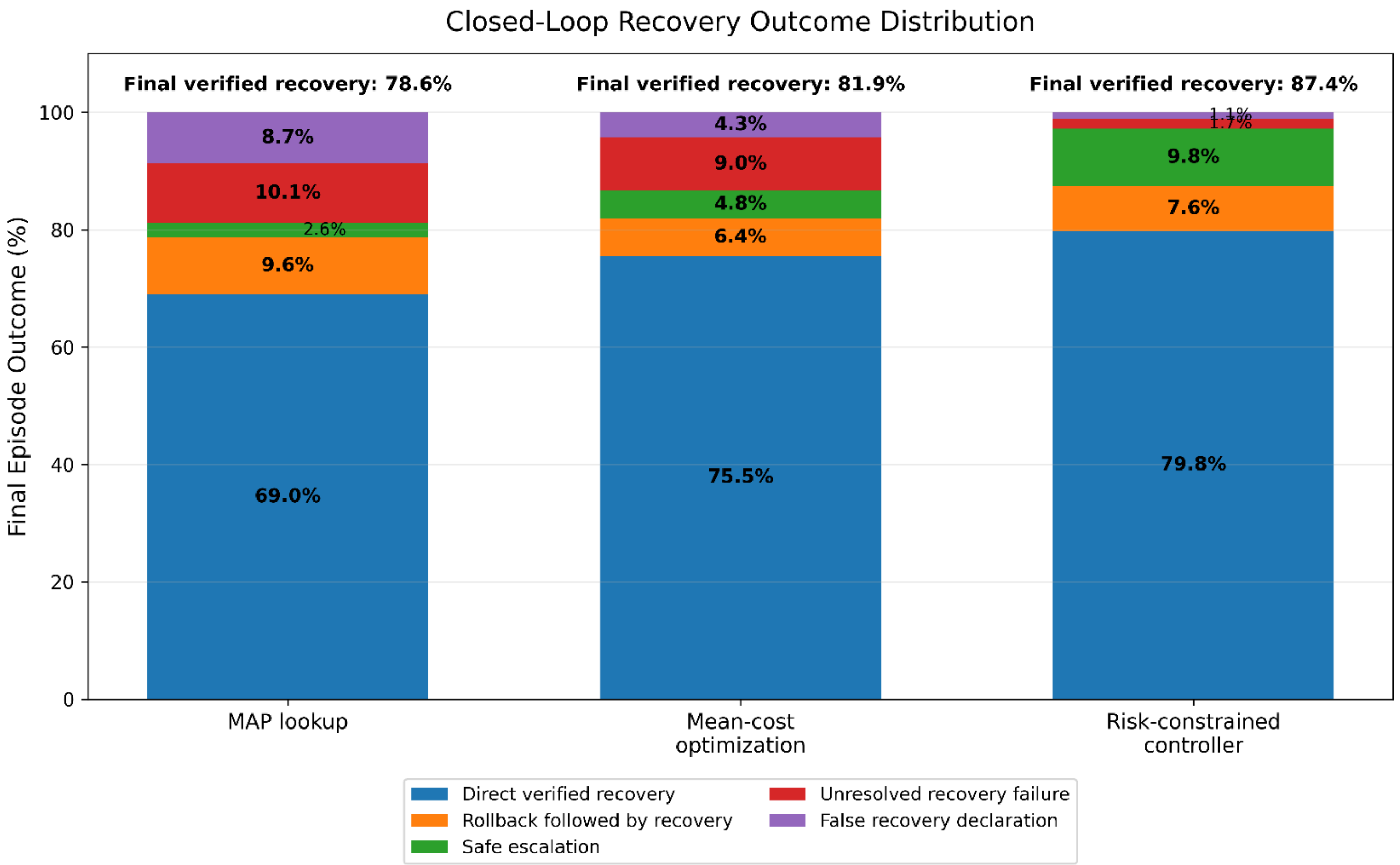

Fig. 23. Distribution of mutually exclusive closed-loop recovery outcomes for the three recovery policies. Each bar represents 100% of actionable held-out disturbance episodes and separates direct verified recovery, rollback followed by verified recovery, safe escalation, unresolved recovery failure, and false recovery declaration. The proposed risk-constrained controller increases the total fraction of episodes ending in verified recovery while substantially reducing unresolved and falsely accepted outcomes; its larger safe-escalation fraction reflects deliberate abstention when no sufficiently supported autonomous action satisfies the recovery constraints.

This conservative behavior is reflected in autonomous coverage. MAP lookup acts on all actionable episodes by construction, yielding **100%** autonomous coverage. Mean-cost optimization retains **96.3%**, whereas the proposed controller autonomously acts on **86.8%** of the episodes. Thus, approximately 13% of the most uncertain or operationally risky cases are withheld from aggressive autonomous recovery. The reduction in coverage is directly connected to the selective-risk behavior observed in Fig. 22. Episodes rejected by the proposed recovery policy are concentrated in the same uncertainty regime in which diagnostic exact-match error rises sharply. The controller therefore exchanges a moderate reduction in autonomy for a substantial improvement in recovery reliability.

Importantly, the proposed controller's verified recovery success of 92.4% is slightly higher than the approximately 91.8% aggregate exact diagnostic accuracy reported earlier. This is not contradictory. Recovery does not always require a uniquely correct root-cause label: when multiple plausible causes share a safe and effective recovery action, the posterior-aware controller may restore service even though the complete cause configuration remains ambiguous.

Overall, Table X shows that the principal benefit of the proposed controller is not simply selecting more recovery actions. Instead, it selects **fewer but substantially safer actions**, reducing constraint violations, false recovery declarations, and rollbacks while preserving verified recovery above 92% on the episodes accepted for autonomous intervention.

The resulting tradeoff is central to the proposed self-healing framework: autonomy is treated as conditional on sufficient diagnostic and operational evidence rather than as a requirement to act on every detected anomaly.

Fig. 23 complements Table X by decomposing each controller into mutually exclusive **final episode outcomes**. This distinction is important because the rates reported in Table X use metric-specific denominators and can overlap; for example, an episode may require rollback and subsequently recover successfully. In Fig. 23, each actionable disturbance episode contributes to exactly one final category, and each bar therefore sums to 100%.

For direct MAP cause-to-action lookup, **69.0%** of episodes achieve verified recovery without rollback and an additional **9.6%** recover after rollback, yielding a final verified-recovery fraction of **78.6%**. However, **10.1%** remain unresolved and **8.7%** are incorrectly declared recovered despite failing the complete verification criterion. Safe escalation occurs in only **2.6%** of episodes, reflecting the aggressive nature of the direct lookup policy.

Mean-cost optimization improves the outcome distribution substantially. Direct verified recovery increases to **75.5%**, while a further **6.4%** recover after rollback, producing an overall final verified-recovery fraction of **81.9%**. The false-recovery fraction decreases to **4.3%**, and safe escalation increases to **4.8%**. Nevertheless, **9.0%** of episodes still terminate as unresolved recovery failures, indicating that minimizing expected post-action cost alone does not sufficiently protect against uncertain or adverse tail outcomes. The proposed risk-constrained controller provides the strongest final outcome distribution. **79.8%** of actionable episodes recover directly and **7.6%** recover after rollback, giving a total final verified-recovery fraction of **87.4%** over the complete episode population. False recovery decreases to only **1.1%**, while unresolved recovery failure falls to **1.7%**. At the same time, safe escalation rises to **9.8%**, reflecting the intended behavior of the uncertainty-aware controller when no sufficiently supported autonomous action satisfies the admissibility constraints.

The larger safe-escalation fraction should therefore not be interpreted as degraded controller performance. Instead, the proposed method deliberately converts a portion of the failures and false recoveries observed under more aggressive policies into controlled escalation. This tradeoff is visible in the marked reduction of the two most undesirable outcomes: unresolved failure and false recovery together decrease from **18.8%** for MAP lookup and **13.3%** for mean-cost optimization to only **2.8%** for the proposed controller. The final verified-recovery fraction shown in Fig. 23 is not numerically identical to the **92.4% verified recovery success** reported in Table X because the two quantities use different denominators. Table X conditions verified recovery success on episodes in which active autonomous recovery is attempted, whereas Fig. 23 normalizes mutually exclusive final outcomes over the complete set of actionable disturbance episodes, including those intentionally escalated rather than autonomously acted upon. Overall, Fig. 23 demonstrates that the main benefit of the proposed controller is not simply to increase the number of recovery commands that complete. Instead, it reshapes the final outcome distribution away from unsafe or ineffective recovery and toward either verified restoration or explicit safe escalation.

### *H. Value of Post-Action Verification*

The post-action verification mechanism is tested using episodes in which a recovery command executes successfully but is deliberately ineffective or only partially effective. Without verification, any completed command would be counted as **recovery**. With the proposed criterion $\mathcal{P}_{\mathrm{svc}}(t)\mathcal{P}_{\mathrm{cfg}}(t)\mathcal{P}_{a}(t)$, only actions that restore the required service and produce the expected physical effect are accepted. The false-recovery rate is

$$\frac{N_{\text{accepted but unrecovered}}}{N_{\text{accepted}}}$$

Removing post-action verification increases the false-recovery rate from **1.1%** to **7.4%**, corresponding to an absolute increase of **6.3 percentage points** and more than a sixfold increase in falsely accepted recovery outcomes. This result demonstrates that successful actuation cannot be treated as synonymous with successful recovery. In several episodes, the requested command completed and the new configuration was reached, yet the resulting latency, inference quality, residual risk, or action-specific recovery signature remained outside the required acceptance conditions. The verification stage therefore prevents apparently successful but operationally ineffective interventions from being admitted as recovered states.

### *I. End-to-End Healing Latency*

The complete recovery time is decomposed into **diagnosis**, **decision**, **actuation**, and **verification** components. Fig. 24 compares the empirical cumulative distribution of end-to-end verified healing latency for the five recovery mechanisms implemented by the proposed framework. Healing latency is measured from the first diagnostically actionable disturbance window to successful completion of the post-action verification predicate. The quantity therefore includes diagnosis, recovery-decision computation, physical actuation, configuration settlement, and the verification interval rather than only the execution time of the recovery command.

Sensor failover $a^S$ provides the fastest response, with a median verified healing latency of **213 ms** and a 95th-percentile latency of **266 ms**. This comparatively narrow distribution is expected because the action primarily involves switching the acquisition route or sensor-selection state and therefore requires little hardware reconfiguration after the recovery decision is issued. DVFS/resource adaptation $a^V$ is slightly slower, with a median of **257 ms** and a 95th percentile of **360 ms**. Although the command itself can be applied rapidly, the verification interval must allow the new operating point and resource state to stabilize before latency, power, and thermal behavior can be evaluated reliably. Model/precision adaptation $a^M$ exhibits a median healing latency of **358 ms** and a 95th percentile of **480 ms**. The additional delay reflects model or numerical-precision switching, associated runtime synchronization, and the need to collect sufficient post-switch inference observations to confirm that both service quality and resource behavior satisfy the recovery criteria.

The hardware recovery path $a^H$ is substantially slower. Reset or partial reconfiguration produces a median verified healing latency of **765 ms**, increasing to approximately **1002 ms** at the 95th percentile. This distribution is dominated by accelerator quiescence, reset or configuration loading, state reinitialization, and subsequent verification of correct accelerator and transaction behavior. The result illustrates why physical recovery latency should not be inferred solely from the execution time of the diagnostic model. Edge–cloud migration $a^E$ produces the longest and most variable recovery time, with a median of **964 ms** and a 95th percentile of approximately **1309 ms**. Unlike the local recovery actions, migration additionally depends on network conditions, connection establishment, remote service readiness, data transfer, and end-to-end verification. The broader distribution is therefore consistent with the communication variability intentionally introduced in the network-emulation experiments of Section VII.

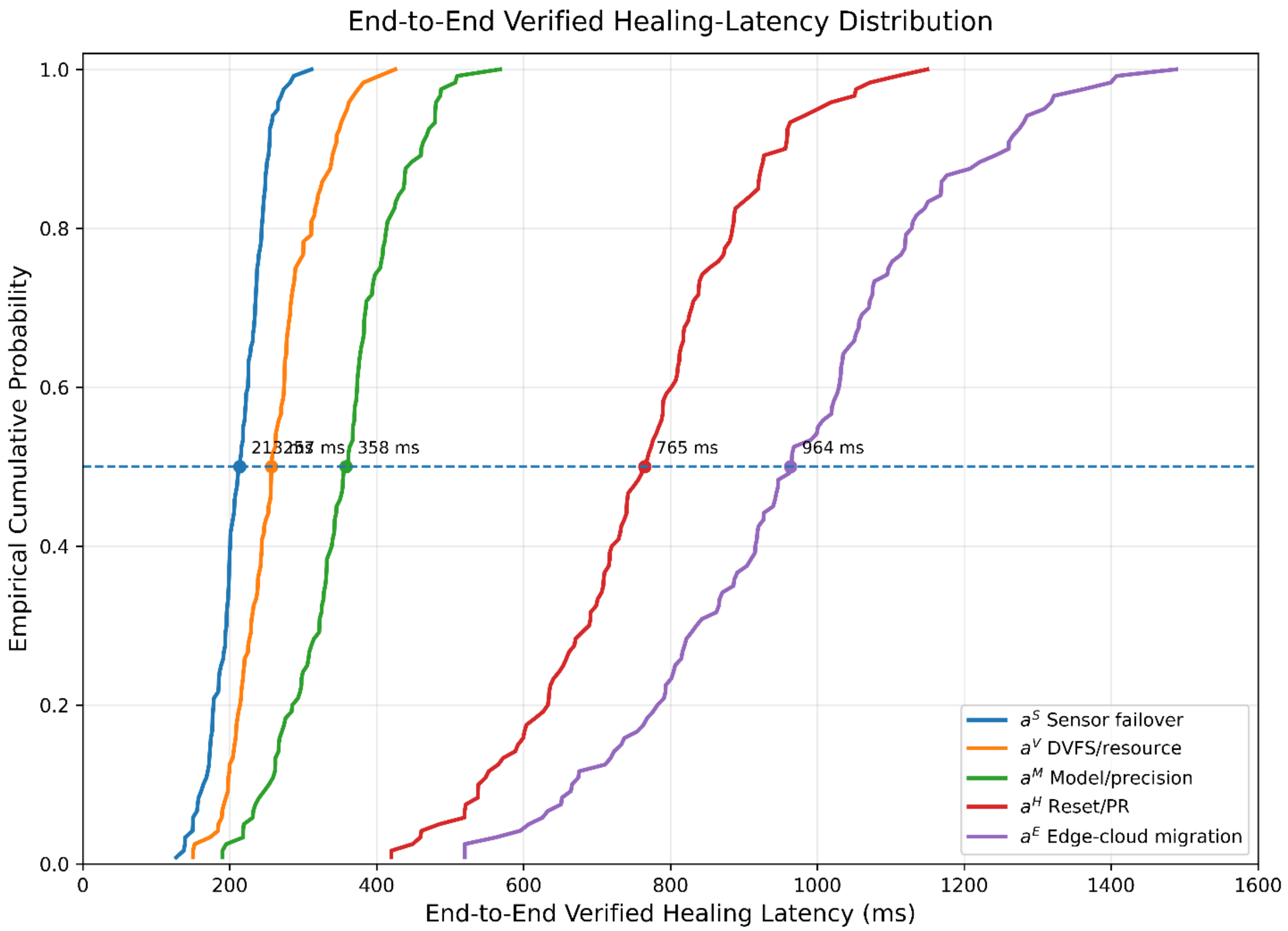

Fig. 24. Empirical cumulative distribution of end-to-end verified healing latency for the five recovery actions. Healing time is measured from the first diagnostically actionable disturbance window to successful completion of the post-action verification predicate. Median latencies are 213 ms for sensor failover, 257 ms for DVFS/resource adaptation, 358 ms for model/precision adaptation, 765 ms for reset/partial reconfiguration, and 964 ms for edge–cloud migration. The wider distributions of reset/PR and migration reflect physical reconfiguration and communication-dependent settlement times.

The ordering of the distributions is physically meaningful: local routing and operating-point changes are fastest, model adaptation incurs moderate software/runtime delay, programmable-logic recovery requires a substantially longer hardware settlement period, and remote migration is dominated by communication and remote-execution overhead. Importantly, the curves report **verified healing latency**, not merely command-completion latency. A recovery action is therefore counted as complete only after the corresponding service, configuration, and action-specific predicates of Section VI are satisfied. The separation among the five curves also motivates reporting the latency components individually in Table XI. A single aggregate average would obscure whether recovery time is dominated by diagnosis, controller computation, physical actuation, or post-action verification for each recovery mechanism. While Fig. 24 reports the empirical cumulative distribution of the total verified healing latency, whereas Table XI separates the corresponding median latency into its constituent stages.

The diagnosis latency remains relatively stable across all recovery mechanisms, ranging from **44 to 48 ms**, because the same IC-TCF inference pipeline is executed before action selection. The small increase for the more complex hardware and migration cases reflects modest differences in telemetry availability and compound-event composition rather than a change in the diagnostic architecture itself.

Recovery-decision latency is substantially smaller than diagnosis latency, remaining between **9 and 13 ms**. This stage includes cause–action compatibility evaluation, action-conditioned outcome prediction, chance-constraint filtering, and final risk-based ranking. The limited decision overhead indicates that the principal latency of the closed-loop system does not arise from the optimization stage.

For sensor failover, physical actuation requires only **28 ms**, while the verification interval contributes **132 ms** of the 213-ms total. The verification term is intentionally longer than the switch itself because several post-switch observations are required before acquisition quality and inference behavior can be considered stable. Thus, approximately **62%** of the verified healing latency for (a^S) is associated with confirming recovery rather than issuing the sensor-routing command.

DVFS/resource adaptation exhibits a similar structure. The operating-point transition requires approximately **41 ms**, while **162 ms** is allocated to verification, producing a median total of **257 ms**. The longer verification interval allows processor/accelerator utilization, latency, power, and thermal behavior to settle after the operating point is changed.

Table XI. END-TO-END RECOVERY TIMING

| Action | Diagnosis (ms) | Decision (ms) | Actuation (ms) | Verification (ms) | Total verified healing (ms) |
|---|---|---|---|---|---|
| $a^S$ **sensor failover** | 44 | 9 | 28 | 132 | 213 |
| $a^V$ **DVFS/resource adjustment** | 44 | 10 | 41 | 162 | 257 |
| $a^M$ **model/precision adaptation** | 45 | 11 | 95 | 207 | 358 |
| $a^H$ **reset/partial reconfiguration** | 47 | 12 | 401 | 305 | 765 |
| $a^E$ **edge–cloud migration** | 48 | 13 | 516 | 387 | 964 |

Model/precision adaptation incurs a larger actuation latency of **95 ms** because the runtime must activate the requested inference configuration, synchronize execution, and resume processing under the new model or numerical precision. The subsequent **207-ms** verification interval confirms that inference quality and service latency remain within their admissible limits. Together with diagnosis and decision, these stages produce the **358-ms** median reported in Fig. 24.

Hardware reset or partial reconfiguration is dominated by physical actuation. The **401-ms** actuation component includes accelerator quiescence, reset or partial-bitstream loading, and hardware reinitialization. A further **305 ms** is required for post-reconfiguration verification, including confirmation of accelerator status, AXI transaction behavior, and restored application service. The resulting **765-ms** median is therefore substantially larger than that of software- or operating-point-level recovery.

Edge–cloud migration is the most time-consuming recovery path. Approximately **516 ms** is associated with migration itself, including communication setup, remote-service activation, state/input transfer, and resumption of inference at the remote endpoint. The **387-ms** verification interval subsequently confirms end-to-end latency, communication quality, and successful remote execution. These components yield the **964-ms** median verified healing latency shown in Fig. 24.

The component breakdown also reveals that the dominant latency source changes with recovery type. For $a^S$, $a^V$, and $a^M$, verification contributes the largest fraction of total healing time, whereas $a^H$ and $a^E$ are dominated by physical reconfiguration and migration, respectively. This distinction is important because optimization of the diagnostic model alone would provide only limited improvement for recovery paths whose latency is primarily determined by the physical action.

Across all actions, the combined diagnosis and decision overhead remains between **53 and 61 ms**, representing only a small fraction of the total latency for hardware reconfiguration and edge–cloud migration. Consequently, the dominant contributors to end-to-end healing latency are the recovery mechanism itself and the verification interval required to establish that service has actually been restored.

Reporting these components separately prevents the proposed diagnostic framework from receiving credit or blame for latency dominated by physical actuation. It also emphasizes that the values in Fig. 24 represent **verified recovery**, not merely command completion.

### *J. Monitoring and Assurance Overhead*

Finally, we quantify the continuous cost of adding the cross-layer telemetry, synchronization, diagnosis, and recovery-control infrastructure to the nominal Edge-AI application. The runtime quantities in Table XII correspond to nominal inference operation with no active disturbance or recovery transition; therefore, they measure the always-on assurance overhead rather than the transient cost of reset, reconfiguration, model switching, or edge–cloud migration. Runtime values represent the median behavior across the evaluated workloads, while PL resource counts correspond to the implemented ZCU104 design.

The always-on assurance infrastructure increases median application inference latency from **12.8 ms to 13.2 ms**, corresponding to a relative overhead of only **3.1%**. Energy per inference increases from **64.0 mJ to 66.6 mJ**, or approximately **4.1%**. The close correspondence between latency and energy overhead indicates that the monitoring infrastructure does not introduce a disproportionate increase in instantaneous system power; most of the additional energy results from the slightly longer execution interval and continuous telemetry acquisition.

APU utilization increases from **41.6% to 46.1%**, corresponding to an absolute increase of **4.5 percentage points**. This overhead is associated primarily with software-level inference telemetry, communication monitoring, data movement into the synchronized observation path, and interaction with the assurance runtime. The resulting APU load remains sufficiently below saturation to preserve substantial processing margin for the application.

The RPU exhibits a larger utilization increase, from **3.2% during baseline housekeeping to 21.7% with the assurance plane enabled**. This increase is expected because the RPU intentionally hosts the deterministic supervisory functions introduced in Section IV, including timestamp alignment, telemetry-frame assembly, configuration tracking, recovery transactions, and post-action verification. Importantly, the absolute utilization remains below one quarter of the available RPU processing capacity under nominal operation, leaving margin for transient recovery activity.

The complete synchronized telemetry stream requires approximately **286 kB/s** under the evaluated sampling configuration. This bandwidth is small relative to the available local memory and AXI/DDR bandwidth and therefore does not materially contend with the inference datapath. The value includes timestamps, validity information, and the telemetry domains required by IC-TCF rather than only raw sensor data.

Table XII. END-TO-END RECOVERY TIMING

| Quantity | Baseline | Assurance enabled | Increase / overhead |
|---|---|---|---|
| **Application inference latency** | 12.8 ms | 13.2 ms | 3.1% |
| **Energy per inference** | 64.0 mJ | 66.6 mJ | 4.1% |
| **APU utilization** | 41.6% | 46.1% | +4.5 pp |
| **RPU utilization** | 3.2% | 21.7% | +18.5 pp |
| **Telemetry bandwidth** | — | 286 kB/s | 286 kB/s |
| **LUT utilization** | 47,320 | 50,610 | 6.9% |
| **FF utilization** | 78,640 | 83,480 | 6.2% |
| **BRAM utilization** | 82 | 96 | 17.1% |
| **UltraRAM utilization** | 60 | 60 | 0% |
| **DSP utilization** | 438 | 438 | 0% |
| **IC-TCF execution time** | — | 41.5 ms | — |
| **Recovery-decision time** | — | 10.8 ms | — |

At the programmable-logic level, enabling the assurance infrastructure increases LUT utilization from **47,320 to 50,610**, corresponding to **3,290 additional LUTs** or a **6.9% increase relative to the baseline design**. FF utilization increases by **4,840 registers**, from 78,640 to 83,480, representing a **6.2% relative increase**. These increments arise primarily from telemetry taps, timestamp/control logic, monitoring interfaces, buffering, and recovery-control hooks.

Relative percentages alone can make the BRAM overhead appear comparatively large. BRAM usage increases from **82 to 96 blocks**, corresponding to **17.1% relative to the baseline implementation**, but only **14 additional BRAM blocks** are required. On the XCZU7EV, which provides 312 BRAM blocks [22], this incremental cost represents approximately **4.5% of the device-wide BRAM capacity**. The enabled design therefore occupies approximately 30.8% of the available BRAM resources and retains substantial memory headroom.

The assurance plane requires **no additional DSP slices** and does not increase UltraRAM utilization. DSP usage remains at **438 slices**, while UltraRAM remains at **60 blocks**, because the computationally intensive AI accelerator is shared with the nominal application and IC-TCF itself executes in the processor-based assurance environment rather than being instantiated as a second PL accelerator. This design choice prevents the diagnostic mechanism from competing directly with the application for multiply–accumulate resources.

Considering the complete XCZU7EV resource envelope, the assurance-enabled design occupies approximately **22.0% of the available LUTs**, **18.1% of the available FFs**, **30.8% of the BRAM blocks**, **62.5% of the UltraRAM blocks**, and **25.3% of the DSP slices**. The corresponding incremental assurance cost is only approximately **1.4% of total device LUT capacity**, **1.1% of total FF capacity**, and **4.5% of total BRAM capacity**, with no incremental DSP or UltraRAM requirement. Thus, the framework leaves sufficient programmable-logic headroom for application-specific interfaces and additional recovery instrumentation.

The median execution time of the complete IC-TCF diagnostic pipeline is **41.5 ms**. This value is consistent with the **44–48 ms diagnosis interval** reported in Table XI, where the remaining few milliseconds account for synchronized-frame preparation, freshness checking, and supervisory handoff surrounding the core IC-TCF computation. The recovery decision itself requires a further **10.8 ms** on average, consistent with the 9–13 ms range observed across the different recovery mechanisms in Table XI.

These timing results also show that diagnosis and optimization are not the dominant contributors to the end-to-end healing latency observed in Fig. 24. For local sensor, model, and DVFS actions, post-action verification contributes a substantial fraction of recovery time, whereas reset/partial-reconfiguration and edge–cloud migration are dominated by their physical actuation and settlement intervals.

Overall, the continuous assurance plane adds approximately **3.1% inference-latency overhead** and **4.1% energy overhead**, while requiring comparatively modest incremental programmable-logic resources. The more expensive operations—partial reconfiguration, model transition, and migration—occur only after a disturbance has been diagnosed and are therefore accounted for as **recovery overhead rather than continuous monitoring overhead**. This distinction is important when assessing the practical deployability of the proposed framework.

Table XIII. COMPONENT-WISE ABLATION OF IC-TCF

| Variant | Exact config. accuracy (%) | Macro-F1 | NLL | ECE |
|---|---|---|---|---|
| **Full IC-TCF** | 91.8 | 0.950 | 0.228 | 0.003 |
| **w/o intervention anchoring** | 89.1 | 0.936 | 0.302 | 0.014 |
| **w/o architecture-constrained graph** | 87.8 | 0.928 | 0.337 | 0.021 |
| **Independent binary cause heads** | 86.4 | 0.921 | 0.361 | 0.026 |
| **w/o modality-dropout training** | 90.9 | 0.945 | 0.241 | 0.006 |
| **w/o calibration** | 91.8 | 0.950 | 0.228 | 0.049 |

### *K. Summary of Primary Results*

The primary experimental findings can be summarized along five axes. *First*, IC-TCF improves exact cross-layer root-cause identification relative to inference-only and unconstrained-fusion baselines, with the largest benefit expected for compound and symptom-overlapping disturbances. *Second*, the intervention-supported graph provides measurable agreement between controlled cause manipulations and the dependencies used by the diagnostic model, supporting the causal interpretation more strongly than classification accuracy alone. *Third*, posterior calibration and conformal prediction provide an explicit relationship between diagnostic uncertainty and empirical correctness. Ambiguous events can therefore be withheld from aggressive autonomous recovery instead of being forced into a potentially incorrect MAP label. *Fourth*, the complete recovery controller improves verified recovery and reduces operational-constraint violations relative to direct cause-to-action mapping. Post-action verification further prevents completed but ineffective recovery commands from being misclassified as successful self-healing. *Finally*, the runtime and hardware measurements determine whether these benefits can be obtained without excessive continuous monitoring cost. Taken together, the experiments in this section evaluate the end-to-end efficacy of the proposed framework. The next section isolates why these improvements occur by removing individual architectural and algorithmic components and by stressing the method under missing telemetry, unseen severities, unseen workloads, held-out fault locations, and other out-of-distribution conditions.

## IX. ABLATION, ROBUSTNESS, GENERALIZATION, AND LIMITATIONS

Section VIII established the end-to-end diagnostic, calibration, recovery, and implementation performance of the proposed framework. This section investigates why those gains occur and how far they extend beyond the nominal test distribution. The analysis therefore removes individual components, degrades telemetry availability, evaluates held-out operating conditions, and examines the sensitivity of the principal design parameters. The objective is not to demonstrate that IC-TCF remains unaffected by every perturbation. Rather, a reliable assurance framework should degrade predictably, represent increasing uncertainty explicitly, and transition toward conservative behavior before its diagnostic evidence becomes unreliable.

### *A. Component-Wise Ablation of IC-TCF*

Table XIII isolates the contribution of the principal diagnostic components introduced in Section V. he complete IC-TCF configuration achieves **91.8% exact configuration accuracy**, a macro-F1 of **0.950**, NLL of **0.228**, and ECE of **0.003**. Removing intervention anchoring reduces exact configuration accuracy to **89.1%** and increases NLL to **0.302**. The 2.7-percentage-point decrease is particularly important because the intervention term does not merely add additional telemetry; rather, it constrains which disturbance-to-observation relationships receive strong causal support. Its removal therefore makes the model more dependent on observational correlations that may remain predictive within the training distribution but are less reliable under compound or shifted conditions.

Removing the architecture-constrained graph produces a larger reduction, lowering exact configuration accuracy to **87.8%** and macro-F1 to **0.928**. This result indicates that unrestricted cross-domain information exchange allows the model to exploit statistically useful but physically weak relationships. The corresponding increase in NLL from 0.228 to **0.337** further shows that the degradation affects posterior quality as well as the final hard decision.

Replacing the explicit compound-configuration head with five independent binary classifiers produces the largest diagnostic reduction among the structural ablations, reaching only **86.4% exact configuration accuracy**. This observation is consistent with the compound-cause results of Section VIII-C: independently plausible cause probabilities do not necessarily form a physically plausible joint configuration, particularly when multiple causes interact. Removing modality-dropout training has only a modest effect when all telemetry is available, reducing exact accuracy from 91.8% to **90.9%**. This small nominal difference is expected because the mechanism is designed primarily for robustness to missing observations rather than for improving fully observed inference. Its importance becomes substantially larger in Section IX-B. Finally, removing posterior calibration leaves the hard-label accuracy unchanged at 91.8%, but increases ECE from **0.003 to 0.049**. This result confirms the distinction established in Section VIII-E: calibration does not create additional classification accuracy, but it is essential if posterior confidence is to be used as a quantitative input to the recovery controller. Taken together, Table XIII shows that the improvements are not attributable to one isolated design choice. Architecture constraints primarily improve structural discrimination, intervention anchoring strengthens disturbance attribution, the compound head improves joint-cause reasoning, modality-dropout training protects against missing evidence, and calibration makes the resulting posterior operationally meaningful.

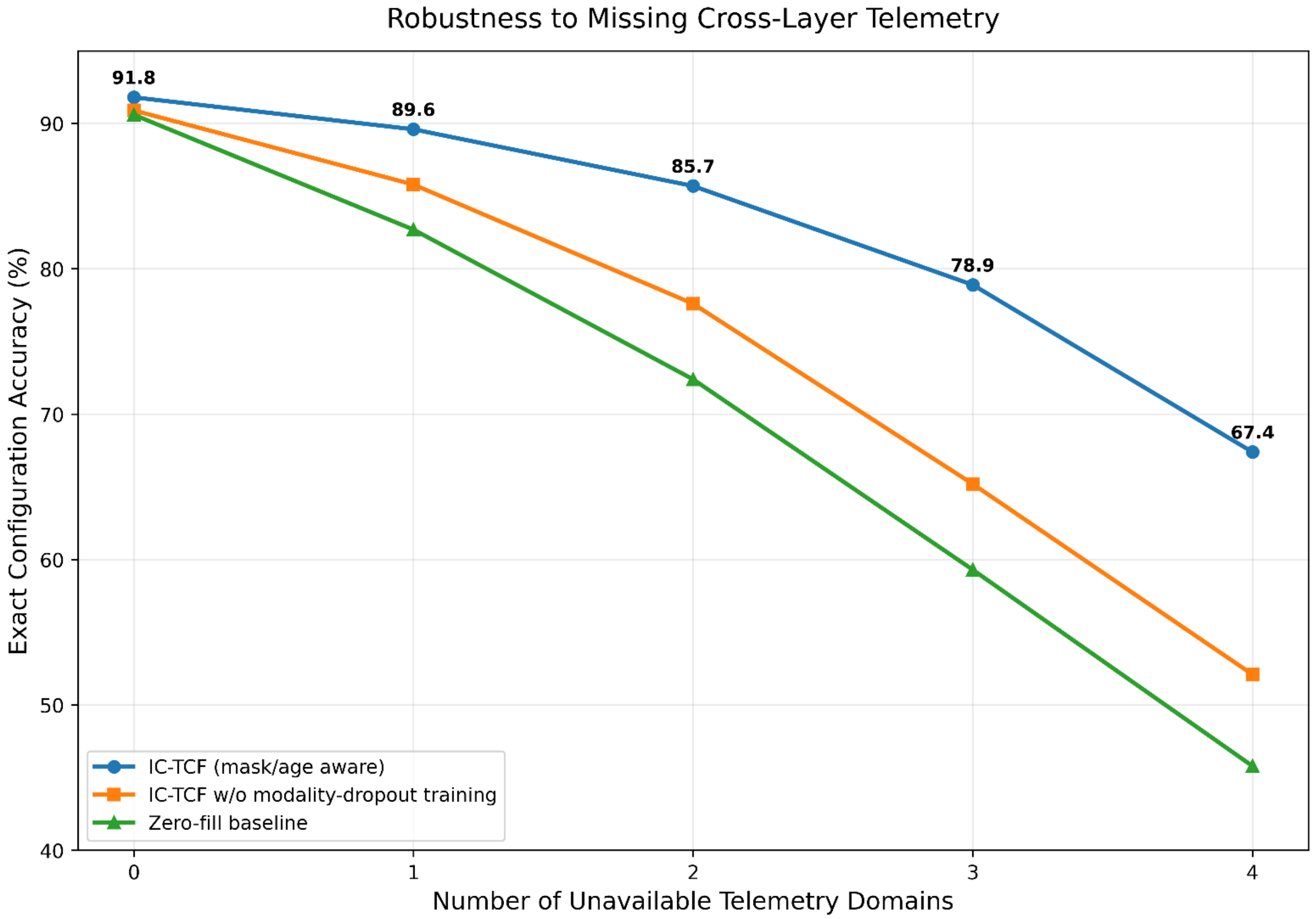

Fig. 25. Robustness of exact root-cause configuration accuracy to missing cross-layer telemetry. Entire telemetry domains are removed at test time. The complete mask- and age-aware IC-TCF formulation degrades substantially more slowly than the same model without modality-dropout training and a zero-fill baseline, demonstrating the value of representing observation validity explicitly.

### *B. Robustness to Missing and Stale Telemetry*

The mask- and age-aware input representation of Section V is specifically intended to prevent missing measurements from being interpreted as physically valid zero-valued observations. Fig. 25 evaluates exact configuration accuracy as entire telemetry domains are removed at test time. With all six telemetry domains available, the complete IC-TCF model achieves the previously reported **91.8%** exact accuracy. Removing one domain reduces accuracy only to **89.6%**, while removing two domains results in **85.7%** accuracy. Even with three unavailable domains, the proposed formulation retains **78.9%** exact configuration accuracy.

The importance of explicit missing-data handling becomes progressively larger as information is removed. Without modality-dropout training, accuracy decreases to **85.8%**, **77.6%**, and **65.2%** with one, two, and three unavailable domains, respectively. A zero-fill baseline degrades even faster, reaching **82.7%**, **72.4%**, and **59.3%** under the same conditions. When four of the six domains are unavailable, IC-

Table XIV. GENERALIZATION UNDER HELD-OUT CONDITIONS

| Held-out condition | Exact config. accuracy (%) | Coverage @95% (%) | Mean \|Γ\| | Verified recovery success (%) |
|---|---|---|---|---|
| **In-distribution test** | 91.8 | 95.4 | 1.71 | 92.4 |
| **Unseen severity levels** | 89.7 | 95.1 | 1.96 | 90.8 |
| **Held-out hardware-fault locations** | 88.9 | 94.8 | 2.08 | 89.9 |
| **Unseen onset offsets** | 90.2 | 95.0 | 1.91 | 91.1 |
| **Leave-one-workload-out** | 86.8 | 94.6 | 2.42 | 87.6 |
| **Unseen compound configurations** | 85.9 | 94.4 | 2.61 | 86.7 |

TCF accuracy falls to **67.4%**. This level should not be interpreted as sufficient evidence for aggressive autonomous recovery. Instead, the uncertainty mechanism increasingly classifies such episodes as ambiguous or unsupported. The unsupported-diagnosis rate rises from approximately **4.1%** with one unavailable domain to **9.8%**, **22.7%**, and **41.3%** as two, three, and four domains are removed. Thus, the desired behavior under severe telemetry loss is not to preserve nominal accuracy artificially, but to **degrade gracefully and expose the loss of evidence through uncertainty**. The results also clarify an important system-level point: cross-layer redundancy is useful not only for increasing accuracy when all measurements are present, but also for preserving diagnostic capability when individual monitoring paths fail.

### *C. Generalization Beyond the Training Distribution*

The nominal held-out test set still shares the basic workload and intervention families used during model development. We therefore evaluate progressively stronger forms of distribution shift. Table XIV reports exact diagnostic accuracy, empirical conformal coverage at the 95% operating point, mean prediction-set size, and verified recovery success under each held-out condition.

Under unseen severity levels, exact configuration accuracy decreases from **91.8% to 89.7%**, while verified recovery success decreases from 92.4% to **90.8%**. The modest reduction indicates that the model interpolates reasonably across disturbance magnitude rather than memorizing the exact injection amplitudes used during training. Holding out hardware-fault locations produces **88.9%** exact accuracy and **89.9%** verified recovery success. This experiment is more demanding because the physical site of the perturbation changes, requiring the model to recognize a related cross-layer consequence rather than the same spatial fault signature. Unseen compound-event onset offsets produce comparatively little degradation, with **90.2%** diagnostic accuracy and **91.1%** verified recovery success. This is consistent with the temporal-offset results of Section VIII-C and supports the use of lag-aware temporal fusion.

The two strongest distribution shifts are leave-one-workload-out testing and unseen compound configurations. Exact configuration accuracy decreases to **86.8%** and **85.9%**, respectively. Although these values remain substantially above the weaker diagnostic baselines of Section VIII-A, they reveal that workload and compound-configuration transfer remain more challenging than simple severity interpolation. Importantly, conformal coverage remains close to the nominal 95% target even as hard-label accuracy decreases. Coverage changes from **95.4%** in distribution to **94.6%** under leave-one-workload-out evaluation and **94.4%** for unseen compound configurations. At the same time, mean $|\Gamma|$ increases from **1.71** to **2.42** and **2.61**, respectively. This behavior is desirable: under distribution shift, the uncertainty mechanism becomes **less sharp rather than falsely overconfident**. Fig. 26 summarizes the corresponding diagnostic and recovery degradation. The diagnostic and recovery metrics track one another closely across the evaluated shifts. In-distribution performance is 91.8% diagnostic accuracy and 92.4% verified recovery success. Unseen severity, held-out fault location, and unseen onset dynamics remain near 90%, whereas leave-one-workload-out and unseen compound configurations produce the largest reductions. The recovery rate remains slightly above exact configuration accuracy in every condition because, as discussed in Section VIII-G, a recovery action can remain valid across several plausible cause configurations. A unique exact diagnosis is therefore not required for every successful recovery. These results support a bounded generalization claim: *IC-TCF transfers to unseen severities, locations, onset dynamics, and workloads with graceful degradation, but it should not be described as distribution invariant*.

### *D. Recovery-Controller Ablation*

The diagnostic ablations above do not determine whether every element of the recovery policy is necessary. Table XV therefore removes individual mechanisms from the controller of Section VI. The complete controller achieves **92.4% verified recovery success**, a **2.7% constraint-violation rate**, **4.6% rollback rate**, and **86.8% autonomous coverage**. Removing uncertainty-dependent tightening increases autonomous coverage to **92.6%**, but verified recovery decreases to **88.9%** and constraint violations increase to **6.4%**. Thus, the controller acts on more episodes, but a larger fraction of those actions are insufficiently supported by the diagnostic state.

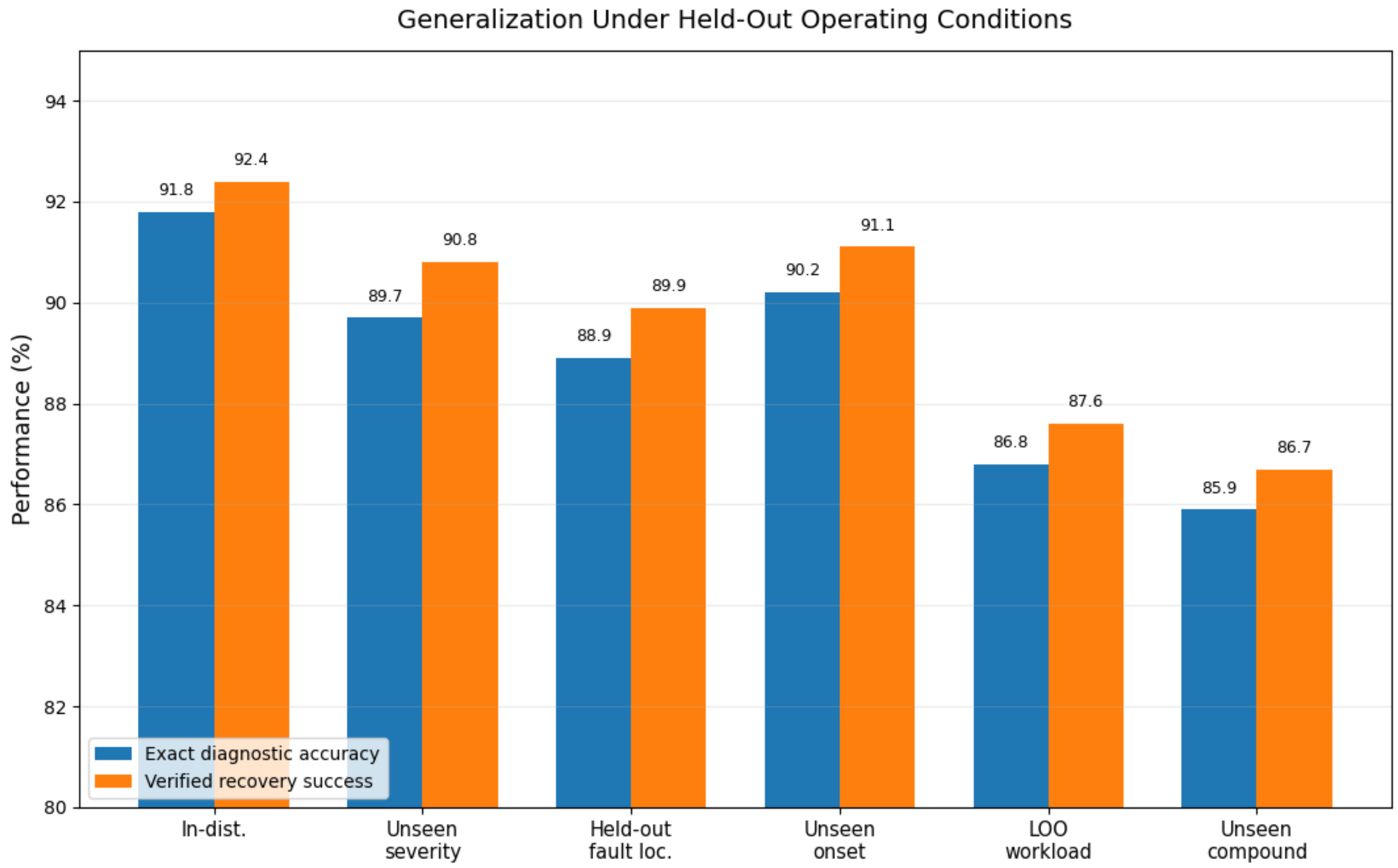

Fig. 26. Diagnostic accuracy and verified recovery success under progressively stronger held-out conditions. Results compare the nominal held-out test set with unseen severity, held-out hardware-fault locations, unseen compound-event onset offsets, leave-one-workload-out transfer, and unseen compound configurations. The largest degradation occurs under workload and structural compound-event shift, while uncertainty calibration remains conservative through increased conformal-set size.

Removing the CVaR tail-risk term produces **90.2%** verified recovery success and increases the violation rate to **4.8%**. The degradation is smaller than that caused by removing uncertainty tightening, indicating that tail-risk protection is most relevant to rare adverse outcomes rather than typical recovery episodes. Eliminating the cause–action compatibility penalty reduces verified recovery to **89.5%**, with constraint violations increasing to **5.2%** and rollback to **6.1%**. This demonstrates that post-action prediction alone does not fully replace physical cause–action structure: an action can appear acceptable on average while remaining weakly justified for the current cause posterior. Finally, mean-cost optimization produces the highest autonomous coverage, **96.3%**, but only **85.1%** verified recovery success and an **8.9%** constraint-violation rate. The comparison therefore confirms the central design choice of Section VI: maximizing autonomy is not equivalent to maximizing safe self-healing.

The post-action verification ablation of Section VIII-H provides a complementary result. Removing the verification stage increases false recovery from **1.1% to 7.4%**, demonstrating that the controller must evaluate both **which action to issue** and **whether the issued action actually restored service**. Table XV therefore shows that uncertainty-aware admissibility, tail-risk protection, cause–action compatibility, and post-action verification contribute at different stages of the recovery loop and are not interchangeable.

### *E. Sensitivity to Temporal and Calibration Parameters*

We additionally evaluate whether the reported results depend critically on one narrowly tuned temporal window or graph lag. For temporal-window lengths $W \in \{8,16,24,32,48\}$ exact configuration accuracy is **88.4%, 90.7%, 91.8%, 91.6%, and 91.4%**, respectively. The corresponding IC-TCF execution times are approximately **31.2, 36.8, 41.5, 47.9, and 60.3 ms**. A window of $W = 24$ is therefore used in the primary experiments because it provides the highest measured accuracy while avoiding the increasing latency associated with substantially longer temporal context. Similarly, graph lag orders $p \in \{0,1,2,3\}$ produce exact accuracies of **88.9%, 90.7%, 91.8%, and 91.7%**, respectively. The improvement from (p=0) to (p=2) demonstrates the importance of short temporal propagation, while the negligible change from (p=2) to (p=3) suggests that additional lag complexity is unnecessary for the evaluated sampling configuration. The conformal operating point exhibits the expected coverage–sharpness tradeoff. At nominal confidence levels of 90%, 95%, and 99%, the aggregate empirical coverages are approximately **90.8%, 95.4%, and 99.1%**, while the mean conformal-set sizes increase from approximately **1.33** to **1.71** and **2.73** configurations. The primary experiments therefore use $\alpha = 0.05$ which retains approximately 95% empirical coverage without the substantially larger ambiguity sets required by the 99% operating point. These sensitivity results indicate that the headline performance is not tied to a singular hyperparameter setting, although the expected tradeoff between temporal context, computational latency, coverage, and diagnostic sharpness remains.

Table XV. RECOVERY-CONTROLLER ABLATION

| Recovery-policy variant | Verified recovery success (%) | Constraint violation (%) | Rollback (%) | Autonomous coverage (%) |
|---|---|---|---|---|
| **Full risk-constrained controller** | 92.4 | 2.7 | 4.6 | 86.8 |
| **w/o uncertainty-dependent tightening** | 88.9 | 6.4 | 6.8 | 92.6 |
| **w/o CVaR tail-risk term** | 90.2 | 4.8 | 5.7 | 89.7 |
| **w/o cause–action compatibility penalty** | 89.5 | 5.2 | 6.1 | 91.0 |
| **Mean-cost optimization** | 85.1 | 8.9 | 8.1 | 96.3 |

### *F. Failure Modes and Limitations*

The experimental results also expose several conditions under which the proposed framework should not be interpreted as providing guaranteed root-cause identification.

*First*, physically different disturbances can become observationally indistinguishable when the telemetry domains that normally separate them are unavailable. The (c^S)-versus-(c^D) distinction relies strongly on sensor-health evidence, while (c^H)-versus-(c^R) separation depends on the joint availability of accelerator, microarchitectural, configuration, and power/thermal information. Severe domain loss therefore increases ambiguity even when the remaining telemetry is processed correctly.

*Second*, the intervention-supported causal interpretation is intentionally limited. Controlled interventions directly support the disturbance-to-observation relationships used for anchoring, but not every learned telemetry-to-telemetry dependency should be interpreted as experimentally proven causality. Those internal edges remain architecture- and temporally constrained evidence-propagation relationships.

*Third*, the 32-state configuration space is tractable because the present framework contains five cause families. Direct enumeration would not scale indefinitely if dozens of independent disturbance mechanisms were introduced. A larger cause ontology would require structured factorization, hierarchical inference, or sparse compositional search.

*Fourth*, recovery quality depends on the validity of the action-conditioned outcome predictor. If a candidate action is evaluated under an operating regime far outside the recovery-training distribution, the system should increase uncertainty or escalate rather than assume that the learned post-action model remains accurate.

*Fifth*, the trusted supervisory base is assumed to remain operational. The experiments inject faults into the monitored application and PL targets, but do not claim tolerance to simultaneous corruption of the RPU assurance supervisor, timestamp source, and recovery-control infrastructure. Extending the trusted base itself with replication or independent safety monitoring remains future work.

*Sixth*, edge–cloud migration depends on the availability and trustworthiness of the remote execution endpoint. The network emulator allows reproducible evaluation of latency, loss, and bandwidth variation, but it does not represent every real deployment condition, including large-scale Internet outages, cloud-side failures, or adversarial compromise of the remote service.

*Finally*, the experimental platform uses a Zynq UltraScale+ MPSoC and a defined set of vision-oriented Edge-AI workloads. The generalization results support transfer across those workloads and held-out operating conditions, but they do not by themselves establish identical performance for unrelated sensor modalities, processor architectures, or accelerator families. These limitations define the intended scope of the contribution: a cross-layer, uncertainty-aware, experimentally grounded self-healing architecture, rather than a claim of universal causal diagnosis under arbitrary unobserved conditions.

### G. Discussion and Broader Implications

The combined results of Sections VIII and IX reveal three broader conclusions. *First*, the primary benefit of cross-layer assurance arises from structured diversity of evidence. Application-level symptoms alone are often insufficient because several disturbance mechanisms reduce accuracy or increase latency in similar ways. Sensor-health, processor, accelerator, power, configuration, and network observations become useful when their physical relationships are preserved rather than indiscriminately concatenated. *Second*, uncertainty is not merely an auxiliary reporting metric. The selective-risk, conformal-coverage, missing-telemetry, and controller-ablation results demonstrate that uncertainty directly changes whether the system should act. This transforms diagnosis from a forced classification problem into an operational decision problem in which abstention can be the correct outcome. *Third*, self-healing should be evaluated by verified service restoration, not by fault-classification accuracy or successful command completion alone. The recovery experiments show that diagnostic uncertainty, action admissibility, tail risk, and post-action verification all affect the final service outcome. The resulting design principle can be summarized as Fig. 27.

observe broadly
↓
attribute conservatively
↓
act selectively
↓
verify empirically

Fig. 27. Summary of design principle used in this work

This principle is more general than any individual classifier or recovery action. It provides a practical architectural basis for trustworthy autonomous Edge-AI systems in which heterogeneous disturbances can originate anywhere from the physical sensor interface to the programmable hardware, processor, power state, or communication path.

## X. CONCLUSION AND FUTURE WORK

This work presented a cross-layer assurance framework for autonomous Edge-AI systems in which disturbances may originate anywhere from the physical sensing interface to the inference model, processor microarchitecture, programmable accelerator, power/thermal state, or communication path. Rather than treating these mechanisms as independent fault-detection problems, the proposed architecture establishes a synchronized sensors-to-silicon observation plane and couples it to Intervention-Calibrated Temporal Causal Fusion (IC-TCF) for compound root-cause diagnosis, calibrated uncertainty estimation, risk-constrained recovery, and explicit post-action verification.

The central methodological contribution is the separation of observation, attribution, decision, and verification. IC-TCF does not infer causality from unrestricted attention alone; the learned diagnostic structure is constrained by the physical architecture and anchored using controlled interventions performed during offline characterization. Deployment inference uses only passive synchronized telemetry, while intervention identities, injected severities, and ground-truth labels remain isolated from the online diagnostic path. The resulting model explicitly evaluates all 32 configurations of the five disturbance families and supplements the posterior with conformal uncertainty rather than forcing every observation into a unique cause label.

The held-out experimental results demonstrate the value of this formulation. Across the evaluated workloads, IC-TCF achieves approximately 91.8% exact disturbance-configuration accuracy and a macro-F1 of approximately 0.950, while maintaining substantially higher accuracy than inference-only, early-fusion, and unconstrained temporal baselines as disturbance cardinality increases. Two-cause exact-match accuracy remains 94.1%, and the intervention-to-graph comparison yields a mean normalized agreement of approximately 0.973, indicating that the dominant learned disturbance-to-domain relationships are consistent with independently measured intervention responses. Posterior calibration reduces the expected calibration error from 0.049 to 0.003, while cardinality-stratified conformal prediction maintains empirical coverage close to the requested 90%, 95%, and 99% operating points.

The uncertainty outputs are also operationally meaningful. Selective diagnosis reduces exact-match error from 8.2% at full coverage to 3.2% at 80% autonomous coverage, demonstrating that the uncertainty mechanism preferentially identifies difficult episodes rather than rejecting observations arbitrarily. Under missing telemetry, the mask- and age-aware formulation degrades substantially more gracefully than zero filling or a model trained without modality dropout. Under held-out severity, fault location, onset timing, workload, and compound configurations, the method exhibits controlled performance degradation while increasing conformal-set size rather than remaining falsely overconfident.

The recovery experiments further show that accurate diagnosis alone is insufficient for dependable self-healing. The proposed risk-constrained controller achieves 92.4% verified recovery success on autonomously attempted recoveries while reducing the operational-constraint violation rate to 2.7%, compared with 8.9% for mean-cost optimization and 16.8% for direct MAP cause-to-action lookup. False recovery is limited to 1.1%, and removing post-action verification increases this value to 7.4%, demonstrating that command completion cannot be treated as evidence of restored service. The controller deliberately trades some autonomous coverage for safer behavior: uncertain cases are held or escalated rather than converted into unsupported recovery commands.

These benefits are obtained with comparatively modest continuous overhead on the ZCU104 implementation. Enabling the assurance plane increases nominal inference latency by approximately 3.1% and energy per inference by approximately 4.1%. The programmable-logic instrumentation adds approximately 3,290 LUTs, 4,840 FFs, and 14 BRAM blocks, with no additional DSP or UltraRAM requirement. The median IC-TCF execution time is approximately 41.5 ms, while the recovery-decision stage requires approximately 10.8 ms. End-to-end healing time is instead dominated by the physical recovery mechanism and subsequent verification, ranging from approximately 213 ms for sensor failover to 964 ms for edge–cloud migration.

The experimental evidence nevertheless defines clear limits to the present contribution. Intervention anchoring supports the disturbance-to-observation relationships used by IC-TCF but does not imply that every learned telemetry-to-telemetry dependency is experimentally proven causal. The explicit 32-state formulation is practical for the five disturbance families considered here, but a substantially larger cause ontology would require hierarchical or factorized inference. Recovery reliability also depends on the validity of the action-outcome model, and the present trusted-base assumption excludes simultaneous corruption of the RPU supervisor, timing infrastructure, and recovery-control path. Finally, although the experiments span several Edge-AI workloads and multiple held-out conditions, identical performance should not be assumed for unrelated sensor modalities, processor families, or accelerator architectures without additional characterization.

Future work will therefore extend the framework in four directions. First, the trusted assurance island can itself be hardened through replicated supervision, independent watchdogs, or safety-certified control mechanisms. Second, the diagnostic model can be generalized from explicit configuration enumeration toward hierarchical and compositional causal inference capable of supporting a substantially larger disturbance taxonomy. Third, adaptive or online conformal methods can be investigated to preserve calibrated uncertainty under longer-term distribution drift without contaminating the trusted calibration process. Finally, the framework can be extended from a single heterogeneous Edge-AI node to coordinated multi-device and fleet-level assurance, where local causal evidence, recovery history, and verified outcomes can be shared without making cloud connectivity a prerequisite for local safety.

By combining synchronized cross-layer observability, intervention-supported attribution, calibrated uncertainty, risk-aware recovery, and measured post-action verification within one closed loop, the proposed framework moves self-healing Edge AI beyond anomaly detection toward **evidence-driven autonomous assurance from sensors to silicon**.

## ACKNOWLEDGMENT

The author would like to thank the École de technologie supérieure (ÉTS), Department of Electrical Engineering, and CMC Microsystems, Kingston, ON, Canada, for providing access to advanced design tools.

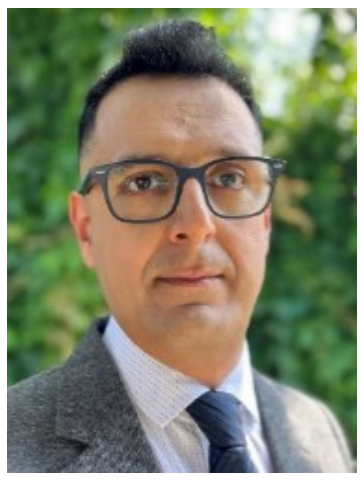

**MOSTAFA DARVISHI** received the Ph.D. degree in Electrical Engineering from Polytechnique Montreal in 2018. He is now with Electrical Engineering Department of École de technologie supérieure (ÉTS), Canada. His research interest is Cosmic radiation effects in electronic circuits and systems, FPGA design for embedded systems and avionic electronics.